\documentclass[twoside,11pt]{article}

\usepackage{jmlr2e}

\usepackage{array}
\usepackage{url}

\usepackage{times}
\usepackage{siunitx}
\usepackage{algorithmicx}
\usepackage{algpseudocode}
\usepackage{mathtools}
\usepackage{setspace}
\usepackage{caption}
\usepackage{subcaption}
\usepackage{booktabs}
\usepackage{adjustbox}
\usepackage{framed}
\usepackage{dsfont}

\usepackage{todonotes}

\usepackage{longtable}

\DeclareMathOperator*{\argmax}{argmax}
\DeclarePairedDelimiter{\ceil}{\lceil}{\rceil}

\newcolumntype{A}{>{\centering\arraybackslash}m{1cm}}

\newcolumntype{B}{>{\centering\arraybackslash}m{1.5cm}}

\newcolumntype{D}{>{\centering\arraybackslash}m{2cm}}

\renewcommand*\Call[2]{\textproc{#1}(#2)}

\algnewcommand{\IIf}[1]{\State\algorithmicif\ #1\ \algorithmicthen}
\algnewcommand{\ElseIIf}{\unskip\ \algorithmicelse}
\algnewcommand{\EndIIf}{\unskip\ \algorithmicend\ \algorithmicif}

\usepackage{algorithm}
\makeatletter
\newcommand\fs@norules{\def\@fs@cfont{\bfseries}\let\@fs@capt\floatc@ruled
	\def\@fs@pre{}%
	\def\@fs@post{}%
	\def\@fs@mid{\kern3pt}%
	\let\@fs@iftopcapt\iftrue}
\makeatother
\floatstyle{norules}
\restylefloat{algorithm}

\newcolumntype{H}{@{}>{\lrbox0}l<{\endlrbox}}

\jmlrheading{18}{2017}{}{07/17}{TBD}{Tom Rainforth and Frank Wood}

\ShortHeadings{Canonical Correlation Forests}{Rainforth and Wood}
\firstpageno{1}

\begin{document}

\title{Canonical Correlation Forests}

\author{\name Tom Rainforth \email twgr@robots.ox.ac.uk \\
       \addr Department of Engineering Science\\
       University of Oxford\\
       Parks Road, Oxford, OX1 3PJ, UK
       \AND
       \name Frank Wood \email fwood@robots.ox.ac.uk \\
       \addr Department of Engineering Science\\
       University of Oxford\\
       Parks Road, Oxford, OX1 3PJ, UK}

\editor{}

\maketitle

\begin{abstract} %
	We introduce canonical correlation forests (CCFs), a new decision tree ensemble method for classification and regression. 
	Individual canonical correlation trees are binary decision trees with hyperplane splits based on local canonical correlation coefficients calculated during training. 
	Unlike axis-aligned alternatives, the decision surfaces of CCFs are not restricted to the coordinate system of the inputs features and therefore more naturally represent data with correlated inputs. 
		CCFs naturally accommodate multiple outputs, provide a similar computational complexity to random forests,
		and inherit their impressive robustness to the choice of input parameters.
	As part of the CCF training algorithm, we also introduce projection bootstrapping, a novel alternative to bagging for oblique decision tree ensembles which maintains use of the full dataset in selecting split points, often leading to improvements in predictive accuracy.
	Our experiments show that, even without parameter tuning, CCFs out-perform axis-aligned random forests and other state-of-the-art tree ensemble methods on both classification and regression problems, delivering both improved predictive accuracy
	and faster training times.
	We further show that they outperform all of the 179 classifiers considered in a recent extensive survey.
\end{abstract}


\section{Introduction}
\label{sec:Intro}

\begin{figure*}[p]
	\centering
	\begin{subfigure}[t]{0.48\textwidth}
		\caption{Single CART}
		\includegraphics[width=0.95\textwidth]{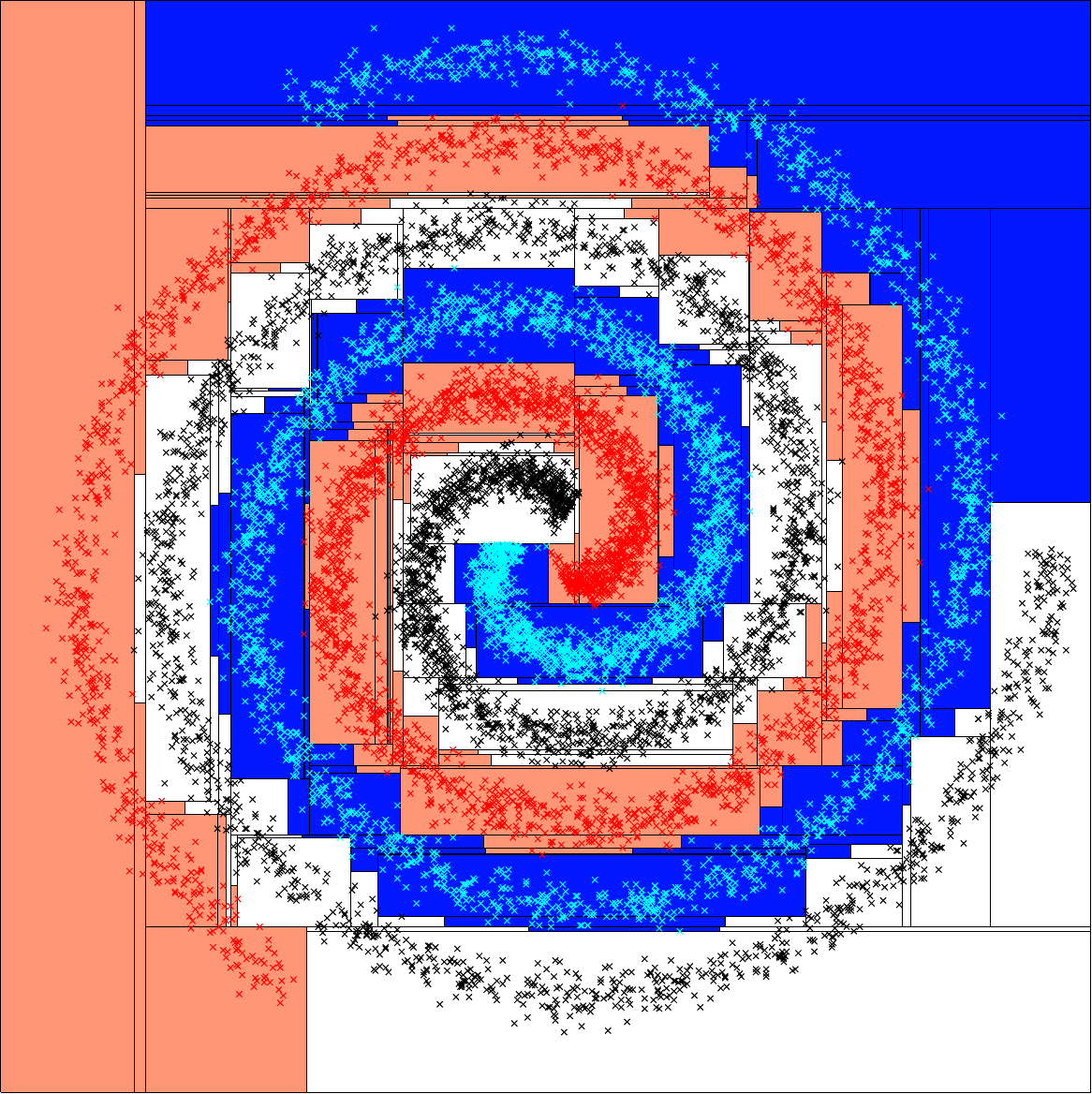}
		\centering
	\end{subfigure}
	~
	\begin{subfigure}[t]{0.48\textwidth}
		\caption{RF with 200 Trees}
		\includegraphics[width=0.95\textwidth]{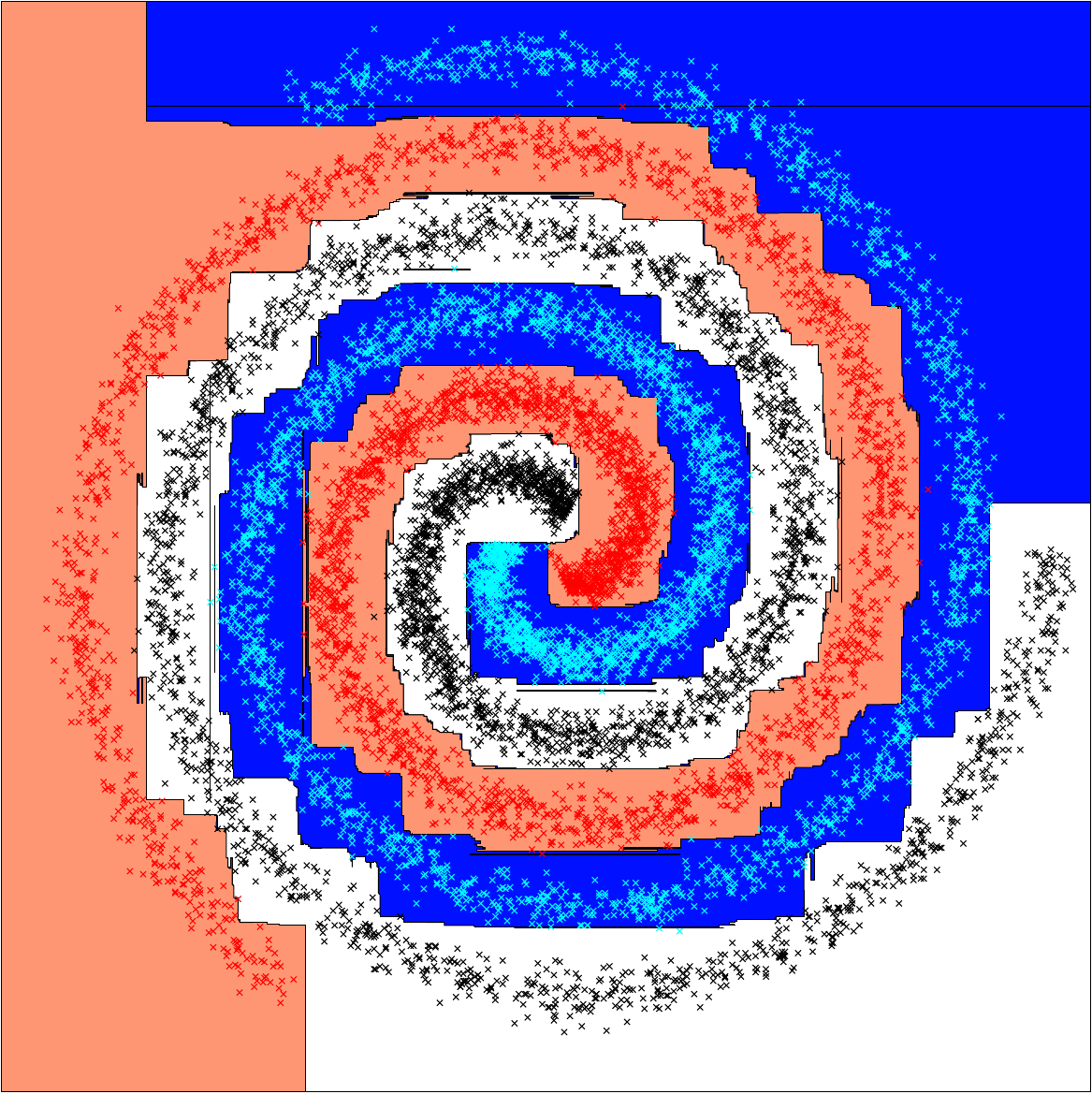}
		\centering
	\end{subfigure}	
	\\
	\vspace{10pt}
	\begin{subfigure}[t]{0.48\textwidth}
		\caption{Single CCT}
		\includegraphics[width=0.95\textwidth]{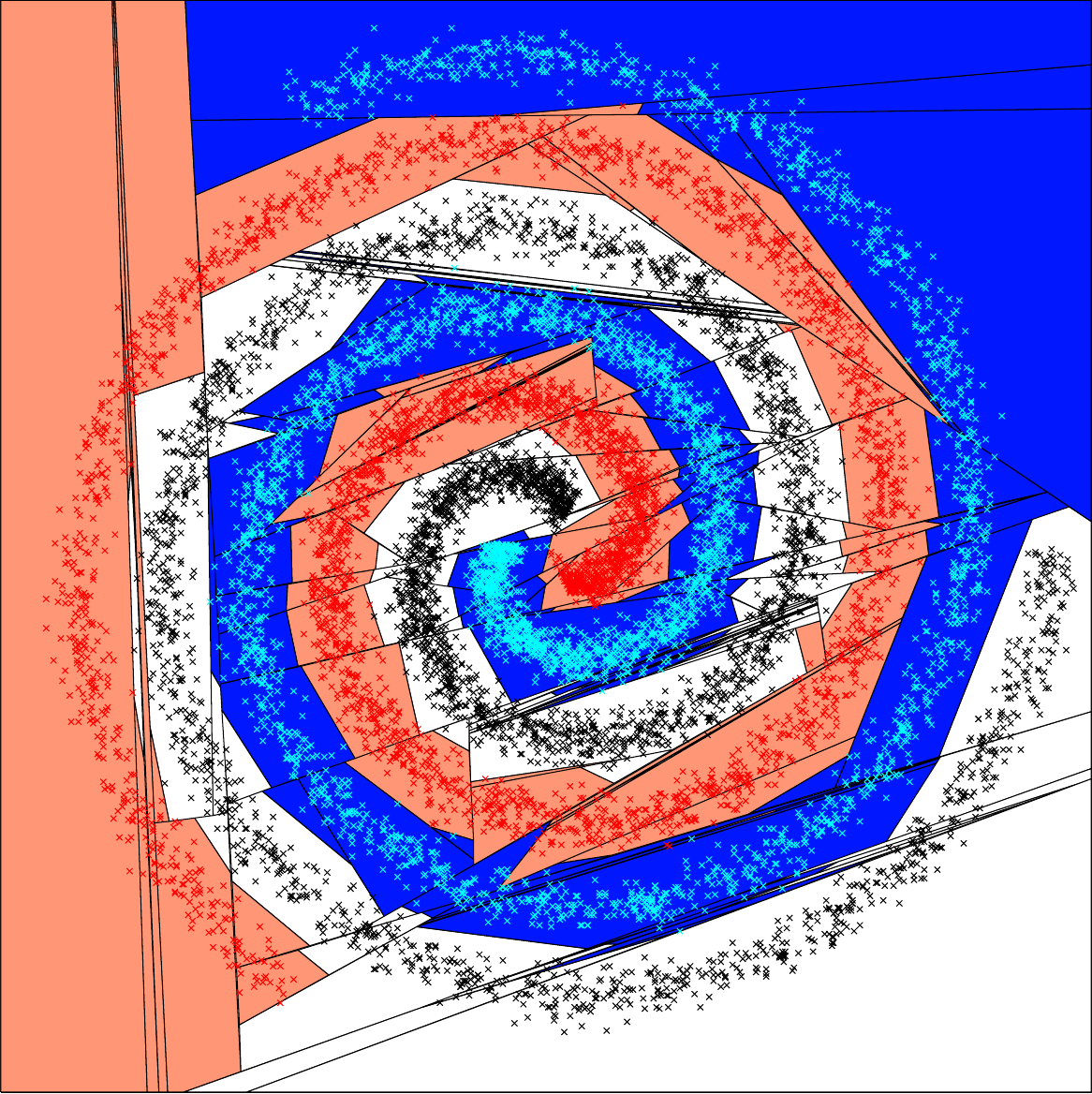}
		\centering
	\end{subfigure}	
	~
	\begin{subfigure}[t]{0.48\textwidth}
		\caption{CCF with 200 Trees}
		\includegraphics[width=0.95\textwidth]{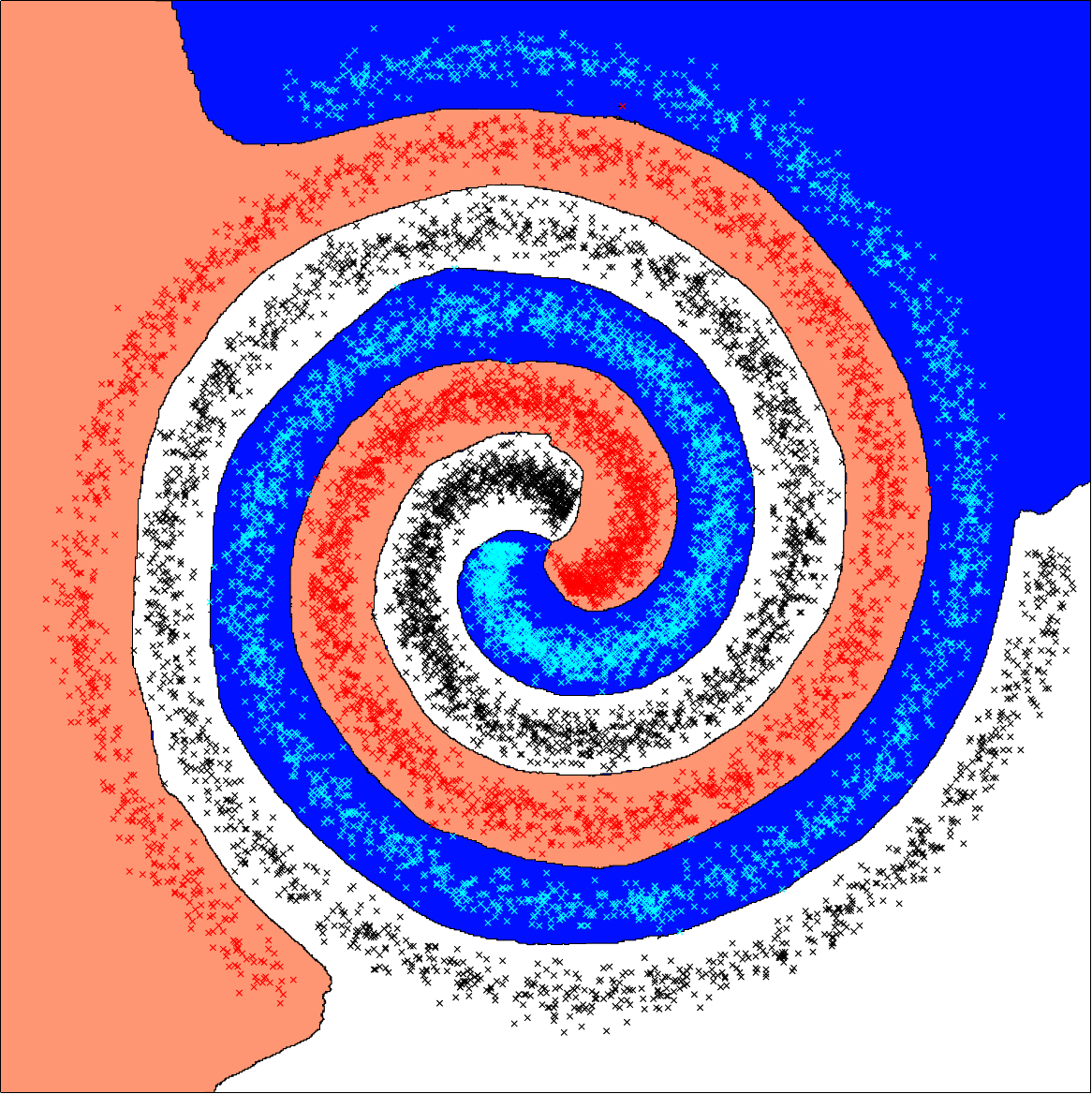}
		\centering
	\end{subfigure}		
	\caption{Decision surfaces for classification of artificial spirals dataset.
		The blue, red, and black crosses correspond to data points from three
		separate classes.  The shading of the background shows the class predicted
		for that area by the respective classification algorithm.
		(a) Shows the hierarchical partitions and surface for a single axis aligned tree while (b) shows the 
		RF decision surface arising from averaging over 200 axis aligned trees using the RF algorithm.  
		(c) Shows a single canonical correlation tree (CCT) and (d) shows the 
		CCF decision surface arising from averaging over 200 CCTs. Comparing  (b) and (d)		
		shows that the CCF has a smoother
		decision surface which better represents the data than the RF.  More precisely the
		decision surface is piecewise linear for the CCF and piecewise orthogonal for the RF.
		\label{fig:decisionSurface}}
\end{figure*}

Decision tree ensemble methods such as random forests (RF) \citep{breiman2001random},
extremely randomized trees \citep{geurts2006extremely}, and boosted decision trees \citep{friedman2001greedy} are widely employed methods for classification and 
regression due to their scalability, fast out of sample prediction, and 
tendency to require little parameter tuning.  
In many cases, they give predictive performance close to, or even equalling, state of the art when used in an out-of-the-box fashion \citep{fernandez2014we}, i.e.
when parameters are set to default values.
The individual trees used in such algorithms are, however, typically axis-aligned, restricting the ensemble decision surfaces
to be piecewise axis-aligned, even when there is little evidence for this in the data.  Canonical correlation forests (CCFs) overcome this problem by instead using carefully chosen hyperplane splits, leading to a more powerful classifier (or regressor) that naturally incorporates correlation between the features, an example for which shown in Figure~\ref{fig:decisionSurface}.
In this paper we demonstrate that this innovation regularly leads to a significant increase in out-of-sample predictive accuracy over previous state-of-the-art tree ensemble methods, whilst maintaining speed and black-box applicability.
Furthermore, we demonstrate that running CCFs without parameter tuning outperforms all of the 179 classifiers considered in the recent survey of \cite{fernandez2014we} over a large selection of datasets, even when
parameter tuning is employed by the competing methods.  
We provide an open source code for CCFs,\footnote{\scriptsize \url{https://github.com/twgr/ccfs/}} delivering both training and testing in a single short line of code.  Usage requires no expertise on the part of the user - all results presented in the paper are generated by using the package out-of-the-box, requiring only the data as input.

The two key factors underlying the performance of decision tree ensembles are the accuracy of the 
individual trees and the diversity in their predictions \citep{kuncheva2003measures,elghazel2011trading}.  
These often form conflicting aims as methods used for decorrelating tree predictions, such as
bagging \citep{breiman1996bagging} or random subspacing \citep{ho1998subspace}, typically rely on randomizing
the tree training process, degrading the accuracy of individual trees.
Not providing enough randomization in the tree training process
causes high levels of correlation between tree predictions, diminishing the gains 
from aggregation and leading to reduced predictive accuracy and potentially overfitting \citep{brown2005diversity}.  
Inserting too much randomness degrades the predictive accuracy of the individual trees, again reducing the 
performance of the ensemble \citep{sollich1996learning}.

The use of axis-aligned splits by most decision tree ensemble methods can be
detrimental to both these key factors.
Individual tree performance can be hampered by an inability of axis-aligned splits to provide a
good representation of the true decision surface, while the smaller set of possible splits 
reduces the ensemble diversity.
The effect of the former on the ensemble performance can be particularly pronounced, 
resulting in sensitivity in the predictive accuracy 
to rotation of, and to correlation between, the input features \citep{menze2011oblique}.  
Recent work has looked to overcome this by randomly rotating or projecting the input features,
either as preprocessing step before training each tree \citep{blaser2016random}, or
independently at each decision node \citep{breiman2001random,tomita2015randomer,tomita2017roflmao}.
Such approaches can improve the ensemble diversity, alleviate sensitivity to rotation, 
and improve performance for many datasets. 
However, these gains are by no means ubiquitous, with the approaches ignoring potentially useful information in the choice of axes, such as when there are unordered categorical features.
Furthermore, they do little to overcome the sensitivity of the performance 
to correlation between the input features. 

CCFs instead use carefully chosen hyperplane splits based on applying canonical correlation analysis (CCA) \citep{canoncorr} at each decision node during training.  
By using CCA, both output information and the correlation between the input features is naturally incorporated into the choice
of projection.  
Further, as the hyperplane splits are calculated at each node separately, CCFs are better at incorporating local correlations than previous approaches.  
This also has the effect of reducing correlation between trees
compared to axis-aligned approaches, as successive splits are no longer constrained to be orthogonal.
CCFs maintain a very similar computational complexity to RFs and our empirical timings suggest that they are typically faster to train, due to evaluating less candidates splits on average at each node.
Given that similar overall accuracy can often be achieved with only a small fraction of the number of trees
compared with existing approaches, CCFs offer
the potential for both substantial improvements in predictive accuracy and training time.

\section{Background}
\label{sec:background}


\subsection{Decision Tree Ensembles}
\label{sec:decTree}
%
%

A decision tree is a predictive model that imposes sequential divisions of an input space to form a set of partitions known as leafs, each containing a local classification or regression model. Out-of-sample prediction is performed by using the partitioning to assign each new input to a particular leaf and then using the corresponding local predictive model.  Typically, the leaf models are taken to be independent of each other and the outputs are assumed to be independent of input features given the leaf assignments. 

Classical decision tree learning algorithms work in a greedy top down fashion, exhaustively searching the space of unique axis-aligned splits and choosing the best based on a splitting criterion, such as the Gini gain or information gain used in CART \citep{CART} and C4.5 \citep{C4.5} respectively.  
Once a split is chosen, it is used to assigned each data point in the training set 
to one of the newly generated child nodes. The process then recurses in a self-similar manner for each of the child nodes,
continuing until no further split is advantageous or some user-set limit is reached for all the generated nodes. 
For classification with continuous features this typically only occurs once each leaf is ``pure," containing only data points of a single class.   When used as individual classifiers or regressors, trees are usually ``pruned" after being grown to prevent overfitting.  This is a regularization 
process which involves collapsing sub-branches of the tree to single leaf nodes.

It was established by \cite{ho1995random} that combining individual trees to form a decision tree ensemble, or forest, can simultaneously improve predictive performance and provide regularization against overfitting without the need for pruning.  In a forest, each tree is separately trained, with prediction based on 
averaging predictions from the individual trees,
resulting in a predictor that typically outperforms any single constituent tree \citep{rokach2010ensemble}.  As classical decision tree algorithms are deterministic procedures, such combination requires the introduction of probabilistic elements into the generative process to prevent identical trees.  One possible approach, the random subspace method \citep{ho1995random,ho1998subspace}, involves only searching splits using a randomly selected subset of the features at each node.  Bagging \citep{breiman1996bagging} instead trains each tree on a bootstrap sample of the original dataset.  Arguably the most widely used decision tree ensemble approach, random forests
(RFs) \citep{breiman2001random}, combines these randomization approaches, providing improved out-of-sample predictive 
performance to using either approach in isolation.

\subsection{Oblique Decision Trees and Forests}

Oblique decision trees (ODTs) extend classical decision trees by splitting using linear combinations of the available features.  Some algorithms such as OC1 \citep{murthy1994OC1} attempt to directly optimize for the hyperplane representing the best split, while others, such as functional trees \citep{gama2004functional} and QUEST \citep{loh1997QUEST}, carry out a linear discriminant analysis (LDA) \citep{fisher1936lda,rao1948utilization} to find a projection which optimizes the discriminant criterion and then search over possible splits in this projected space.  Although ODTs generally produce better results than axis aligned trees, many existing algorithms suffer from a number of common issues such as a failure to effectively deal with multiple classes, numerical instability or significant increase in computational cost compared with an axis aligned alternative.
Many also only apply to classification problems and do not naturally extend to regression or multiple output settings.

A number of approaches for constructing oblique decision forests using random projections 
have been suggested in the literature.
Random rotation ensembles \citep{blaser2016random} apply a
separate random rotation to the input features for each tree in the ensemble, training the tree
in the standard axis-aligned 
way on this rotated space.  As such, each tree uses orthogonal splits, but in a different co-ordinate system.
Forest-RC \citep{breiman2001random} instead uses random linear combinations of subsets of the input features 
to generate hyperplane splits at each node, such that each tree is itself an oblique decision tree.  
Randomer forests \citep{tomita2015randomer}
also use random linear combinations of features at each node, but instead of sub-sampling fixed numbers
of features for each linear combination, they use sparse random projections of all the features.  
All these methods
have been shown to offer improvements in predictive accuracy to RFs on some datasets, but
also worse performance on others.  In particular, they tend to perform
poorly on problems with unordered categorical features.

Rotation forests \citep{rodriguez2006rotation} take a more structured approach.  Instead of using
bagging or random subspacing,  they apply a rotation pre-processing step using 
principle component analysis (PCA) on random groupings of the input features.
This significantly reduces sensitivity to correlation between features and delivers significant performance improvements over both RFs and randomly rotated alternatives.
However, this improvement comes at a significant computational cost, as the method
considers all the features when splitting each node, as opposed to using a small
subset of them as done by methods employing random subspacing.
Furthermore, as all splits in a particular tree are orthogonal to one another and as PCA does not incorporate class information,
the performance of rotation forests is still sensitive to localized or class dependent correlations.
Perhaps because of the faster speed of training, axis-aligned RFs are still more prevalently
used than rotation forests, despite the typically improved predictive accuracy of the latter.

\cite{lemmond2008discriminant} and \cite{menze2011oblique} both introduced the idea of creating forests of oblique decision trees for classification using splits based on LDA projections, providing noticeable
improvements in predictive accuracy over RFs.
  However, both methods only apply to binary classification and neither carries out LDA in a manner that is both numerically stable and computationally efficient.  The latter paper also introduces the idea of carrying out a
  ridge regression adjusted LDA where there is regularization towards the principle component directions.  However, their results suggest no advantage is gained by this regularization compared to using LDA directly.

In work developed independently to our own, \citet{zhang2014random} consider three more generally applicable approaches: PCA-RF, LDA-RF, and RF-ensemble.
PCA-RF uses PCA to project the features at each node, and then calculates splits in this projected space.
LDA-RF instead uses a multiclass form of LDA to calculate the projections.
RF-ensemble constructs an ensemble where each tree uses either PCA, LDA, or axis-aligned splits depending on which
approach provided the best split at the root node for that tree.  They show that all three approaches deliver
regular improvements over RFs on a selection of (predominantly) UCI datasets~\citep{Lichman2013UCI}, though
they do not provide comparisons with rotation forests.  Their results also suggest that LDA-RF outperforms
PCA-RF and that RF-ensemble provides small further improvements
on both.  Weaknesses of their
methods include not applying to regression or multi-output problems,\footnote{
	Though it is not suggested in the paper, PCA-RF 
	could in theory be applied to regression and multiple output problems.
	LDA-RF and RF-ensemble, on the other hand, do not generalize beyond single output classification.}
a combinatorial computational complexity in number of categories for categorical features, and numerical instability
in the LDA calculation \citep{de2012least}.


\subsection{Canonical Correlation Analysis}
\label{sec:CCA}

Canonical correlation analysis (CCA) \citep{canoncorr} is a deterministic method for calculating pairs of linear projections that maximise the correlation between two matrices in the co-projected space.  It is a co-ordinate free process that is unaffected by rotation, translation or global scaling of the inputs.  Consider applying CCA between the arbitrary matrices $W \in \mathbb{R}^{n \times d}$ and $V \in \mathbb{R}^{n \times k}$.  The first pair of canonical coefficients are given by
\begin{align}
\label{eq:CanoncorrRed}
\left\{A_1, B_1\right\} = \argmax\limits_{a \in \mathbb{R}^d, b \in \mathbb{R}^k} \left( \mathrm{corr} \left(W a , V b \right) \right)
\end{align}
subject to the conditions $\lVert a \rVert_2 = 1$ and $\lVert b \rVert_2 = 1$.  The corresponding canonical correlation components are given by $W A_1$ and $V B_1$.  Another $\nu_{\max}-1$ pairs 
are created where $\nu_{\max}= \min \left(\mathrm{rank}\left(W\right),\mathrm{rank}\left(V\right) \right)$
by repeating the same optimization with the additional constraints that the new components are uncorrelated with all previous components, e.g.:
\begin{align}
\label{eq:canonConstr}
\left(W A_1\right)^T W A_2 = 0 \quad \text{and} \quad \left(V B_1\right)^T V B_2 = 0.
\end{align}
As shown by, for example, \cite{cohen1969computation,borga2001canonical}, the solution of~\eqref{eq:CanoncorrRed} 
 has a simple closed form (see Appendix~\ref{sec:numStabCCASup}).
However, this solution can be numerically unstable as it requires an inversion of typically degenerate covariance matrices.  \cite{bjorck1973numerical} demonstrated that the solution for CCA can also be found in a numerically stable fashion using a combination of QR \citep{householder1958unitary} and SVD \citep{golub1970singular} decompositions. Details of our numerically stable approach 
based on this work are given in Appendix \ref{sec:numStabCCASup}.

A key motivation for using CCA is that it is applies for any
two matrices.  This means that, by using a 1-of-K class encoding for classification problems,
we can use CCA, and therefore CCFs, for regression, classification, 
multi-output classification, and multivariate regression, 
all of which will be considered later in the paper.
We note that for the case of single-output classification, then the feature projection matrix produced by 
our CCA approach is analytically equivalent to that produced by multi-class
LDA~\citep{hastie1995penalized,de2012least}.  For a single-output regression, 
CCA is instead equivalent to projecting onto the hyperplane produced by
the output of a linear regression~\citep{sun2008least}.
Along with unifying these cases and generalizing to multiple outputs, there are
also numerical advantages to using CCA.  
For example, the CCA approach given in Appendix \ref{sec:numStabCCASup}
provides more numerical stability and easier regularization than standard approaches for
calculating LDA.  This is particularly important given that the analysis is generally
done on a bootstrap sample of the original data, such that the problem is highly likely to be degenerate.

\section{Canonical Correlation Forests}
\label{sec:CCFs}

%

\subsection{Overview}
\label{sec:ccf:overview}

We start by providing a high level overview of CCFs, before introducing formal notation
and an in-depth description later in the section.
As with RFs and most other decision tree ensemble methods, the trees in a CCF
are independently trained in a greedy, self-similar, top down procedure that successively
chooses the best split (according to some split criterion)
by exhaustively searching over a set of candidates.  The algorithm starts
with a root node containing all the training data and each time a new split is selected, this
creates two new child nodes with each data points passed down to either the left or right
child depending on what side of the split the data point falls.  The same procedure for choosing
the best split is then applied to each of the child nodes and this process continues recursively
until all nodes have met a stopping criterion or contain only a single unique data point, at which
point they become leaf nodes.

\begin{figure}[p]
	\centering
	\adjincludegraphics[width=0.95\textwidth,trim={0 0 0 0},clip]{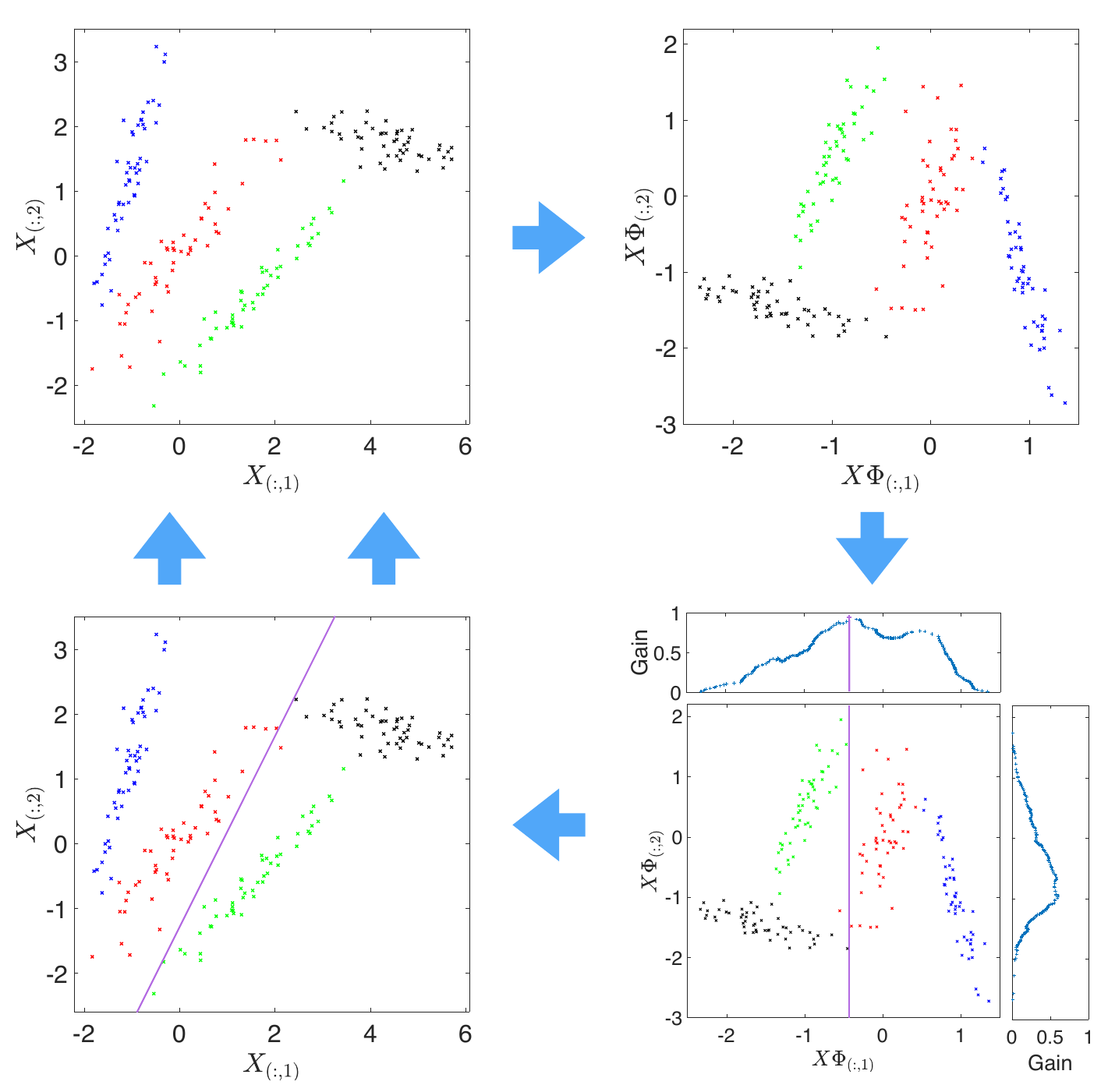}	
	\caption{High level demonstration of CCT training process for a classification task.  
		Given the original training data at the node, $\lambda$ ($=2$ in this example) features are subsampled 
		for consideration (top left). 
		CCA with projection bootstrapping  (see Section~\ref{sec:projBoot})
		is carried out to project the data into a space that maximally correlates the inputs with the outputs
		(top right).  Note how knowing only one of the features in this projected space is significantly
		more informative about the class identity than knowing only one of the features in the original space.
		For example, knowing the horizontal axis value in the projected space is enough to uniquely
		identify whether a point is in the blue class or not.
		The next step is to calculate a split criterion (e.g. information gain) for each of the possible unqiue 
		splits (splits are always taken halfway between adjacent points) in this 
		projected space (bottom right).  The split with the highest gain is selected for
		splitting, implying a hyperplane split in the original space (bottom left).  
		The self-similar process is then repeated separately on each of the two newly created partitions, 
		terminating when each partition has no further advantageous split (e.g. because it only contains
		data points of a particular class). The tree is complete (see Figure \ref{fig:expTrees}) 
		when all branches have reached termination.
		\label{fig:exp}}
\end{figure}

\begin{figure}[t]
	\centering
	\adjincludegraphics[width=0.48\textwidth,trim={0 0 0 0},clip]{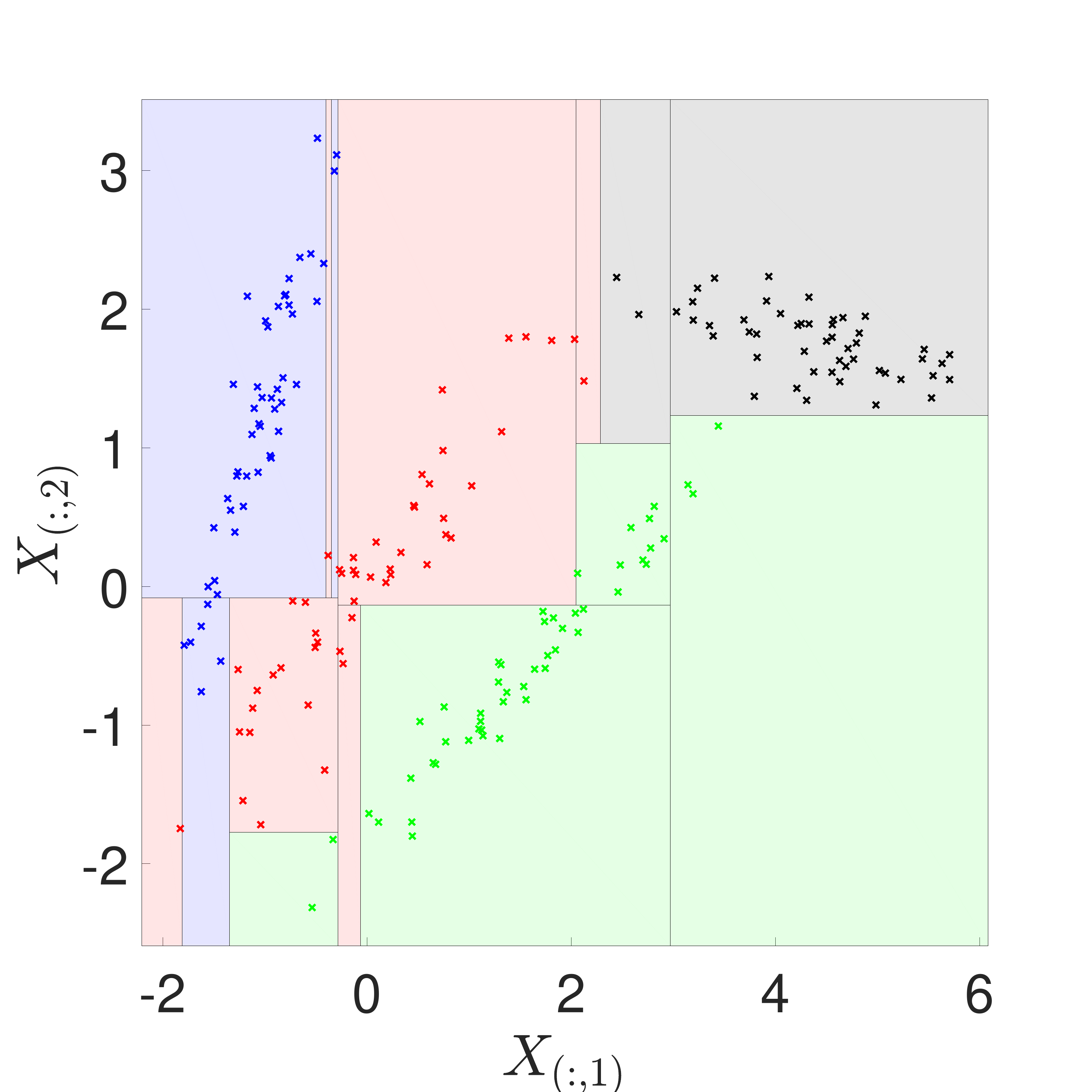}	
	~~~
	\adjincludegraphics[width=0.48\textwidth,trim={0 0 0 0},clip]{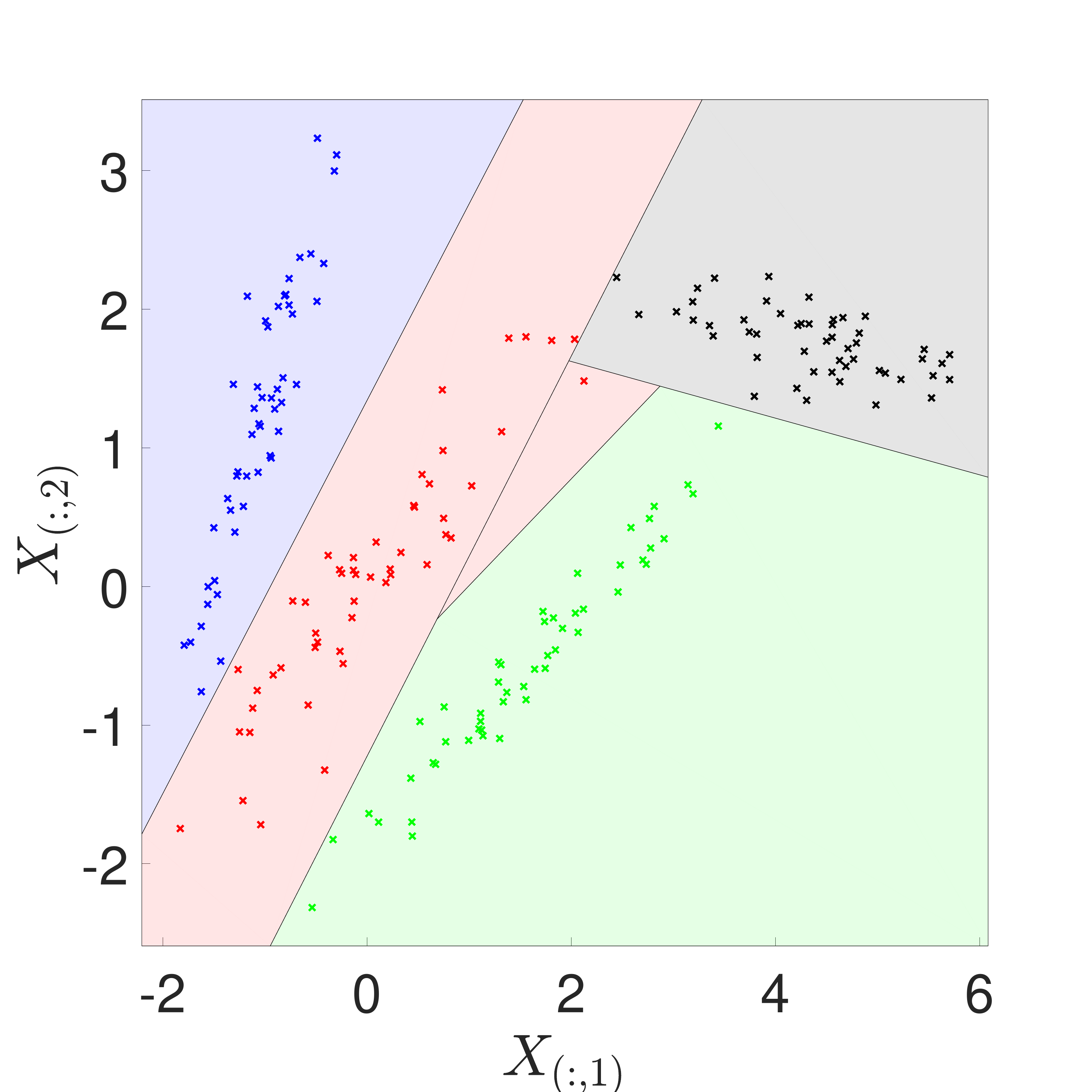}	
	\caption{Tree resulting from the axis aligned decision tree algorithm CART \citep{CART} (left) and CCT training (right) on the dataset from Figure \ref{fig:exp}.
		\label{fig:expTrees}}
\end{figure}

There are two key differences between the training algorithms for CCFs and RFs.  
Firstly, for RF training each tree is trained on a
bootstrap sample of the data in a process known as bagging, while for CCFs, each tree is
trained on the full training dataset.  Secondly, CCFs consider a different set of split
candidates.  In RF training then a random subset of the features are considered at each node and the
set of split candidates corresponds to all unique axis aligned partitions of the data using these
features.  In CCF training a random subset of the features is also taken, but CCA 
with projection bootstrapping (see Section~\ref{sec:projBoot}) is 
used first to project
the features into canonical component space, with the set of split candidates corresponding to the
unique partitions in this projected space.  The chosen partition then implies a hyperplane
split that can be used directly at test time.
This training process is summarized in Figure~\ref{fig:exp}, using an example
classification problem. The resulting CCT along with an axis aligned alternative is shown in
Figure~\ref{fig:expTrees}.

\subsection{Forest Definition and Prediction}
\label{sec:Notation}

Though CCFs are able to deal with categorical inputs, missing data, and multiple outputs 
(see Sections~\ref{sec:DataPrepro} and \ref{sec:multi-out}), we neglect these cases for now in 
the interest of clarity.
Our aim is to make predictions $v \in \mathbb{V}$ about outputs $y \in \mathbb{Y}$, given corresponding vectors of input features $x \in \mathbb{R}^D$.
For regression, the outputs $y$ are single numeric values $y \in \mathbb{R}$.
For classification, the outputs are class labels $k \in \{1, \dots, K\}$, but
for notational convenience we represent these using a 1-of-K encoding 
$y \in \mathbb{I}^{K}$, where for class $k$ the $k^{\mathrm{th}}$ element of
$y$ is set to $1$ and all other values are set to $0$.  Though the predictions
convey information about the outputs, it need not be the case that 
$\mathbb{V}=\mathbb{Y}$.  For example, for classification we consider 
predicting relative class probabilities, $v \in [0,1]^K$, as we are uncertain
about the true class.
For regression we will only consider predicting a point
estimate for the output, i.e. $\mathbb{V}=\mathbb{Y}$, though it is possible
to also calculate uncertainty estimates in the same way as for
RFs (see for example \cite{criminisi2011decision}).  

CCFs represent a supervised learning approach and so require training using labelled input
pairs.  Once trained, prediction can be carried out at arbitrary input points.
We consider using a training dataset comprising of
$N$ inputs $X = \{x_n\} _{n=1}^N$ and corresponding outputs $Y = \{y_n\} _{n=1}^N$.
Here $X$ and $Y$ can be conveniently represented as matrices, with the first index denoting
the data point by convention.
We use the notation $X_{(u,v)}$ for indexing, where $u$ and $v$ can both be
either scalars or vectors of indexes and $:$ indicates all instances in that dimension.
Thus, for example, $Y_{(1,:)}$ corresponds to $y_1$ and $X_{([1,4],[2,3])}$ denotes a matrix
containing the second and third features of the first and fourth data points.
Vectors are indexed in the same way. For example $x_{([1,3])}$ represents the first and
third dimensions (i.e. features) of $x$.

Let $T = \left\{t_i \right\}_{i = 1 \dots L}$ denote a CCF, comprising of $L$ individual canonical correlation trees (CCTs) $t_i$.   When clear from the context, we will omit the tree index $i$ to avoid clutter.
As with classical decision trees, CCTs define a hierarchical partitioning on the input space.  Each individual tree $t_i = \left\{\Psi, \Theta \right\}$ is defined by a set of discriminant nodes $\Psi = \left\{\psi_j\right\}_{j \in \mathcal{J} \backslash \partial \mathcal{J}}$ and a set of leaf nodes $\Theta = \left\{\theta_j\right\}_{j \in \partial \mathcal{J}}$ where $\mathcal{J} \subset \mathbb{Z}^{\geq 0}$ is the set of node indices and $\partial \mathcal{J} \subseteq \mathcal{J}$ is the subset of leaf node indices.  Each discriminant node is defined by the tuple $\psi_j = \left\{j, \delta_j, \phi_j, s_j, \chi_{j,\ell}, \chi_{j,r} \right\}$ where $j$ is the unique node identifier,
$\delta_j \in \{1,\dots,D\}^{\lambda}$ is a vector of indexes giving the features used for splitting at the node, $\phi_j \in \mathbb{R}^{\lambda}$ is a weight vector used to project these features, $s_j \in \mathbb{R}$ is the point at which the split occurs in the projected space $x^T_{(\delta_j)} \phi_j$, and $\left\{\chi_{j,\ell}, \chi_{j,r}\right\} \subseteq \mathcal{J} \backslash j$ are the two child node ids.  

Let $B \left(j, t\right)$ denote the partition of the input space associated with node $j$ of tree $t$ such that $B \left(0, t\right) = \mathbb{R}^D$ and $B\left(j, t\right) = B\left(\chi_{j,\ell}, t\right) \cup B\left(\chi_{j,r}, t\right)$.  The partitioning procedure is then defined such that 
\begin{align}
\label{eq:partitioning}
\begin{split}
B \left(\chi_{j,\ell}, t\right) & = B\left(j, t\right) \cap \left\{x \in \mathbb{R}^D : x^T_{(\delta_j)} \phi_j\le s_{j} \right\} \\
B \left(\chi_{j,r}, t\right) & = B\left(j, t\right) \cap \left\{x \in \mathbb{R}^D : x^T_{(\delta_j)} \phi_j> s_{j} \right\}.
\end{split}
\end{align}
Thus $\Psi$ defines a hierarchical partitioning procedure that deterministically 
assigns inputs to leaf nodes.  We further introduce the notation $\omega_j$ to
indicate the indices of the training points which fall into the partition of node $j$,
i.e.
\begin{align}
\label{eq:omega}
\omega_j = \left\{n \in 1\dots N \colon x_n \in B(j,t)\right\},
\end{align}
and $N_j = \lVert\omega_j\rVert_0$ as the corresponding number of training
points at the node.

Each leaf node is itself defined by a local prediction
model $\theta_j : \mathbb{R}^D \rightarrow \mathbb{V}$.\footnote{Technically 
	speaking, the prediction space of individual trees need not be same 
	as that of forest, see for example \cite{criminisi2011decision}.  
	However, we omit consideration of this scenario for clarity.}
Although more complicated leaf 
models are possible, using for example logistic regression \citep{gama2004functional} 
or kernels \citep{geurts2006kernelizing}, in this paper we only consider the case 
where the leaf model is a deterministic assignment to the average
of the outputs reaching that leaf in the training set. Specifically
\begin{align}
\label{eq:theta}
\theta_j (x)= \frac{1}{N_j}\sum_{n \in \omega_j} y_n,
\end{align}
such that the tree's prediction is independent of the input, given
the leaf assignment.  
We therefore drop the dependence on $x$,
such that each leaf node is fully defined by its identifier $j$ and prediction $\theta_j \in \mathbb{V}$.

As with RFs, out of sample prediction in CCFs is done using an equally weighted voting 
scheme of the tree predictions. Slightly abusing notation, let $t_i \left(x\right) \in \mathbb{V}$ 
denote the prediction of $t_i$ for input $x$.  The forest's prediction is then given by
\begin{align}
\label{eq:voting}
T(x) = \frac{1}{L} \sum_{i=1}^{L} t_i \left(x\right).
\end{align}
We note that more complicated methods of combining the tree predictions could
be used instead, see, for example, \cite{robnik2004improving} and 
\citep{geurts2006kernelizing}.


\begin{figure}[p]
	\vspace{-0.2em}
	\rule{\columnwidth}{1.3pt} 	
	\vspace{-1.8em}
	\begin{algorithm}[H]
		\small
		\captionsetup{labelfont=bf, justification=justified,singlelinecheck=false}
		\caption{CCF training algorithm \label{alg:CCFtrain}}
		\vspace{-0.5em}
		\rule{\columnwidth}{0.5pt}
		\begin{algorithmic}[1]
			\renewcommand{\algorithmicrequire}{\textbf{Inputs:}}
			\renewcommand{\algorithmicensure}{\textbf{Outputs:}}				 
			\Require features $X \in \mathbb{R}^{N \times D}$, outputs $Y \in \mathbb{Y}^N$, 
			number of trees $L \in \mathbb{Z}^+$ (default is $500$),
			number of features to sub-sample $\lambda \in \left\{1,\dots,D\right\}$ (default is  
			$\ceil*{\log_2 (D)+1}$),
			impurity measure $g : \mathbb{Y}^{\mathbb{Z}^+} \rightarrow \mathbb{R}$ (default is~\eqref{eq:entropy} for classification
			and~\eqref{eq:mse-gain} for regression), stopping criteria $c$ (see Section~\ref{sec:algorithm})
			\Ensure CCF $T$
			\State Preprocess $X$ \Comment See Section~\ref{sec:param}
			\For{$i = 1 \colon L$}
			\State Randomly assign missing values in $X$ \Comment See Section~\ref{sec:param}
			\State $ t_i \leftarrow \Call{growTree}{0, X, Y, \lambda, g, c}$ \Comment Each tree trained independently
			\EndFor	
			\State \Return $T = \left\{t_i \right\}_{i = 1 \dots L}$
		\end{algorithmic}
		\vspace{-0.5em}
		\rule{\columnwidth}{0.5pt}
		\vspace{-2em}
	\end{algorithm}
	\rule{\columnwidth}{1.3pt} 	
	\vspace{-1.8em}	
	\begin{algorithm}[H]
		\small
		\captionsetup{labelfont=bf, justification=justified,singlelinecheck=false}
		\caption{\textsc{GrowTree} \label{alg:growTree}}
		\vspace{-0.5em}
		\rule{\columnwidth}{0.5pt}
		\begin{algorithmic}[1]	
			\renewcommand{\algorithmicrequire}{\textbf{Inputs:}}
			\renewcommand{\algorithmicensure}{\textbf{Outputs:}}
			\Require unique node identifier for root node $j$, $X^j = X_{(\omega_j,:)} \in \mathbb{R}^{N_j \times D}$, 
			$Y^j = Y_{(\omega_j,:)} \in \mathbb{Y}^{N_j}$, $\lambda$, $g$, $c$
			\Ensure subtree $\{\Psi_j,\Theta_j\}$ where $\Psi_j$ are discriminant nodes and $\Theta_j$ leaf nodes
			\State Subsample features ids $\delta_j$ by sampling from $\left\{1,\dots,D\right\}$ $\lambda$ times without replacement.  \label{growTree:line:subspace1}
			\State Set $\mathcal{X} \leftarrow X^j_{(:,\delta_j)}$ \label{growTree:line:subspace2}
			\State Construct a bootstrap sample $\left\{\mathcal{X}', \mathcal{Y}'\right\}$ by sampling $N_j$ data points with replacement from $\left\{\mathcal{X}, Y^j\right\}$ \label{growTree:line:boot}
			\State $\left\{\Phi, \Omega\right\} \leftarrow \Call{CCA}{\mathcal{X}', \mathcal{Y}'}$ \Comment Calculate CCA coefficients using bootstrap sample \label{growTree:line:cca}
			\State $U \leftarrow \mathcal{X} \Phi$ \Comment Project \textit{original features} into canonical component space \label{growTree:line:project}
			\State $G_{\mathrm{base}} \leftarrow g\left(Y^j \right)$ \Comment Impurity of current node
			\State $\nu_{\max}= \min \left(\mathrm{rank}\left(\mathcal{X}'\right),\mathrm{rank}\left(\mathcal{Y}'\right) \right)$
			\Comment  Number of projections
			\For{$\nu\in1:\nu_{\max}$} \Comment Evaluate splits for each projection
			\State $u \leftarrow \Call{Sort}{U_{(:,\nu)}}$ \label{growTree:line:u}
			\For{$i\in2:N_j$} \Comment Exhaustive search on unique splits
			\State $S_{(i,\nu)} \leftarrow (u_{(i-1)}+u_{(i)})/2$ \label{growTree:line:S}
			\Comment Split halfway between consecutive points
			\State $\tau^{\ell} \leftarrow \left\{n \in \left\{1,\dots,N_j\right\} : U_{(n,\nu)} \le S_{(i,\nu)}\right\}$
			\Comment Points that would be assigned to left child
			\State $N_{\chi_{j,\ell}}\leftarrow$ number of elements in $\tau^{\ell}$
			\State $\tau^{r} \leftarrow \left\{1,\dots,N_j\right\} \backslash \tau^{\ell}$
			\State $G_{(i,\nu)} \leftarrow G_{\mathrm{base}}
			-\frac{N_{\chi_{j,\ell}}}{N_j} g\left(Y_{(\tau^{\ell} ,:)}\right)
			-\frac{N_j-N_{\chi_{j,\ell}}}{N_j} g\left(Y_{(\tau^{r} ,:)}\right)$ 
			\label{growTree:line:gain}
			\Comment Split gain
			\EndFor
			\EndFor
			\State $\left\{i^*, \nu^*\right\} \leftarrow \argmax_{i,\nu} G_{(i,\nu)}$ 
			\label{growTree:line:optgain}
			\Comment Choose best split
			\If{$G_{(i^*, \nu^*)} \le 0$ (i.e. no split is beneficial) or any stopping of criteria $c$ satisfied} 
			\label{growTree:line:bLeaf}
			\Comment Node is a leaf
			\State $\theta_j \leftarrow \frac{1}{N_j} \sum_{n=1}^{N_j} Y^j_{(n,:)}$ \Comment Predictive model is average of points at leaf
			\State \Return $\left\{\cdot,\left\{j,\theta_j\right\}\right\}$
			\Else \Comment Node is a discriminant node
			\State Generate unique identifiers for children $\chi_{j,\ell}$ and $\chi_{j,r}$
			\State $\phi_j \leftarrow \Phi_{(:,\nu^*)}, \quad s_j \leftarrow S_{(i^*, \nu^*)}$ \Comment Chosen projection and split point
			\State $\psi_j \leftarrow \left\{j, \delta_j,  \phi_j, s_j, \chi_{j,\ell}, \chi_{j,r} \right\}$ 
			\State $\tau^{\ell} \leftarrow \left\{n \in \left\{1,\dots,N_j\right\} : U_{(n,\nu^*)} \le S_{(i^*,\nu^*)}\right\}$
			\label{growTree:line:child1}
			\Comment Assign data points to children
			\State $\tau^{r} \leftarrow \left\{1,\dots,N_j\right\} \backslash \tau^{\ell}$
			\label{growTree:line:child2}
			\State $\left\{\Psi_{\ell},\Theta_{\ell}\right\} \leftarrow 
			\Call{growTree}{\chi_{j,\ell}, X^j_{(\tau^{\ell},:)}, Y^j_{(\tau^{\ell},:)}, \lambda, g, c}$
			\Comment Recurse for left child and right child
			\State $\left\{\Psi_{r},\Theta_{r}\right\} \leftarrow 
			\Call{growTree}{\chi_{j,r}, X^j_{(\tau^{r},:)}, Y^j_{(\tau^{r},:)}, \lambda, g, c}$
			\label{growTree:line:lastrecurse}
			\State \Return $\left\{\psi_j \cup \Psi_{\ell} \cup \Psi_r,  \Theta_{\ell} \cup \Theta_{r}\right\}$
			\EndIf
		\end{algorithmic}
		\vspace{-0.5em}
		\rule{\columnwidth}{0.5pt}
		\vspace{-2em}
	\end{algorithm} 
\end{figure}

\subsection{Forest Training}
\label{sec:algorithm}

Algorithms~\ref{alg:CCFtrain} and~\ref{alg:growTree} give a step by step walk-through for training a CCF, outlining
the forest training and tree growing processes respectively.
In both cases some simplifications have been made for clarity,
with more comprehensive algorithm blocks provided in Appendix~\ref{sec:detail-alg}, additionally detailing things
such as categorical features and missing data.  As we can see from Algorithm~\ref{alg:CCFtrain}, CCF training
requires as inputs the data $\{X,Y\}$, a number number of trees $L\in \mathbb{Z}^+$, a number of features
to sub-sample $\lambda\in \left\{1,\dots,D\right\}$, an impurity measure $g : \mathbb{Y}^{\mathbb{Z}^+} \rightarrow \mathbb{R}$
(see~\eqref{eq:entropy} and~\eqref{eq:mse-gain}), and stopping criteria $c$ (see Section~\ref{sec:param}).  
Other than the data, all of these required inputs have default values, meaning that
CCFs can be trained without parameter tuning, as is done in all of our experiments.  Setting these
options will be discussed in~\ref{sec:param}, for now we just note that all these options are shared with RFs and
therefore the intuition for how they should be set predominantly transfers.  We therefore also refer the reader
to previous work that examines the effect these parameters on decision tree ensembles more generally, 
e.g. \cite{bernard2009influence,criminisi2011decision}.

As should hopefully be clear by now, each tree in a CCF is trained independently using the full dataset.
The tree training process, \textsc{GrowTree}, is self similar, starting at the root node
with the full dataset.  At each call it selects an optimal split for a set of generated candidates
(lines~\ref{growTree:line:subspace1} to~\ref{growTree:line:optgain}); decides whether
to assign the current node as a leaf or discriminant node (line~\ref{growTree:line:bLeaf}); 
and then if the node is assigned as a decision 
node, it recursively calls \textsc{GrowTree} again (with appropriate partitions of the data) to 
produce sub-trees for each of the newly generated child nodes (lines~\ref{growTree:line:child1}
to~\ref{growTree:line:lastrecurse}).  
The training process is complete once all generated branches have
terminated as leaf nodes.

The process for deciding whether to assign a node as a leaf node or a decision node is straightforward - 
the node is assigned to be a leaf if no split is beneficial (as per the split gain discussed later) 
or if a stopping criterion is met (see Section~\ref{sec:param}). Otherwise it becomes a decision node.  
The key part of the algorithm is therefore the process for selecting the split, namely selecting the
features used for splitting $\delta_j$, the projection vector $\phi_j$, and the split 
point $s_j$.  This can be further
broken down into generating a set of candidate splits and selecting the optimal split from this candidate
set.  

As in RFs, the process of generating the set of candidate splits for CCFs
starts by randomly
sampling, $\delta_j$, the subset of the features to consider at that node (lines~\ref{growTree:line:subspace1} 
and~\ref{growTree:line:subspace2} of Algorithm~\ref{alg:growTree}).  This corresponds to
sampling $\lambda$ features without replacement from all of those present.  For RFs, 
the set of candidate splits comprises of all unique axis-aligned splits using these features.
For CCFs the candidate splits will be hyperplane splits implied by the possible axis-aligned splits
in a projected space $X \Phi$.  To calculate $\Phi$, we first take a bootstrap sample of the local data at the
node $\{\mathcal{X}',\mathcal{Y}'\}$ (line~\ref{growTree:line:boot} of Algorithm~\ref{alg:growTree}).  Using this bootstrap sample
we then carry out a CCA between the features and the outputs (line~\ref{growTree:line:cca} of Algorithm~\ref{alg:growTree})
\begin{align}
\label{eq:canonFeat}
\left[\Phi, \Omega \right] = \mathrm{CCA} \left(\mathcal{X}', \mathcal{Y}'\right)
\end{align}
where $\Phi$ and $\Omega$ are the canonical coefficients corresponding to $\mathcal{X}'$ and $\mathcal{Y}'$ respectively.  We note that $\Omega$ is not directly used by the algorithm.
Our set of candidate splits is now the unique axis-aligned splits in the projection of the \emph{original} 
input features into the canonical component space, namely (line~\ref{growTree:line:cca} of Algorithm~\ref{alg:growTree})
\begin{align}
\label{eq:U}
U = X_{(\omega_j,\delta_j)} \Phi
\end{align}
remembering that $\omega_j$ is the index of data points present at node $j$.
Choosing a split in the space of $U$
implies a hyperplane split in the space of $X$, given by the projection $\phi_j$, corresponding to
one of the columns of $\Phi$, and the split point $s_j$ in the projected space.  
Note that CCA is only required during the training phase with the splitting 
rule \eqref{eq:partitioning} used directly for out of sample prediction.  

The process of choosing the best split from the set of possible candidates
 is analogous to the exhaustive search approach 
used by most decision tree methods (including RFs). The only difference is that the search
uses $U$ instead of $X_{(\omega_j,\delta_j)}$.  The core idea is to use an impurity
criterion $g : \mathbb{Y}^{\mathbb{Z}^+} \rightarrow \mathbb{R}$
and to choose the split which most reduces the impurity
(lines~\ref{growTree:line:gain} and~\ref{growTree:line:optgain} of Algorithm~\ref{alg:growTree}).
Namely we wish to solve
\begin{align}
\left\{\phi_j,s_j\right\} &= \argmax_{\phi \in \Phi, s \in \mathbb{R}} \; 
G\left(Y_{(\omega_j,:)},\phi,s\right) \nonumber \\
&= \argmax_{\phi \in \Phi, s \in \mathbb{R}} \; \left(g\left(Y_{(\omega_j,:)}\right)
-\frac{N_{\chi_{j,\ell}}}{N_j} g\left(Y_{(\omega_{\chi_{j,\ell}},:)}\right)
-\frac{N_{\chi_{j,r}}}{N_j} g\left(Y_{(\omega_{\chi_{j,r}} ,:)}\right)\right)
\label{eq:gain}
\end{align}
where $G\left(Y_{(\omega_j,:)},\phi,s\right)$ is commonly referred to as the gain of a split and
using~\eqref{eq:partitioning},~\eqref{eq:omega}, and~\eqref{eq:U} we have that
\begin{align}
\omega_{\chi_{j,\ell}} \left(\phi,s\right) &= \left\{n \in \omega_j \colon X_{(n,\delta_j)} \phi \le s\right\} \\
\omega_{\chi_{j,r}} \left(\phi,s\right) &= \left\{n \in \omega_j \colon X_{(n,\delta_j)} \phi > s\right\} 
= \omega_j \backslash \omega_{\chi_{j,\ell}} \left(\phi,s\right).
\end{align}
We see that the gain of a split depends only on how the data points are assigned to each of the children,
such that any $\{\phi,s\}$ that leads to the same partitioning of the training data gives the same gain.
By construction we already have that there are only a finite number of $\phi$ considered and so the set
of unique possible splits is now fully defined by the associated values of $s$ that lead to distinct
partitions of the training data.  We define this by restricting valid splits to be halfway 
between consecutive points in the respective sorted column of $U$ (see lines~\ref{growTree:line:u}
and~\ref{growTree:line:S} of Algorithm~\ref{alg:growTree}).  Consequently there
are $(N_j-1) \nu_{\max}$ possible splits where $\nu_{\max}$ is the number of
columns of $\Phi$ (see Section~\ref{sec:CCA}).  

The choice of impurity criterion depends on whether classification or regression is being performed.
The intuitive property we desire for the impurity criterion is that it is high when the uncertainty in the
output is high or equivalently when homogeneity in the outputs for points at the node is low.
For classification we take as a default the entropy of the node defined as
\begin{align}
\label{eq:entropy}
g\left(Y_{(\omega_j,:)}\right) = -\sum_{k=1}^{K} p_k \log_2 p_k
\end{align}
where $p_k$ is the empirical probability of class $k$ at the node, namely
$p_k = \frac{1}{N_j} \sum_{n\in\omega_j} Y_{(n,k)}$.
The entropy impurity criterion is thus the entropy of the predictive distribution that would be
generated if the node were to become a leaf.  Using the entropy as the impurity metric corresponds
to using information gain as the split criterion, as introduced by~\cite{quinlan1986induction}.
A common alternative for classification is the
Gini impurity $g\left(Y_{(\omega_j,:)}\right) = 1-\sum_{k=1}^{K} p_k^2$ \citep{CART}.  Entropy is taken
as a default due to slightly superior performance in our experiments, but we note that performance
variations were small and the overall improvement was certainly not definitive.

For regression we use the empirical variance,
\begin{align}
\label{eq:mse-gain}
g\left(Y_{(\omega_j)}\right) = \frac{1}{N_j} \sum_{n \in \omega_j} Y_{(n)}^2 - \left(\frac{1}{N_j} \sum_{n \in \omega_j} Y_{(n)}\right)^2,
\end{align}
as an impurity measure~\citep{CART}.  Using this impurity measure
is often referred to as the mean squared error split criterion as it corresponds to the mean
square error that would result from using the mean of the samples (which is the predictive
model taken if the node is a leaf) as the prediction for the training data at that node.
Alternative split criteria, such those based on curvature~\citep{loh2002regression}, could also
be used, but we will not consider these further here.

Given a split and the assignment of the node to be a decision node, all that remains is to
recurse to the children.  This simply involves assigning all datapoints to either the left
or right child using~\eqref{eq:partitioning} (lines~\ref{growTree:line:child1} and~\ref{growTree:line:child2}
in Algorithm~\ref{alg:growTree}) and calling \textsc{GrowTree} for each child using the corresponding partition
of the training data.

\subsubsection{Projection Bootstrapping}
\label{sec:projBoot}

We refer to calculating the projection matrix $\Phi$ using a local bootstrap sampling of the data but
then using the original dataset for choosing the split in the projected space as
\emph{projection bootstrapping}.  The motivation for projection bootstrapping, instead of
say using bagging, is that it retains all the information from the dataset when choosing
$\{\phi_j,s_j\}$ from the candidates, thereby improving the predictive accuracy of individual trees.
This is beneficial because the use of hyperplane splits increases the diversity of the
ensemble compared with axis-aligned forests. Therefore less artificial randomness needs
to be added to the tree training process to sufficiently decorrelate the tree predictions.
Consequently, as shown in Section~\ref{sec:exp}, projection bootstrapping leads to
improved ensemble predictive performance compared to bagging.
We found, on the other hand, that omitting both bagging and projection bootstrapping, and thus
relying on random subspacing alone to decorrelate tree predictions, reduced the randomness in
the tree training process too far, leading to degraded performance.

An important exception to using projection bootstrapping is in the case where $\lambda = D$,
i.e. when we consider all of the features at each node, for which we use bagging instead.
The reason for this is that no
feature subspacing occurs in this scenario and so additional randomness needs to be added into
the training process to prevent overfitting.
Using the default settings for $\lambda$, this scenario occurs only when $D\le2$.  

\subsubsection{Data Preprocessing}
\label{sec:DataPrepro}
The format of our forest definition given in Section \ref{sec:Notation} requires $X$ to be in numerical form.  
This is not a problem for ordered categorical features which can simply be treated as numerical features using the class
index.  For unordered categorical features $x^c \in \mathcal{S}$, where $\mathcal{S}$ represents the space of
arbitrary qualitative attributes, we use a 1-of-K encoding such that the feature is converted into $K$ binary
features.  To ensure equal probability of selecting categorical and
numerical features, the expanded binary array of each categorical feature is still treated as a single feature during
the random subspacing (i.e. sampling $\delta_j$, see~Algorithm~\ref{alg:growTreeDet}).  
In the rest of the paper we refer to numeric, binary, and ordered categorical features 
(i.e. those that do not require this expansion) as ordinal and non-ordered categorical features as non-ordinal.

A second preprocessing step is to centre and rescale the input data so that it has zero mean and unit variance
for each dimension (ignoring any missing data).  As CCA and the split criteria are invariant under affine
transformations of the full dataset, this does not have any direct bearing on the tree training.  However it 
ensures equal influence of the input features in the rank reduction used to ensure stability of the CCA
calculation as described in Section~\ref{sec:numStabCCASup}.

The final preprocessing step is to deal with missing data.  Here we take the simple approach of randomly
assigning each missing value in $X$ to a random draw from a unit Gaussian (noting that we have previously
ensure that each feature has zero mean and unit variance).  This is done independently for each tree so that
each tree is affected in a different way. The same approach is used for
both training and testing.  Though we found this simple approach worked well in practise, one could also
envisage using more complicated schemes, such as sampling from the empirical distribution of the feature over
the training data instead.

%


\subsubsection{Stopping Criteria and other Parameter Settings}
\label{sec:param}

As previously stated, the CCF training algorithm has four parameters of note: the number of trees $L$, the
number of features to sub-sample at each node $\lambda$, the impurity measure $g$, and the stopping
criteria $c$.  We reiterate that all of these parameters are common to RFs.
 
 Setting $g$ has already been discussed in Section~\ref{sec:algorithm}, using entropy
and variance as defaults for classification and regression respectively.  The setting of $L$ need only be based on
computational budget as it is commonly accepted~\citep{criminisi2011decision},
and there are theoretical results suggesting \citep{breiman2001random,biau2012analysis}, 
that more trees is always preferable for common 
decision ensemble methods (with the exception of boosting which can overfit if $L$ is too large).  
We take $L=500$ as a default
value as compromise  between speed and accuracy, noting that, as shown in Section~\ref{sec:noTrees}, the misclassification
rate has usually stabilized by this value of $L$.  It may often, however, be beneficial to use significantly smaller values
for $L$ when speed is paramount none-the-less.  From this perspective CCFs offer the potential for significant
speed improvements over RFs, as we found that on average only $15$ CCTs are required to match the accuracy
of $500$ RF trees (see Section~\ref{sec:noTrees}).  When the number of features is very large, it may also be
beneficial to use more than $500$ trees.

The default number of features to sample at each step is taken to be $\lambda = \ceil*{\log_2 \left(D\right) +1}$ where $\ceil*{\cdot}$ represents the
ceiling function,
with the exception that we set $\lambda=2$ when $D=3$ so that random subspacing and CCA can both be employed.
 This is the default setting taken by, for example, the WEKA RF implementation~\citep{hall2009weka}.  It has the
 advantage of being very fast, as the training cost subsequently scales logarithmically with the number of
 features (see Section~\ref{sec:CompComp}). We also found that it was also an effective choice in terms of 
 empirical predictive accuracy,
 outperforming the other common choice of $\lambda = \ceil*{\sqrt{D}}$.  One could of course also tune $\lambda$
 at the expense of training speed, but we note that results in the recent survey of~\cite{fernandez2014we},
 along we our own comparisons to their results, suggest that there is
 no noticeable difference between the performance of RF packages that tune hyperparameters and those that do not.
 We further refer the reader to the study for the effect of
 $\lambda$ on RF performance by \cite{bernard2009influence}.

Though a number of possible stopping criteria exist~\citep{criminisi2011decision}, the main two 
used in practice are a maximum tree depth, beyond which all nodes are assigned as leaves (our default is $\infty$),
and a minimum number of contained datapoints for which a node is allowed to split (our default is $2$
for classification and $6$ for regression).  Though not technically a stopping criterion, a common associated
parameter (omitted from the algorithm blocks) is a minimum number of datapoints allowed for a generated
leaf node (our default is $1$ for classification and $3$ for regression).  As such, only splits that assign at
least this many points to a leaf are considered.  The rational for these stopping criteria is to try and guard
against potential overfitting, particularly when the $L$ is small.  Our default settings are again
in line with the defaults of the WEKA RF implementation~\citep{hall2009weka}. We note that the default settings for
classification correspond to using no stopping criteria (other than assigning the node to be a leaf when no
split is beneficial).  The default settings for regression are slightly more conservative, reflecting a greater apparent
tendency for decision tree ensembles to overfit for regression problems then classification problems.  However, it
may that this apparent tendency is due more to the error metrics that are typically used for assessing the two
(i.e. mean squared error compared to misclassification rate), 
rather than any inherent tendencies of the algorithms themselves.

\section{Classification Experiments}
\label{sec:exp}


\subsection{Comparison to State-of-the-art Tree Ensembles}
\label{sec:treeComparison}

To investigate the predictive performance of CCFs, we ran comparison tests against
RFs and rotation forest (Rot-For) over a broad variety of datasets.  The results show that CCFs significantly outperformed both methods and create a new benchmark in classification accuracy for out-of-the-box decision tree ensembles, despite being a considerably more computationally efficient than rotation forests as discussed in Section \ref{sec:CompComp}.

To better understand the key factors behind these improvements, we also considered a number of variations of the CCF algorithm. 
Firstly, we considered using bagging, as per RF, instead of projection bootstrapping.  
We refer to this algorithm as CCF-Bag.

Secondly, we considered the effect of uniformly sampling the projection matrix, $\Phi$, from the space of valid rotation matrices, instead of using CCA.
Here our notion of a uniform distribution over rotations is based on the Haar measure as discussed by \cite{diaconis1987subgroup} and \cite{blaser2016random}.
We refer to this approach as random rotation forests (RRFs).  
To carry out the required rotation matrix sampling, RRFs use the procedure laid out in \cite{blaser2016random} based on a QR decomposition \citep{householder1958unitary} of independent draws from a unit Gaussian. 
Note that, unlike \cite{blaser2016random}, the random rotation in RRFs is carried out at separately each node, as opposed to using a single rotation for the whole tree.

Finally we considered using RRFs with the addition of bagging, referring to this as algorithm as RRF-Bag.  
RRF-Bag shares similarity with the Forest-RC algorithm of \cite{breiman2001random}.
A key difference is that Forest-RC independently samples elements of the projection matrix uniformly in the range $\{-1,+1\}$, biasing the projected features, $U = \mathcal{X} \Phi$, away from the original axis compared with uniformly sampling rotations.
Forest-RC also allows sparsity by calculating $n_p$ separate projections, each with a maximum of $n_{v}$ non-zero values, such that $\Phi$ is constructed as a $D\times n_p$ matrix where there are only $n_v$ values are nonzero in each column.  We do not consider Forest-RC directly here, but note that previous results suggest that its
average performance is similar to RFs.

{\sc{Matlab}}'s \texttt{TreeBagger} function was used for the RF implementation, while our own {\sc MATLAB} implementation\footnote{\scriptsize \url{https://github.com/twgr/ccfs/}} was used for CCFs, CCF-Bag, RRFs, and RRF-Bag.  
As per the default parameters given in~\ref{sec:param}, each ensemble was built using $L=500$ trees, 
$\lambda = \ceil*{\log_2 \left(D\right) +1}$ (with the exception $\lambda=2$ when $D=3$), 
the entropy impurity criterion, and no stopping criteria other than stopping when no split is beneficial.
The choice of the entropy impurity criterion over the Gini impurity criterion
and setting $\lambda = \ceil*{\log_2 \left(D\right) +1}$
rather than the common alternative $\lambda = \ceil*{\sqrt{D}}$ was made on the basis that both choices 
gave improved
predictive performance for RFs on average compared with these alternatives.
For CCFs, CCF-Bag, RRFs and RRF-Bag, missing values were sampled randomly for each tree as explain 
in~\ref{sec:DataPrepro} while for RFs they were dealt with internally by the \texttt{TreeBagger} function.

Rotation Forests were implemented in WEKA \citep{hall2009weka} with 500 trees and the default tree options except that we used binary, unpruned, trees and set the minimum number of instances per leaf to 1.  In addition to keeping the implementation of rotation forest as consistent as possible with the other algorithms, these settings dominated rotation forests of the same size with the default options over a single cross validation.  As recommended by Rodriguez et al \citep{rodriguez2006rotation}, 1-of-K encoding was used for non-ordinal features for rotation forests (note rotation forests do not then treat them differently to ordinal variables), while missing values were replaced by the mean.

\begin{table*}[t]                        
	\centering    
	\footnotesize             
	\setlength\tabcolsep{4pt}	  
	\renewcommand{\arraystretch}{1.2}    
	\caption{Dataset summaries for classification experiments.  $K=$ number of classes, $N=$ number of data points, $D_c=$ number of non-ordinal features, $D_r=$ number of ordinal features, and \% Miss = percentage of feature values which are missing. \label{table:summary}}                            
	\begin{tabular}{ccccccHHHHHHHHHHH}                    
		\hline                                                       
		Dataset & $K$ & $N$ & $D_c$ & $D_r$ & \% Miss & CCF & RF & & Rot-For & & CCF-Bag & & RRF & & RRF-Bag & \\
		\hline           
		Balance scale & 3 & 625 & 0 & 4 & 0 & 9.60 $\pm$ 3.73 & 18.66 $\pm$ 4.56 & $\bullet$ & \textbf{7.25 $\pm$ 3.38} & $\circ$ & 8.80 $\pm$ 3.82 & $\circ$ & 13.53 $\pm$ 3.85 & $\bullet$ & 10.85 $\pm$ 3.89 & $\bullet$ \\            
		Banknote & 2 & 1372 & 0 & 4 & 0 & \textbf{0.00 $\pm$ 0.00} & 0.58 $\pm$ 0.64 & $\bullet$ & \textbf{0.00 $\pm$ 0.00} &  & \textbf{0.00 $\pm$ 0.00} &  & \textbf{0.00 $\pm$ 0.00} &  & 0.02 $\pm$ 0.12 &  \\                        
		Breast tissue & 6 & 106 & 0 & 9 & 0 & 28.18 $\pm$ 12.00 & 31.09 $\pm$ 12.38 & $\bullet$ & 28.48 $\pm$ 12.36 &  & 28.36 $\pm$ 11.82 &  & \textbf{26.06 $\pm$ 11.85} & $\circ$ & 26.24 $\pm$ 12.08 & $\circ$ \\                     
		Climate crashes & 2 & 360 & 0 & 18 & 0 & \textbf{5.81 $\pm$ 4.10} & 6.46 $\pm$ 4.10 & $\bullet$ & 6.11 $\pm$ 3.94 & $\bullet$ & 6.56 $\pm$ 4.16 & $\bullet$ & 7.06 $\pm$ 4.15 & $\bullet$ & 7.22 $\pm$ 4.15 & $\bullet$ \\        
		Fertility & 2 & 100 & 0 & 9 & 0 & 13.33 $\pm$ 9.67 & 14.53 $\pm$ 9.45 & $\bullet$ & \textbf{12.40 $\pm$ 8.87} &  & 13.53 $\pm$ 9.70 &  & 12.67 $\pm$ 9.10 & $\circ$ & 12.87 $\pm$ 9.22 &  \\                                      
		Heart-SPECT & 2 & 267 & 0 & 22 & 0 & 18.59 $\pm$ 6.97 & 18.86 $\pm$ 7.25 &  & \textbf{17.56 $\pm$ 7.29} & $\circ$ & 18.12 $\pm$ 7.12 &  & 18.17 $\pm$ 6.88 &  & 18.37 $\pm$ 7.06 &  \\                                            
		Heart-SPECTF & 2 & 267 & 0 & 44 & 0 & 18.64 $\pm$ 7.36 & 18.91 $\pm$ 7.09 &  & 18.57 $\pm$ 7.06 &  & \textbf{18.25 $\pm$ 7.10} &  & 19.26 $\pm$ 7.63 &  & 19.56 $\pm$ 7.73 & $\bullet$ \\                                         
		Hill valley & 2 & 1212 & 0 & 100 & 0 & \textbf{0.00 $\pm$ 0.00} & 39.16 $\pm$ 4.05 & $\bullet$ & 6.32 $\pm$ 2.65 & $\bullet$ & \textbf{0.00 $\pm$ 0.00} &  & 4.17 $\pm$ 1.86 & $\bullet$ & 6.65 $\pm$ 2.28 & $\bullet$ \\         
		Hill valley noisy & 2 & 1212 & 0 & 100 & 0 & \textbf{4.73 $\pm$ 1.90} & 42.08 $\pm$ 4.57 & $\bullet$ & 11.01 $\pm$ 2.80 & $\bullet$ & 5.24 $\pm$ 1.84 & $\bullet$ & 15.78 $\pm$ 3.44 & $\bullet$ & 18.64 $\pm$ 3.88 & $\bullet$ \\
		ILPD & 2 & 640 & 0 & 10 & 0 & 27.56 $\pm$ 5.08 & 29.92 $\pm$ 5.17 & $\bullet$ & 29.02 $\pm$ 5.25 & $\bullet$ & 28.20 $\pm$ 5.25 & $\bullet$ & 27.55 $\pm$ 4.94 &  & \textbf{27.29 $\pm$ 4.98} &  \\                               
		Ionosphere & 2 & 351 & 0 & 33 & 0 & 4.82 $\pm$ 3.44 & 6.53 $\pm$ 3.95 & $\bullet$ & 5.79 $\pm$ 3.56 & $\bullet$ & 5.50 $\pm$ 3.66 & $\bullet$ & \textbf{4.69 $\pm$ 3.43} &  & 5.01 $\pm$ 3.55 &  \\                               
		Iris & 3 & 150 & 0 & 4 & 0 & 2.76 $\pm$ 3.95 & 5.07 $\pm$ 5.22 & $\bullet$ & 4.31 $\pm$ 5.19 & $\bullet$ & \textbf{2.13 $\pm$ 3.65} & $\circ$ & 4.53 $\pm$ 5.31 & $\bullet$ & 4.49 $\pm$ 5.21 & $\bullet$ \\                      
		Landsat satellite & 2 & 6435 & 0 & 36 & 0 & 8.18 $\pm$ 1.06 & 8.03 $\pm$ 1.00 & $\circ$ & \textbf{7.75 $\pm$ 1.02} & $\circ$ & 8.59 $\pm$ 1.05 & $\bullet$ & 8.17 $\pm$ 1.02 &  & 8.58 $\pm$ 1.02 & $\bullet$ \\                  
		Letter & 26 & 20000 & 0 & 16 & 0 & \textbf{2.10 $\pm$ 0.33} & 3.36 $\pm$ 0.39 & $\bullet$ & 2.43 $\pm$ 0.32 & $\bullet$ & 2.43 $\pm$ 0.35 & $\bullet$ & 2.82 $\pm$ 0.37 & $\bullet$ & 3.29 $\pm$ 0.40 & $\bullet$ \\              
		Libras & 15 & 360 & 0 & 90 & 0 & 9.80 $\pm$ 4.68 & 18.54 $\pm$ 5.93 & $\bullet$ & \textbf{9.48 $\pm$ 4.47} &  & 11.26 $\pm$ 4.99 & $\bullet$ & 12.06 $\pm$ 4.95 & $\bullet$ & 13.59 $\pm$ 5.10 & $\bullet$ \\                     
		MAGIC & 2 & 19020 & 0 & 10 & 0 & \textbf{11.54 $\pm$ 0.68} & 11.90 $\pm$ 0.72 & $\bullet$ & 12.67 $\pm$ 0.71 & $\bullet$ & 11.69 $\pm$ 0.72 & $\bullet$ & 12.46 $\pm$ 0.70 & $\bullet$ & 12.59 $\pm$ 0.74 & $\bullet$ \\          
		Nursery & 5 & 12960 & 0 & 8 & 0 & 0.04 $\pm$ 0.07 & 0.19 $\pm$ 0.13 & $\bullet$ & \textbf{0.03 $\pm$ 0.06} & $\circ$ & 0.09 $\pm$ 0.10 & $\bullet$ & 0.10 $\pm$ 0.11 & $\bullet$ & 0.22 $\pm$ 0.15 & $\bullet$ \\                 
		ORL & 40 & 400 & 0 & 10304 & 0 & \textbf{1.58 $\pm$ 2.08} & 1.72 $\pm$ 2.03 &  & - &  & 2.05 $\pm$ 2.17 & $\bullet$ & 2.62 $\pm$ 2.58 & $\bullet$ & 3.55 $\pm$ 3.10 & $\bullet$ \\     
		Optical digits & 10 & 5620 & 0 & 64 & 0 & \textbf{1.26 $\pm$ 0.45} & 1.59 $\pm$ 0.49 & $\bullet$ & 1.29 $\pm$ 0.41 &  & 1.37 $\pm$ 0.46 & $\bullet$ & 1.46 $\pm$ 0.46 & $\bullet$ & 1.62 $\pm$ 0.49 & $\bullet$ \\  
			\hline 
			\end{tabular}      
			\hspace{20pt}
		\begin{tabular}{ccccccHHHHHHHHHHH}   
			\hline                                                       
			Dataset & $K$ & $N$ & $D_c$ & $D_r$ & \% Miss & CCF & RF & & Rot-For & & CCF-Bag & & Rand-Rot & & Rand-Rot Bag & \\
			\hline                                              
		Parkinsons & 2 & 195 & 0 & 22 & 0 & \textbf{5.87 $\pm$ 5.17} & 9.43 $\pm$ 6.17 & $\bullet$ & 7.19 $\pm$ 5.33 & $\bullet$ & 7.47 $\pm$ 6.04 & $\bullet$ & 6.70 $\pm$ 5.27 & $\bullet$ & 7.43 $\pm$ 5.64 & $\bullet$ \\             
		Pen digits & 10 & 10992 & 0 & 16 & 0 & \textbf{0.41 $\pm$ 0.19} & 0.82 $\pm$ 0.28 & $\bullet$ & 0.48 $\pm$ 0.22 & $\bullet$ & 0.45 $\pm$ 0.20 & $\bullet$ & 0.53 $\pm$ 0.23 & $\bullet$ & 0.63 $\pm$ 0.25 & $\bullet$ \\          
		Polya & 2 & 9255 & 0 & 169 & 0 & \textbf{21.02 $\pm$ 1.30} & 21.20 $\pm$ 1.27 & $\bullet$ & - &  & 21.15 $\pm$ 1.25 & $\bullet$ & 21.37 $\pm$ 1.35 & $\bullet$ & 21.89 $\pm$ 1.29 & $\bullet$ \\                      
		Seeds & 3 & 210 & 0 & 7 & 0 & \textbf{4.60 $\pm$ 4.56} & 5.78 $\pm$ 5.01 & $\bullet$ & 4.79 $\pm$ 4.60 &  & 5.46 $\pm$ 5.30 & $\bullet$ & 5.84 $\pm$ 4.91 & $\bullet$ & 6.41 $\pm$ 5.03 & $\bullet$ \\                            
		Skin seg & 2 & 245057 & 0 & 3 & 0 & 0.04 $\pm$ 0.01 & 0.05 $\pm$ 0.01 & $\bullet$ & 0.04 $\pm$ 0.01 &  & 0.04 $\pm$ 0.01 &  & 0.03 $\pm$ 0.01 & $\circ$ & \textbf{0.03 $\pm$ 0.01} & $\circ$ \\                                   
		Soybean & 19 & 683 & 13 & 22 & 3.29 & \textbf{5.22 $\pm$ 2.73} & 5.50 $\pm$ 3.02 &  & 5.67 $\pm$ 2.91 &  & 5.54 $\pm$ 2.94 & $\bullet$ & 5.58 $\pm$ 2.75 & $\bullet$ & 5.71 $\pm$ 2.90 & $\bullet$ \\                             
		Spirals & 3 & 10000 & 0 & 2 & 0 & 0.27 $\pm$ 0.16 & 1.21 $\pm$ 0.33 & $\bullet$ & 1.01 $\pm$ 0.32 & $\bullet$ & 0.27 $\pm$ 0.16 &  & 0.26 $\pm$ 0.16 &  & \textbf{0.26 $\pm$ 0.15} &  \\                                          
		Splice & 3 & 3190 & 60 & 0 & 0 & \textbf{3.04 $\pm$ 0.88} & 3.05 $\pm$ 0.93 &  & 4.21 $\pm$ 1.16 & $\bullet$ & 3.10 $\pm$ 0.87 &  & 10.06 $\pm$ 1.86 & $\bullet$ & 11.32 $\pm$ 2.03 & $\bullet$ \\                                
		Vehicle & 4 & 846 & 0 & 18 & 0 & \textbf{17.34 $\pm$ 4.13} & 25.22 $\pm$ 4.57 & $\bullet$ & 21.12 $\pm$ 4.27 & $\bullet$ & 17.36 $\pm$ 4.06 &  & 22.63 $\pm$ 4.19 & $\bullet$ & 22.95 $\pm$ 4.08 & $\bullet$ \\                   
		Vowel-c & 11 & 990 & 2 & 10 & 0 & 0.85 $\pm$ 0.89 & 2.56 $\pm$ 1.66 & $\bullet$ & 0.95 $\pm$ 0.94 &  & 1.21 $\pm$ 1.03 & $\bullet$ & \textbf{0.69 $\pm$ 0.90} &  & 1.06 $\pm$ 1.12 & $\bullet$ \\                                 
		Vowel-n & 11 & 990 & 0 & 10 & 0 & 1.84 $\pm$ 1.30 & 3.93 $\pm$ 1.98 & $\bullet$ & 1.43 $\pm$ 1.19 & $\circ$ & 2.51 $\pm$ 1.55 & $\bullet$ & \textbf{1.10 $\pm$ 1.06} & $\circ$ & 1.41 $\pm$ 1.10 & $\circ$ \\                     
		Waveform (1) & 3 & 5000 & 0 & 21 & 0 & 13.52 $\pm$ 1.56 & 15.06 $\pm$ 1.69 & $\bullet$ & 13.53 $\pm$ 1.50 &  & 13.46 $\pm$ 1.58 &  & 13.56 $\pm$ 1.56 &  & \textbf{13.37 $\pm$ 1.48} & $\circ$ \\                                 
		Waveform (2) & 3 & 5000 & 0 & 40 & 0 & 13.30 $\pm$ 1.68 & 14.61 $\pm$ 1.64 & $\bullet$ & 13.31 $\pm$ 1.64 &  & \textbf{13.28 $\pm$ 1.61} &  & 13.65 $\pm$ 1.72 & $\bullet$ & 13.81 $\pm$ 1.68 & $\bullet$ \\                      
		Wholesale-c & 2 & 440 & 1 & 7 & 0 & 8.58 $\pm$ 4.06 & \textbf{8.15 $\pm$ 4.01} &  & 8.44 $\pm$ 3.99 &  & 8.47 $\pm$ 4.03 &  & 8.65 $\pm$ 4.17 &  & 8.45 $\pm$ 4.01 &  \\                                                          
		Wholesale-r & 3 & 440 & 0 & 7 & 0 & 30.95 $\pm$ 5.84 & 28.97 $\pm$ 6.10 & $\circ$ & \textbf{28.18 $\pm$ 6.19} & $\circ$ & 28.80 $\pm$ 5.89 & $\circ$ & 32.29 $\pm$ 5.91 & $\bullet$ & 29.62 $\pm$ 5.89 & $\circ$ \\               
		Wisconsin & 2 & 699 & 0 & 9 & 0.25 & 3.23 $\pm$ 2.02 & 3.54 $\pm$ 2.17 & $\bullet$ & \textbf{2.87 $\pm$ 1.82} & $\circ$ & 3.02 $\pm$ 1.90 & $\circ$ & 3.03 $\pm$ 1.97 & $\circ$ & 2.91 $\pm$ 1.83 & $\circ$ \\             
		Yeast & 10 & 1484 & 0 & 8 & 0 & 38.47 $\pm$ 4.04 & 37.89 $\pm$ 4.22 & $\circ$ & \textbf{37.21 $\pm$ 4.23} & $\circ$ & 37.27 $\pm$ 3.94 & $\circ$ & 38.41 $\pm$ 4.15 &  & 37.72 $\pm$ 4.10 & $\circ$ \\                            
		Zoo & 7 & 101 & 0 & 16 & 0 & \textbf{3.33 $\pm$ 5.75} & 5.13 $\pm$ 6.73 & $\bullet$ & 5.47 $\pm$ 6.71 & $\bullet$ & 3.60 $\pm$ 5.83 &  & 4.67 $\pm$ 6.72 & $\bullet$ & 4.67 $\pm$ 6.72 & $\bullet$ \\  &&&&&&&&&&&&&&&&     \\      
		\hline                                         
	\end{tabular}                                                          
\end{table*}  

\begin{table*}[p]                        
	\centering    
	\scriptsize             
	\setlength\tabcolsep{3pt}	  
	\renewcommand{\arraystretch}{1.2}    
	\caption{Mean and standard deviations of percentage of test cases misclassified.  Method with best accuracy is shown in bold. $\bullet$ and $\circ$ indicate that CCFs were significantly better and worse respectively at the 1\% level of a Wilcoxon signed rank test.  The \textit{ORL} and \textit{Polya} datasets could not be run in reasonable time for rotation forest. \label{table:ResultsTable}}                            
	\begin{tabular}{|c|HHHHHc|cc|cc|cc|cc|cc|}                    
		\hline                                                       
		Dataset & $K$ & $N$ & $D_c$ & $D_r$ & \% Miss & CCF & RF & & Rot-For & & CCF-Bag & & RRF & & RRF-Bag & \\
		\hline           
Balance scale & 3 & 625 & 0 & 4 & 0 & 9.60 $\pm$ 3.73 & 18.66 $\pm$ 4.56 & $\bullet$ & \textbf{7.25 $\pm$ 3.38} & $\circ$ & 8.80 $\pm$ 3.82 & $\circ$ & 13.53 $\pm$ 3.85 & $\bullet$ & 10.85 $\pm$ 3.89 & $\bullet$ \\            
Banknote & 2 & 1372 & 0 & 4 & 0 & \textbf{0.00 $\pm$ 0.00} & 0.58 $\pm$ 0.64 & $\bullet$ & \textbf{0.00 $\pm$ 0.00} &  & \textbf{0.00 $\pm$ 0.00} &  & \textbf{0.00 $\pm$ 0.00} &  & 0.02 $\pm$ 0.12 &  \\                        
Breast tissue & 6 & 106 & 0 & 9 & 0 & 28.18 $\pm$ 12.00 & 31.09 $\pm$ 12.38 & $\bullet$ & 28.48 $\pm$ 12.36 &  & 28.36 $\pm$ 11.82 &  & \textbf{26.06 $\pm$ 11.85} & $\circ$ & 26.24 $\pm$ 12.08 & $\circ$ \\                     
Climate crashes & 2 & 360 & 0 & 18 & 0 & \textbf{5.81 $\pm$ 4.10} & 6.46 $\pm$ 4.10 & $\bullet$ & 6.11 $\pm$ 3.94 & $\bullet$ & 6.56 $\pm$ 4.16 & $\bullet$ & 7.06 $\pm$ 4.15 & $\bullet$ & 7.22 $\pm$ 4.15 & $\bullet$ \\        
Fertility & 2 & 100 & 0 & 9 & 0 & 13.33 $\pm$ 9.67 & 14.53 $\pm$ 9.45 & $\bullet$ & \textbf{12.40 $\pm$ 8.87} &  & 13.53 $\pm$ 9.70 &  & 12.67 $\pm$ 9.10 & $\circ$ & 12.87 $\pm$ 9.22 &  \\                                      
Heart-SPECT & 2 & 267 & 0 & 22 & 0 & 18.59 $\pm$ 6.97 & 18.86 $\pm$ 7.25 &  & \textbf{17.56 $\pm$ 7.29} & $\circ$ & 18.12 $\pm$ 7.12 &  & 18.17 $\pm$ 6.88 &  & 18.37 $\pm$ 7.06 &  \\                                            
Heart-SPECTF & 2 & 267 & 0 & 44 & 0 & 18.64 $\pm$ 7.36 & 18.91 $\pm$ 7.09 &  & 18.57 $\pm$ 7.06 &  & \textbf{18.25 $\pm$ 7.10} &  & 19.26 $\pm$ 7.63 &  & 19.56 $\pm$ 7.73 & $\bullet$ \\                                         
Hill valley & 2 & 1212 & 0 & 100 & 0 & \textbf{0.00 $\pm$ 0.00} & 39.16 $\pm$ 4.05 & $\bullet$ & 6.32 $\pm$ 2.65 & $\bullet$ & \textbf{0.00 $\pm$ 0.00} &  & 4.17 $\pm$ 1.86 & $\bullet$ & 6.65 $\pm$ 2.28 & $\bullet$ \\         
Hill valley noisy & 2 & 1212 & 0 & 100 & 0 & \textbf{4.73 $\pm$ 1.90} & 42.08 $\pm$ 4.57 & $\bullet$ & 11.01 $\pm$ 2.80 & $\bullet$ & 5.24 $\pm$ 1.84 & $\bullet$ & 15.78 $\pm$ 3.44 & $\bullet$ & 18.64 $\pm$ 3.88 & $\bullet$ \\
ILPD & 2 & 640 & 0 & 10 & 0 & 27.56 $\pm$ 5.08 & 29.92 $\pm$ 5.17 & $\bullet$ & 29.02 $\pm$ 5.25 & $\bullet$ & 28.20 $\pm$ 5.25 & $\bullet$ & 27.55 $\pm$ 4.94 &  & \textbf{27.29 $\pm$ 4.98} &  \\                               
Ionosphere & 2 & 351 & 0 & 33 & 0 & 4.82 $\pm$ 3.44 & 6.53 $\pm$ 3.95 & $\bullet$ & 5.79 $\pm$ 3.56 & $\bullet$ & 5.50 $\pm$ 3.66 & $\bullet$ & \textbf{4.69 $\pm$ 3.43} &  & 5.01 $\pm$ 3.55 &  \\                               
Iris & 3 & 150 & 0 & 4 & 0 & 2.76 $\pm$ 3.95 & 5.07 $\pm$ 5.22 & $\bullet$ & 4.31 $\pm$ 5.19 & $\bullet$ & \textbf{2.13 $\pm$ 3.65} & $\circ$ & 4.53 $\pm$ 5.31 & $\bullet$ & 4.49 $\pm$ 5.21 & $\bullet$ \\                      
Landsat satellite & 2 & 6435 & 0 & 36 & 0 & 8.18 $\pm$ 1.06 & 8.03 $\pm$ 1.00 & $\circ$ & \textbf{7.75 $\pm$ 1.02} & $\circ$ & 8.59 $\pm$ 1.05 & $\bullet$ & 8.17 $\pm$ 1.02 &  & 8.58 $\pm$ 1.02 & $\bullet$ \\                  
Letter & 26 & 20000 & 0 & 16 & 0 & \textbf{2.10 $\pm$ 0.33} & 3.36 $\pm$ 0.39 & $\bullet$ & 2.43 $\pm$ 0.32 & $\bullet$ & 2.43 $\pm$ 0.35 & $\bullet$ & 2.82 $\pm$ 0.37 & $\bullet$ & 3.29 $\pm$ 0.40 & $\bullet$ \\              
Libras & 15 & 360 & 0 & 90 & 0 & 9.80 $\pm$ 4.68 & 18.54 $\pm$ 5.93 & $\bullet$ & \textbf{9.48 $\pm$ 4.47} &  & 11.26 $\pm$ 4.99 & $\bullet$ & 12.06 $\pm$ 4.95 & $\bullet$ & 13.59 $\pm$ 5.10 & $\bullet$ \\                     
MAGIC & 2 & 19020 & 0 & 10 & 0 & \textbf{11.54 $\pm$ 0.68} & 11.90 $\pm$ 0.72 & $\bullet$ & 12.67 $\pm$ 0.71 & $\bullet$ & 11.69 $\pm$ 0.72 & $\bullet$ & 12.46 $\pm$ 0.70 & $\bullet$ & 12.59 $\pm$ 0.74 & $\bullet$ \\          
Nursery & 5 & 12960 & 0 & 8 & 0 & 0.04 $\pm$ 0.07 & 0.19 $\pm$ 0.13 & $\bullet$ & \textbf{0.03 $\pm$ 0.06} & $\circ$ & 0.09 $\pm$ 0.10 & $\bullet$ & 0.10 $\pm$ 0.11 & $\bullet$ & 0.22 $\pm$ 0.15 & $\bullet$ \\                 
ORL & 40 & 400 & 0 & 10304 & 0 & \textbf{1.58 $\pm$ 2.08} & 1.72 $\pm$ 2.03 &  & - &  & 2.05 $\pm$ 2.17 & $\bullet$ & 2.62 $\pm$ 2.58 & $\bullet$ & 3.55 $\pm$ 3.10 & $\bullet$ \\                                    
Optical digits & 10 & 5620 & 0 & 64 & 0 & \textbf{1.26 $\pm$ 0.45} & 1.59 $\pm$ 0.49 & $\bullet$ & 1.29 $\pm$ 0.41 &  & 1.37 $\pm$ 0.46 & $\bullet$ & 1.46 $\pm$ 0.46 & $\bullet$ & 1.62 $\pm$ 0.49 & $\bullet$ \\                
Parkinsons & 2 & 195 & 0 & 22 & 0 & \textbf{5.87 $\pm$ 5.17} & 9.43 $\pm$ 6.17 & $\bullet$ & 7.19 $\pm$ 5.33 & $\bullet$ & 7.47 $\pm$ 6.04 & $\bullet$ & 6.70 $\pm$ 5.27 & $\bullet$ & 7.43 $\pm$ 5.64 & $\bullet$ \\             
Pen digits & 10 & 10992 & 0 & 16 & 0 & \textbf{0.41 $\pm$ 0.19} & 0.82 $\pm$ 0.28 & $\bullet$ & 0.48 $\pm$ 0.22 & $\bullet$ & 0.45 $\pm$ 0.20 & $\bullet$ & 0.53 $\pm$ 0.23 & $\bullet$ & 0.63 $\pm$ 0.25 & $\bullet$ \\          
Polya & 2 & 9255 & 0 & 169 & 0 & \textbf{21.02 $\pm$ 1.30} & 21.20 $\pm$ 1.27 & $\bullet$ & - &  & 21.15 $\pm$ 1.25 & $\bullet$ & 21.37 $\pm$ 1.35 & $\bullet$ & 21.89 $\pm$ 1.29 & $\bullet$ \\                      
Seeds & 3 & 210 & 0 & 7 & 0 & \textbf{4.60 $\pm$ 4.56} & 5.78 $\pm$ 5.01 & $\bullet$ & 4.79 $\pm$ 4.60 &  & 5.46 $\pm$ 5.30 & $\bullet$ & 5.84 $\pm$ 4.91 & $\bullet$ & 6.41 $\pm$ 5.03 & $\bullet$ \\                            
Skin seg & 2 & 245057 & 0 & 3 & 0 & \textbf{0.03 $\pm$ 0.01} & 0.05 $\pm$ 0.01 & $\bullet$ & 0.04 $\pm$ 0.01 & $\bullet$ & 0.03 $\pm$ 0.01 & $\bullet$ & 0.03 $\pm$ 0.01 & $\bullet$ & 0.03 $\pm$ 0.01 & $\bullet$ \\             
Soybean & 19 & 683 & 13 & 22 & 3.29 & \textbf{5.22 $\pm$ 2.73} & 5.50 $\pm$ 3.02 &  & 5.67 $\pm$ 2.91 &  & 5.54 $\pm$ 2.94 & $\bullet$ & 5.58 $\pm$ 2.75 & $\bullet$ & 5.71 $\pm$ 2.90 & $\bullet$ \\                             
Spirals & 3 & 10000 & 0 & 2 & 0 & 0.27 $\pm$ 0.16 & 1.21 $\pm$ 0.33 & $\bullet$ & 1.01 $\pm$ 0.32 & $\bullet$ & 0.27 $\pm$ 0.16 &  & 0.26 $\pm$ 0.16 &  & \textbf{0.26 $\pm$ 0.15} &  \\                                          
Splice & 3 & 3190 & 60 & 0 & 0 & \textbf{3.04 $\pm$ 0.88} & 3.05 $\pm$ 0.93 &  & 4.21 $\pm$ 1.16 & $\bullet$ & 3.10 $\pm$ 0.87 &  & 10.06 $\pm$ 1.86 & $\bullet$ & 11.32 $\pm$ 2.03 & $\bullet$ \\                                
Vehicle & 4 & 846 & 0 & 18 & 0 & \textbf{17.34 $\pm$ 4.13} & 25.22 $\pm$ 4.57 & $\bullet$ & 21.12 $\pm$ 4.27 & $\bullet$ & 17.36 $\pm$ 4.06 &  & 22.63 $\pm$ 4.19 & $\bullet$ & 22.95 $\pm$ 4.08 & $\bullet$ \\                   
Vowel-c & 11 & 990 & 2 & 10 & 0 & 0.85 $\pm$ 0.89 & 2.56 $\pm$ 1.66 & $\bullet$ & 0.95 $\pm$ 0.94 &  & 1.21 $\pm$ 1.03 & $\bullet$ & \textbf{0.69 $\pm$ 0.90} &  & 1.06 $\pm$ 1.12 & $\bullet$ \\                                 
Vowel-n & 11 & 990 & 0 & 10 & 0 & 1.84 $\pm$ 1.30 & 3.93 $\pm$ 1.98 & $\bullet$ & 1.43 $\pm$ 1.19 & $\circ$ & 2.51 $\pm$ 1.55 & $\bullet$ & \textbf{1.10 $\pm$ 1.06} & $\circ$ & 1.41 $\pm$ 1.10 & $\circ$ \\                     
Waveform (1) & 3 & 5000 & 0 & 21 & 0 & 13.52 $\pm$ 1.56 & 15.06 $\pm$ 1.69 & $\bullet$ & 13.53 $\pm$ 1.50 &  & 13.46 $\pm$ 1.58 &  & 13.56 $\pm$ 1.56 &  & \textbf{13.37 $\pm$ 1.48} & $\circ$ \\                                 
Waveform (2) & 3 & 5000 & 0 & 40 & 0 & 13.30 $\pm$ 1.68 & 14.61 $\pm$ 1.64 & $\bullet$ & 13.31 $\pm$ 1.64 &  & \textbf{13.28 $\pm$ 1.61} &  & 13.65 $\pm$ 1.72 & $\bullet$ & 13.81 $\pm$ 1.68 & $\bullet$ \\                      
Wholesale-c & 2 & 440 & 1 & 7 & 0 & 8.58 $\pm$ 4.06 & \textbf{8.15 $\pm$ 4.01} &  & 8.44 $\pm$ 3.99 &  & 8.47 $\pm$ 4.03 &  & 8.65 $\pm$ 4.17 &  & 8.45 $\pm$ 4.01 &  \\                                                          
Wholesale-r & 3 & 440 & 0 & 7 & 0 & 30.95 $\pm$ 5.84 & 28.97 $\pm$ 6.10 & $\circ$ & \textbf{28.18 $\pm$ 6.19} & $\circ$ & 28.80 $\pm$ 5.89 & $\circ$ & 32.29 $\pm$ 5.91 & $\bullet$ & 29.62 $\pm$ 5.89 & $\circ$ \\               
Wisconsin cancer & 2 & 699 & 0 & 9 & 0.25 & 3.23 $\pm$ 2.02 & 3.54 $\pm$ 2.17 & $\bullet$ & \textbf{2.87 $\pm$ 1.82} & $\circ$ & 3.02 $\pm$ 1.90 & $\circ$ & 3.03 $\pm$ 1.97 & $\circ$ & 2.91 $\pm$ 1.83 & $\circ$ \\             
Yeast & 10 & 1484 & 0 & 8 & 0 & 38.47 $\pm$ 4.04 & 37.89 $\pm$ 4.22 & $\circ$ & \textbf{37.21 $\pm$ 4.23} & $\circ$ & 37.27 $\pm$ 3.94 & $\circ$ & 38.41 $\pm$ 4.15 &  & 37.72 $\pm$ 4.10 & $\circ$ \\                            
Zoo & 7 & 101 & 0 & 16 & 0 & \textbf{3.33 $\pm$ 5.75} & 5.13 $\pm$ 6.73 & $\bullet$ & 5.47 $\pm$ 6.71 & $\bullet$ & 3.60 $\pm$ 5.83 &  & 4.67 $\pm$ 6.72 & $\bullet$ & 4.67 $\pm$ 6.72 & $\bullet$ \\           
		\hline                                         
	\end{tabular}                                                          
\end{table*}  
	
The majority of the 37 datasets considered were taken from the UCI machine learning 
database\footnote{\scriptsize \url{https://archive.ics.uci.edu/ml/datasets.html}}
 \citep{Lichman2013UCI} with the exceptions of the \textit{ORL} face recognition dataset \citep{samaria1994parameterisation}, the \textit{Polyadenylation Signal Prediction} (\textit{polya}) dataset \citep{li2002kent}, and the artificial spiral dataset from Figure \ref{fig:decisionSurface}.    Summaries of the datasets are given in Table \ref{table:summary}.  Note for the \textit{vowel-c} dataset the sex and identifier for the speaker are included, whereas these are omitted for the \textit{vowel-n} dataset which is otherwise identical.  The \textit{wholesale-c} and \textit{wholesale-r} datasets correspond to predicting the \textit{channel} and \textit{region} attributes respectively.  

For each dataset, 15 different 10-fold cross-validation tests were performed, 
results from which are given in Table~\ref{table:ResultsTable}.  
Table \ref{table:nVics} shows a summary of results over all the datasets, 
giving the number of datasets for which the performance of one dataset was 
significantly better than another at the 1\% level of a Wilcoxon signed rank test.  
Table~\ref{table:rank} shows the average performance ranks of the algorithms.
These results show that CCFs performed excellently, comfortably outperforming
all the other approaches.  
The improvement compared to RFs was particularly large, with RFs having on average 1.74 times as many 
misclassifications as CCFs (ignoring the two datasets where CCFs had perfect accuracy).  To give 
another perspective, if one were using RFs and decided to switch to CCFs, then the number of 
misclassifications would be reduced by a factor of 28.7\% on average for the tested datasets.  
We emphasise that these are improvements over a generic selection of datasets, rather than being
carefully chosen problems. Therefore the magnitude and regularity ($33$ of $37$ datasets had
lower error for CCFs than RFs) of the improvements represents a substantial advancement.
In Section~\ref{sec:effectOfCorrelation} we will show that there are certain circumstances,
namely where there is high correlation between the input features, for which the improvements are
typically significantly larger than this.  For example, on the highly correlated~\emph{hill-valley} dataset the error
was reduced from $39.16\%$ for RF to $0\%$ for CCFs.
However, the results also show that the improvements do not come from better dealing with
input correlations alone, with a number of datsets whose inputs are not correlated showing
significant improvements.  For example, the error for the~\emph{balance scale} dataset
was roughly halved, despite there being zero correlation between its inputs features.
\begin{table}[p]	
	\renewcommand{\arraystretch}{1.2}
	\footnotesize
	\centering
	\caption{Number of victories column vs row at 1\% significance level of Wilcoxon signed rank test for classification experiments (37 total).
		CCF victories and losses are given in blue and red respectively. \label{table:nVics}}
	\begin{tabular}{lcccccc} \toprule
		& \text{\textbf{\color{blue} CCF}} & \text{RF} & \text{Rotation Forest} & \text{CCF-Bag} & \text{RRF} & \text{RRF-Bag}  \\ \midrule
		\textbf{\color{red} CCF} & - & \textbf{\color{red} 3} & \textbf{\color{red} 8} & \textbf{\color{red} 5} & \textbf{\color{red} 4} & \textbf{\color{red} 6} \\  
		\text{RF} & \textbf{\color{blue} 28} & - & 27 & 26  & 24 & 21\\  
		\text{Rotation Forest} & \textbf{\color{blue} 15}& 2 & - & 10 & 10 & 8\\  
		\text{CCF-Bag} & \textbf{\color{blue} 19} & 1 & 11 & - & 5 & 5 \\  
		\text{RRF} & \textbf{\color{blue} 22} & 8 & 15 & 18 & - & 4\\ 
		\text{RRF-Bag} & \textbf{\color{blue} 24} & 9 & 17 & 22 & 20 & - \\ \bottomrule
	\end{tabular}
\end{table}
\begin{table}[p]
	\renewcommand{\arraystretch}{1.2}
	\footnotesize
	\centering
	\caption{Mean rank across datasets of mean test misclassification rate.  Lower
		rank corresponds to better performance (average rank $=3.5$). \textit{ORL} and \textit{Polya} datasets
		are omitted from comparison to avoid biasing against rotation forests. \label{table:rank}}
	\begin{tabular}{lc} \toprule
		\text{Algorithm} & \text{Mean Rank}  \\ \midrule
		\text{CCF} & 2.43 \\
		\text{CCF-Bag} & 2.91 \\  
		\text{Rotation Forest} & 2.96 \\  
		\text{RRF} &  3.60 \\ 
		\text{RRF-Bag} &  4.13 \\ 
		\text{RF} & 4.97 \\  \bottomrule
	\end{tabular}
\end{table}

Elsewhere, the domination of CCFs over CCF-Bag highlights the improvement from using projection 
bootstrapping instead of bagging, while the good performance on a large variety of datasets 
demonstrates the robustness and wide ranging applicability of CCFs.  The more computationally
intensive rotation forest algorithm performed
reasonably well, but was still comfortably outperformed by CCFs on average.

Another interesting result was that RRFs delivered significantly better predictive performance than RRF-Bag.
Previous approaches using per-node random rotations, such as Forest-RC \citep{breiman2001random} and
randomer forests~\citep{tomita2015randomer}, have employed bagging in addition to random rotations.  These
results suggest that this is actually detrimental to the ensemble performance and supports the
motivation for projection bootstrapping provided in Section~\ref{sec:projBoot}.

%

\begin{table}[p]                        
	\centering    
	\scriptsize             
	\setlength\tabcolsep{4pt}	  
	\renewcommand{\arraystretch}{1.2}          
	\caption{Mean and standard deviations of percentage of test cases misclassified for CCF and algorithms from \cite{zhang2014random}.  All results are from an ensemble of 100 trees.  Method with best accuracy is shown in bold. $\bullet$ and $\circ$ indicate that CCFs were significantly better and worse respectively at the 1\% level of a t-test. \label{table:ldarf}}                    
	\begin{tabular}{|c|c|cc|cc|cc|}                    
		\hline                                                       
		Dataset & CCF & PCA-RF & & LDA-RF & & RF-ensemble &\\
		\hline           
		Balance scale & \textbf{9.70 $\pm$ 3.57} & 12.41 $\pm$ 3.53 & $\bullet$ & 11.46 $\pm$ 3.44 & $\bullet$ & 11.91 $\pm$ 3.41 & $\bullet$ \\
		Breast tissue & 27.82 $\pm$ 11.79 & \textbf{26.80 $\pm$ 13.39} &  & 27.64 $\pm$ 13.01 &  & 28.00 $\pm$ 13.93 &  \\                      
		Climate crashes & 5.78 $\pm$ 3.83 & 8.81 $\pm$ 3.38 & $\bullet$ & 7.70 $\pm$ 3.50 & $\bullet$ & \textbf{2.72 $\pm$ 3.42} & $\circ$ \\   
		Iris & \textbf{2.58 $\pm$ 3.92} & 4.53 $\pm$ 4.84 & $\bullet$ & 4.63 $\pm$ 5.06 & $\bullet$ & 4.61 $\pm$ 5.21 & $\bullet$ \\            
		Libras & \textbf{9.72 $\pm$ 4.75} & 12.24 $\pm$ 5.57 & $\bullet$ & 15.13 $\pm$ 5.63 & $\bullet$ & 14.10 $\pm$ 5.69 & $\bullet$ \\       
		Nursery & \textbf{0.05 $\pm$ 0.07} & 0.29 $\pm$ 0.15 & $\bullet$ & 0.19 $\pm$ 0.13 & $\bullet$ & 0.26 $\pm$ 0.18 & $\bullet$ \\         
		Parkinsons & \textbf{6.57 $\pm$ 5.53} & 8.23 $\pm$ 6.15 & $\bullet$ & 9.89 $\pm$ 6.93 & $\bullet$ & 8.77 $\pm$ 6.38 & $\bullet$ \\      
		Seeds & \textbf{4.89 $\pm$ 4.55} & 6.13 $\pm$ 5.21 &  & 6.04 $\pm$ 4.50 &  & 6.06 $\pm$ 4.80 &  \\                                      
		Splice & \textbf{3.30 $\pm$ 0.97} & 9.04 $\pm$ 1.62 & $\bullet$ & 7.90 $\pm$ 1.79 & $\bullet$ & 5.78 $\pm$ 1.36 & $\bullet$ \\          
		Vehicle & \textbf{17.44 $\pm$ 4.19} & 21.33 $\pm$ 4.47 & $\bullet$ & 20.66 $\pm$ 4.60 & $\bullet$ & 21.77 $\pm$ 4.45 & $\bullet$ \\     
		Waveform (1) & \textbf{13.53 $\pm$ 1.57} & 13.81 $\pm$ 1.53 &  & 14.12 $\pm$ 1.50 & $\bullet$ & 13.78 $\pm$ 1.51 &  \\                  
		Waveform (2) & 13.76 $\pm$ 1.65 & \textbf{13.72 $\pm$ 1.62} &  & 14.12 $\pm$ 1.54 &  & 13.79 $\pm$ 1.60 &  \\                           
		Wisconsin & 3.31 $\pm$ 2.11 & 2.92 $\pm$ 2.01 &  & 3.02 $\pm$ 1.91 &  & \textbf{2.84 $\pm$ 1.91} &  \\                           
		Yeast & 38.79 $\pm$ 4.13 & 37.82 $\pm$ 3.65 &  & \textbf{37.77 $\pm$ 3.69} &  & 37.92 $\pm$ 3.65 &  \\       
		\hline                                         
	\end{tabular}                                                            
\end{table}  

\subsubsection{Comparison to Random Forests with ensemble of feature spaces}
\label{sec:relatedWork}

We next make comparisons to the PCA-RF, LDA-RF, and RF-ensemble methods introduced by
\cite{zhang2014random}.  
As \cite{zhang2014random} performed the same cross validation tests on 14 of the common datasets, we make 
direct comparisons to their quoted results.  Taking the first 100 trees from our experiments to match the ensemble sizes gives the comparisons provided in Table~\ref{table:ldarf}.  We see that CCFs consistently outperform
all three methods, giving win/draw/loss ratios at the 1\% significance level of a t-test for CCFs of 8/6/0,  9/5/0, and
7/6/1 compared to PCA-RF, LDA-RF and RF-ensemble respectively.  On average PCA-RF, LDA-RF, and RF-ensemble had 1.66, 1.51 and 1.49 times as many misclassifications as CCFs respectively.
We reiterate that, other than PCA-RF, these methods do not extend to the regression and multiple output
cases like CCFs and the other approaches we consider do.

\subsection{Comparison to Other Classifiers}
\label{sec:OtherClassifiers}

In order to provide comparison to a wider array of classifiers, we also tested CCFs using the experimental setup of \cite{fernandez2014we} from their recent survey of 179 classifiers applied to 121 datasets.  We used the same partitions which were a mix of 4-fold cross validations and predefined train / test splits.\footnote{\raggedright Partitions and results from \cite{fernandez2014we} are available at http://persoal.citius.usc.es/manuel.fernandez.delgado/papers/jmlr/}  We omitted datasets containing non-ordinal features on the basis that they pre-processed such features by treating the category indexes as numeric features, which is substantially different to our recommended procedure (see Section \ref{sec:DataPrepro}).  
For consistency with their approach, we fixed the missing features values (to the feature mean), rather than 
using the procedure laid out in Section~\ref{sec:DataPrepro}.
As a check that tests were run correctly, we also ran \textsc{matlab}'s {TreeBagger} algorithm, with $500$ trees and $\lambda = \ceil*{\log_2 \left(D\right) +1}$, which as expected gave performance on par with the very similar {rforest\_R} algorithm.  
An exception to this was the \textit{image-segmentation} dataset for which we were unable to replicate the {rforest\_R} results 
to within any plausible level of variability.  We therefore omitted this dataset from the results.
In total 82 datasets were compared, for which summaries along with detailed results are available in Appendix \ref{sec:detailedResults}, while a summary of the results for the top 20 performing classifiers (based on average accuracy rank) is given in Table \ref{table:otherClass}.
 We note that our quoted results for the competing methods differs
from those of the original paper due to using only a subset of the original datasets
(averages were reprocessed using the individual dataset scores provided by~\cite{fernandez2014we}).
We also note that the relatively poor
performance of rotation forests that is quoted most likely stems from the fact that
they set $L=10$ for rotation forests and $L=500$ for most of the RF implementations, constituting an arguably
unfair comparison.

\begin{table}[t!]                        
	\centering    
	\scriptsize             
	\setlength\tabcolsep{3.25pt}	  
	\renewcommand{\arraystretch}{1.2}    
	\caption{Comparison of top 20 performing classifiers on 82 UCI datasets. R is the mean rank over all 181 classifiers according to error rate; E is the mean error rate (\%); $\kappa$ is the mean Cohen's $\kappa$ (unlike others, higher is better); $\text{R}_{\text{CCF}}$, $\text{E}_{\text{CCF}}$ and $\kappa_{\text{CCF}}$ are the respective values for CCFs on the datasets where the competing classifier successfully ran (note CCFs successfully ran on all datasets); $N_v$ and $N_l$ are the number of datasets where the CCFs $\kappa$ was higher and lower than the classifier respectively;\protect\footnotemark and p is the p-value for whether the CCFs $\kappa$ mean is higher using a Wilcoxon signed rank test.  Classifier types: SVM = support vector machine, NNET = neural net, RF = random forest, Bag = bagging, and BST = boosting.  Results for CCF, rForest\_R, TreeBagger, and RotationForest are all
		statistically valid, while results for the other datasets are potentially inflated, such that these classifiers
		may actually perform worse than quoted, see main text for details.
		For further details on classifiers see \cite{fernandez2014we}. \label{table:otherClass}}                            
	\begin{tabular}{|c|c|c|c|c|c|c|c|c|c|}                    
		\hline                                                       
		Classifier & R & $\text{R}_{\text{CCF}}$  & E & $\text{E}_{\text{CCF}}$ & $\kappa$ & $\kappa_{\text{CCF}}$ & $N_v$ & $N_l$ & p \\
		\hline           
		\textbf{CCF} & \textbf{28.87} & - & \textbf{14.08} & - & \textbf{70.67} & - & - & - & - \\                                             
		svmPoly (SVM) & 31.53 & 30.57 & 15.73 & 14.27 & 65.10 & 69.61 & 54 & 25 & 2.3e-4 \\                              
		svmRadialCost (SVM) & 31.84 & 30.57 & 15.33 & 14.27 & 66.55 & 69.61 & 43 & 36 & 0.11 \\                           
		svm\_C (SVM) & 32 & 29.21 & 15.67 & 14.18 & 67.65 & 70.49 & 46 & 32 & 0.18 \\                                        
		elm\_kernel (NNET) & 32.19 & 30.14 & 15.20 & 14.54 & 69.01 & 69.75 & 42 & 36 & 0.16 \\                       
		parRF (RF) & 33.03 & 28.87 & 15.54 & 14.08 & 67.73 & 70.67 & 52 & 27 & 0.014 \\                                   
		svmRadial (SVM) & 33.77 & 30.57 & 15.68 & 14.27 & 65.88 & 69.61 & 50 & 28 & 1.6e-3 \\                             
		rf\_caret (RF) & 34.48 & 29.21 & 15.56 & 14.18 & 67.67 & 70.49 & 54 & 23 & 6.3e-4 \\                               
		rforest\_R (RF) & 40.70 & 29.21 & 15.82 & 14.18 & 66.67 & 70.49 & 57 & 20 & 2e-5 \\                                
		\textbf{TreeBagger} (RF) & 40.91 & 28.87 & 15.75 & 14.08 & 67.51 & 70.67 & 55 & 23 & 3.5e-5 \\                              
		Bag\_LibSVM (Bag) & 42.28 & 29.48 & 16.65 & 14.25 & 58.13 & 70.27 & 70 & 12 & 3.2e-12 \\                          
		C50\_t (BST) & 42.61 & 28.87 & 16.85 & 14.08 & 66.11 & 70.67 & 57 & 22 & 4.0e-4 \\                                   
		nnet\_t (NNET) & 42.87 & 28.87 & 18.74 & 14.08 & 64.72 & 70.67 & 54 & 26 & 1.5e-3 \\                                  
		avNNet\_t (NNET) & 43.26 & 28.87 & 18.77 & 14.08 & 64.88 & 70.67 & 50 & 29 & 1.0e-3 \\                                 
		RotationForest & 44.62 & 28.87 & 16.64 & 14.08 & 65.34 & 70.67 & 64 & 15 & 7.1e-9 \\                       
		pcaNNet\_t (NNET) & 45.86 & 28.87 & 19.28 & 14.08 & 63.83 & 70.67 & 54 & 25 & 1.5e-4 \\                              
		mlp\_t (NNET) & 46.06 & 28.87 & 17.38 & 14.08 & 66.75 & 70.67 & 54 & 26 & 1.6e-3 \\                                   
		LibSVM (SVM) & 46.50 & 28.87 & 16.65 & 14.08 & 63.80 & 70.67 & 57 & 21 & 2.9e-6 \\                               
		MB\_LibSVM (BST) & 46.90 & 28.87 & 16.82 & 14.25 & 64.47 & 70.27 & 61 & 17 & 1.8e-6 \\                         
		Regularized RF\_t (RF) & 49.56 & 28.87 & 16.71 & 14.18 & 66.30 & 70.49 & 59 & 20 & 1.1e-5 \\        
		
		\hline                                         
	\end{tabular}                                                          
\end{table}   
\footnotetext{Note that these are \textit{not} the number significant victories / losses as quoted elsewhere
	due to the small number of folds considered.}

Whereas \cite{fernandez2014we} included an, often intensive, parameter tuning step in their tests, we used CCFs in an out-of-the-box fashion, taking the same parameters as in Section~\ref{sec:param}.   
Despite this, CCFs outperformed all other tested classifiers based on every calculated performance metric.  This superior performance is particularly impressive in light of the
observation by~\cite{wainberg2016random}
that the method~\cite{fernandez2014we} use for the parameter tuning
positively biases the estimated predictive accuracy and $\kappa$ score~\citep{carletta1996assessing} 
 (an thus negatively biases the error). 
Namely, their parameter tuning
step makes use of the full dataset (train and test), such that the quoted performance on the test set is inflated.
Crucially, this does not affect methods that do not use a parameter tuning step - 
specifically CCFs, {rForest\_R}, {TreeBagger}, and rotation forests from those in shown in Table~\ref{table:otherClass}.
As the results for others are all biased towards better performance, 
improvements of CCFs over these methods should always be
at least as large as those quoted and never less.  Similarly for cases were the improvement for CCFs is shown to
be statistically significant, then this significance result still holds.

As also noted by~\cite{wainberg2016random}, the procedure of~\cite{fernandez2014we} was further biased in
its consideration of failed runs as the quoted ranks, average accuracies, and $\kappa$ scores
omit failed runs in the quoted average for each classifier.
This causes bias because, for example, a classifier that failed to run a dataset where the best attainable
error rate was relatively high will have its average error negatively biased.  We note that experiments for the CCFs and
TreeBagger runs had no such failed runs.  
This bias does not effect the significance test comparisons or number of victories / losses compared with CCFs 
as these comparisons are based only on the datasets were the competing classifier 
successfully ran on all folds.  Note also that none of the top 20
classifiers had more than 4 failures so the effects of failed runs was relatively small.  
However, in the interest of correctly comparing the mean rank,
mean error, and mean $\kappa$ between CCFs and the alternatives, we also provide the average metrics for CCFs on the subset
of datasets where each classifier successfully run (see Table~\ref{table:otherClass}).  These show that CCFs
outperformed all of the other classifiers on all three metrics.  Similarly, we found that for each classifier, the
number of times CCFs outperformed that classifier was larger than the number of times
it performed worse.  

\section{Regression Experiments}
\label{sec:reg}

\begin{figure*}[p]
	\centering
	\begin{subfigure}[t]{0.48\textwidth}
		\caption{Ground Truth}
		\includegraphics[width=0.95\textwidth]{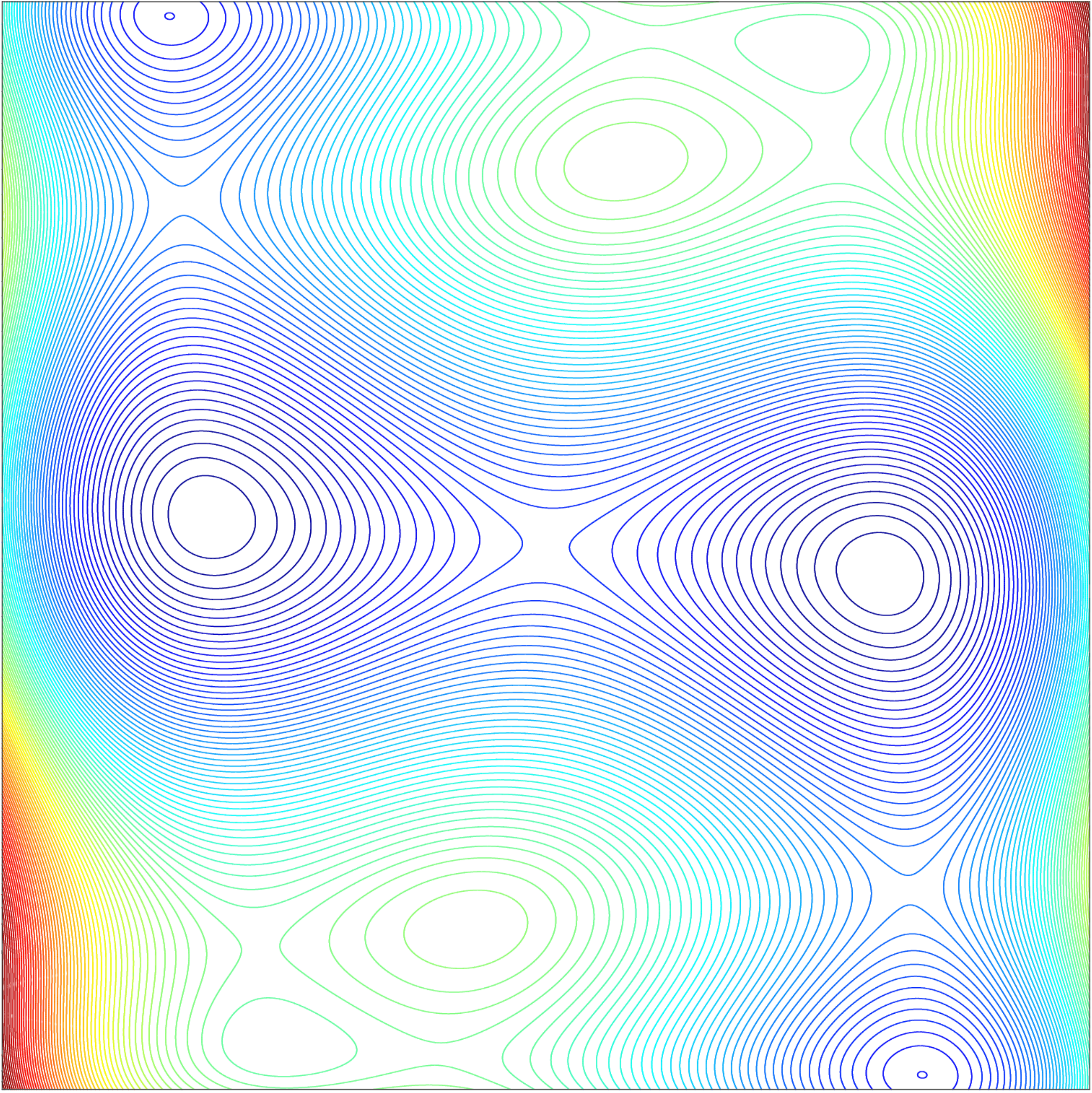}
		\centering
	\end{subfigure}
	\\
	\vspace{20pt}
	\begin{subfigure}[t]{0.48\textwidth}
		\caption{RF prediction}
		\includegraphics[width=0.95\textwidth]{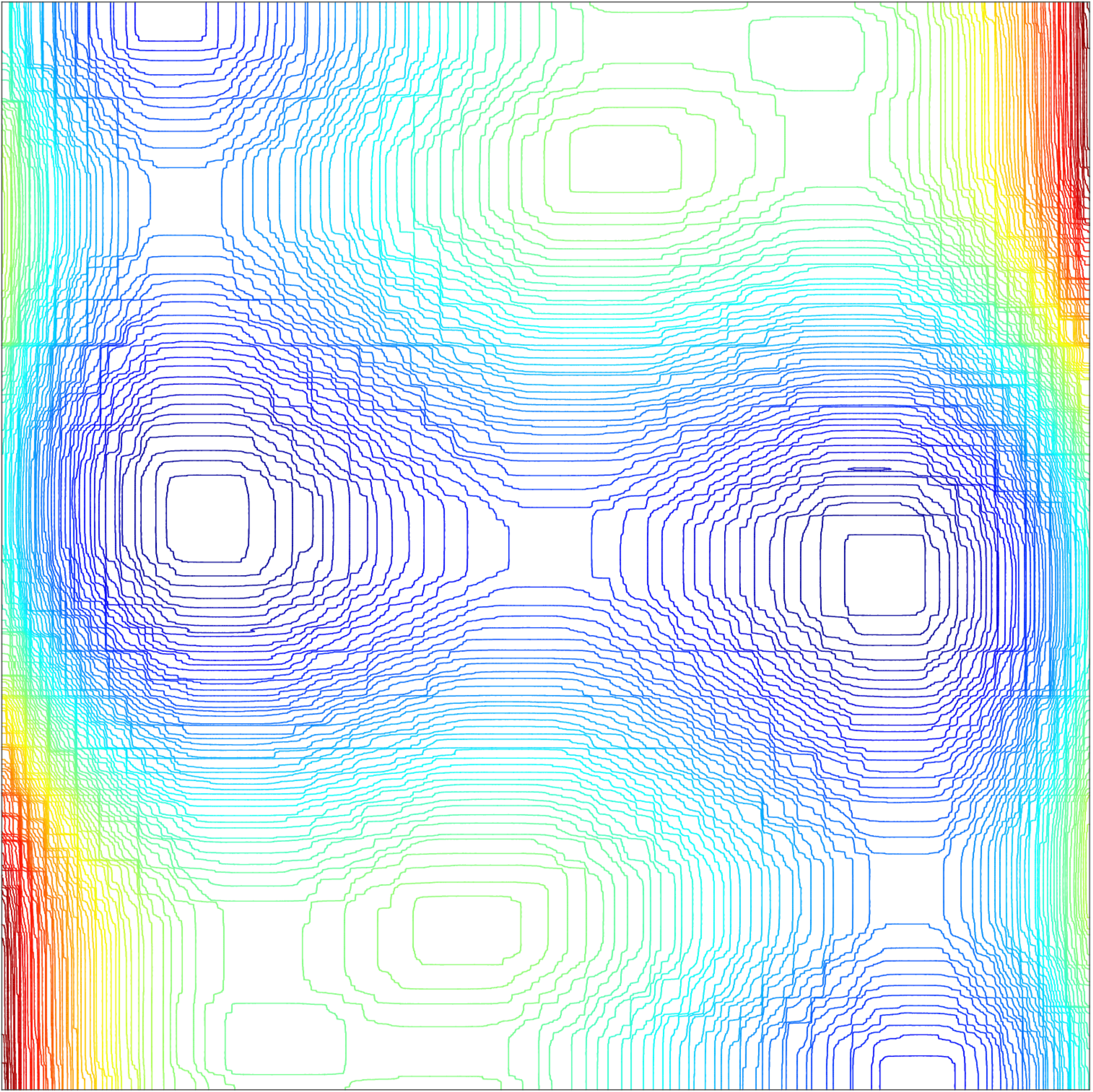}
		\centering
	\end{subfigure}	
	~
	\begin{subfigure}[t]{0.48\textwidth}
		\caption{CCF prediction}
		\includegraphics[width=0.95\textwidth]{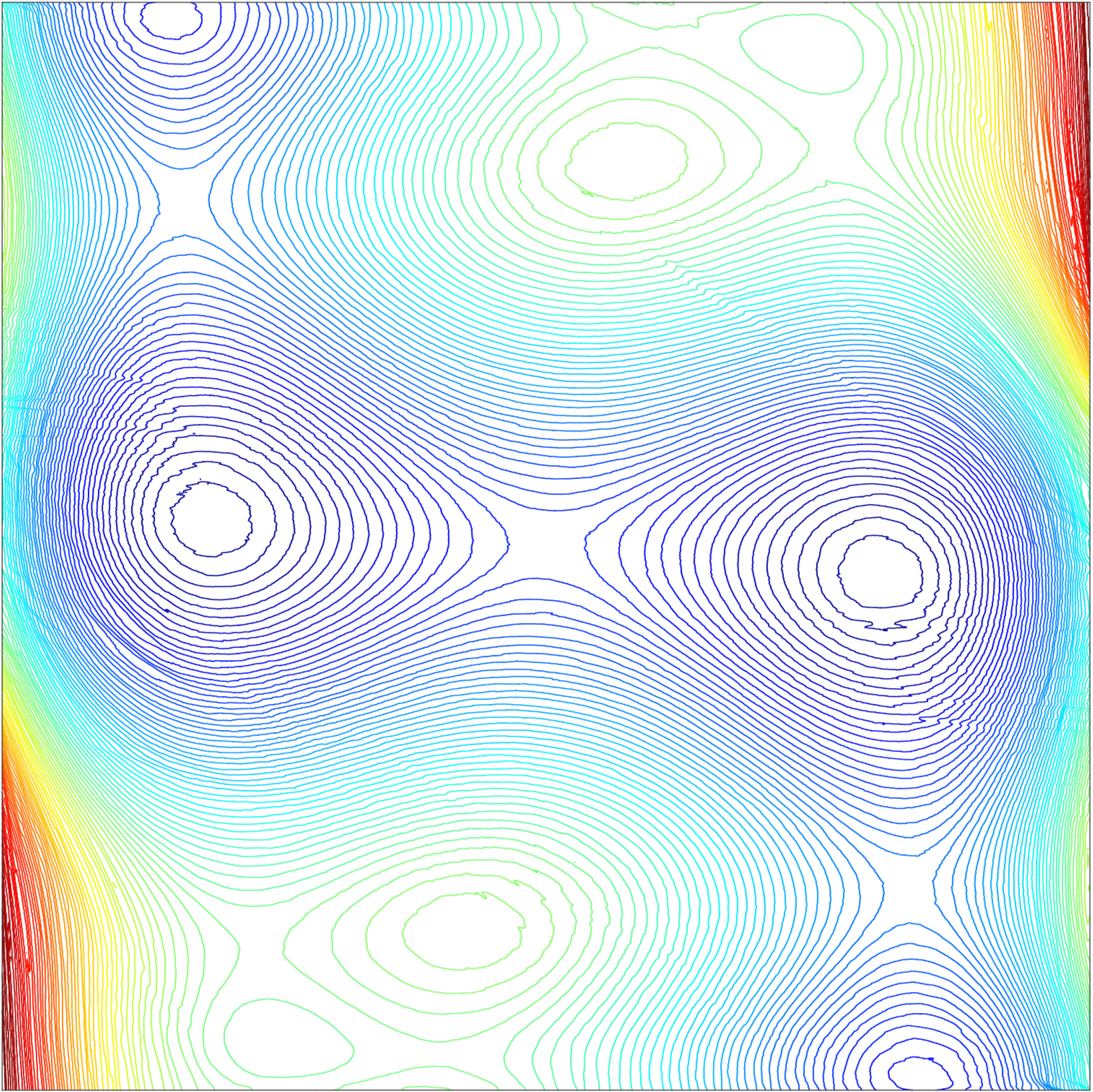}
		\centering
	\end{subfigure}		
	\caption{Contour plots for regression of the six hump camel function~\eqref{eq:camel}
		in the range $x_1 \in [-1.15, 1.15]$, $x_2 \in [-1.75,1.75]$.  a) shows the ground truth, corresponding to a contour plot for a $1000\times1000$ grid.  The training dataset was formed of $20000$ random samples without replacement from these $10^6$ points.  The trained regressor was then used to predict all $10^6$ points in the grid to construct the contour plots shown in b) and c).  Mean squared errors for the $9.8 \times 10^5$ points not used in training were $8.45 \times 10^{-4}$ and $3.40 \times 10^{-4}$ for the RF and CCF respectively.
		\label{fig:camel}}
\end{figure*}

We now investigate the predictive performance of CCFs for regression.  We start with  an illustrative
example to show the benefits of CCFs over axis aligned approaches.  Namely, we consider the six hump
camel function~\citep{molga2005test}
\begin{align}
\label{eq:camel}
f(x_1,x_2) = \left(4-2.1x_2^2+\frac{x_2^4}{3}\right)x_2^2+x_1x_2+(-4+4x_1^2)x_1^2
\end{align}
in the range $x_1 \in [-1.15, 1.15]$, $x_2 \in [-1.75,1.75]$.  Figure~\ref{fig:camel} shows
the contour plots resulting from training a CCF and RF on a random sample of points
in this range.   These show that the RF predictions fail to catch the true
contours of the function, with the RF contours noticeably aligned with the axis to the detriment of the regressor.
The CCF on the other hand is able to accurately capture the regression surface.

We next consider a more rigorous comparison of CCFs to RFs, rotation forests, CCF-Bag, RRFs, and RRF-bag.  Parameters
were again set as per Section~\ref{sec:param}, meaning that we used the mean squared
error criterion and set the minimum number of points at a leaf node to $3$ for all methods.
All other parameters and the test
procedure (i.e. $15$ separate $10-$fold cross-validation tests)
were as per the classification experiments in Section~\ref{sec:treeComparison}. As
in~\cite{pardo2013rotation}, we used the 61 regression datasets from the WEKA dataset
collection\footnote{\scriptsize\url{http://www.cs.waikato.ac.nz/ml/weka/datasets.html}}~\citep{hall2009weka}, 
30 of which were
assembled by Luis Torgo.\footnote{\scriptsize\url{http://www.dcc.fc.up.pt/~ltorgo/Regression/DataSets.html}}
Summary information about each dataset is provided in Table~\ref{table:regdata}, the full results are given in
Table~\ref{table:RegressionTable}, summaries for the significance test results are shown in Table~\ref{table:nVicsReg},
and the mean ranks are given in Table~\ref{table:rank-reg}.  

\begin{table*}[p]                        
	\centering    
	\footnotesize             
	\setlength\tabcolsep{4pt}	  
	\renewcommand{\arraystretch}{1.2}    
	\caption{Dataset summaries for regression experiments.  $N=$ number of data points, $D_c=$ number of non-ordinal features, $D_r=$ number of ordinal features, and \% Miss = percentage of feature values which are missing. \label{table:regdata}}                            
	\begin{tabular}{cccccHHHHHHHHHHH}                    
		\hline     
				Dataset & $N$ & $D_c$ & $D_r$ & \% Miss & CCF & RF & & Rot For & & CCF-Bag & & RRF & & RRF-Bag &\\
				\hline                                                     
2D planes & 40768 & 0 & 10 & 0 & 6.37 $\pm$ 0.12 & \textbf{5.57 $\pm$ 0.10} & $\circ$ & 5.87 $\pm$ 0.11 & $\circ$ & 6.09 $\pm$ 0.11 & $\circ$ & 6.57 $\pm$ 0.12 & $\bullet$ & 6.08 $\pm$ 0.11 & $\circ$ \\                           
Abalone & 4177 & 1 & 7 & 0 & 40.83 $\pm$ 4.91 & 43.88 $\pm$ 5.39 & $\bullet$ & 42.08 $\pm$ 5.11 & $\bullet$ & \textbf{40.72 $\pm$ 4.86} & $\circ$ & 43.62 $\pm$ 5.18 & $\bullet$ & 43.24 $\pm$ 5.16 & $\bullet$ \\                   
Ailerons & 13750 & 0 & 40 & 0 & 17.46 $\pm$ 1.15 & 18.60 $\pm$ 1.24 & $\bullet$ & \textbf{17.23 $\pm$ 1.10} & $\circ$ & 17.48 $\pm$ 1.14 &  & 19.03 $\pm$ 1.30 & $\bullet$ & 19.74 $\pm$ 1.35 & $\bullet$ \\                         
Auto 93 & 93 & 3 & 19 & 0.68 & \textbf{27.31 $\pm$ 32.94} & 34.42 $\pm$ 38.42 & $\bullet$ & 34.92 $\pm$ 34.50 & $\bullet$ & 28.15 $\pm$ 34.09 & $\bullet$ & 30.46 $\pm$ 35.70 & $\bullet$ & 34.41 $\pm$ 39.39 & $\bullet$ \\         
Auto horsepower & 203 & 8 & 17 & 1.04 & \textbf{10.89 $\pm$ 14.89} & 15.37 $\pm$ 18.44 & $\bullet$ & 12.11 $\pm$ 15.11 & $\bullet$ & 11.23 $\pm$ 15.10 & $\bullet$ & 13.21 $\pm$ 16.42 & $\bullet$ & 15.00 $\pm$ 17.55 & $\bullet$ \\
Auto MPG & 398 & 1 & 6 & 0.22 & 11.53 $\pm$ 4.56 & 12.61 $\pm$ 5.44 & $\bullet$ & 11.85 $\pm$ 4.91 & $\bullet$ & 11.68 $\pm$ 4.53 & $\bullet$ & \textbf{11.44 $\pm$ 4.73} &  & 11.89 $\pm$ 4.82 & $\bullet$ \\                       
Auto price & 159 & 0 & 15 & 0 & \textbf{10.91 $\pm$ 7.57} & 12.20 $\pm$ 9.60 & $\bullet$ & 11.83 $\pm$ 9.62 & $\bullet$ & 11.95 $\pm$ 8.57 & $\bullet$ & 12.75 $\pm$ 11.81 & $\bullet$ & 14.59 $\pm$ 13.41 & $\bullet$ \\            
Bank 32NH & 8192 & 0 & 32 & 0 & 48.67 $\pm$ 3.83 & 50.35 $\pm$ 3.94 & $\bullet$ & \textbf{48.61 $\pm$ 3.88} &  & 48.89 $\pm$ 3.82 & $\bullet$ & 52.50 $\pm$ 4.02 & $\bullet$ & 53.59 $\pm$ 4.05 & $\bullet$ \\                       
Bank 8FM & 8192 & 0 & 8 & 0 & \textbf{3.70 $\pm$ 0.25} & 4.19 $\pm$ 0.29 & $\bullet$ & 4.38 $\pm$ 0.27 & $\bullet$ & 3.74 $\pm$ 0.25 & $\bullet$ & 4.58 $\pm$ 0.28 & $\bullet$ & 4.86 $\pm$ 0.30 & $\bullet$ \\                      
Basketball & 96 & 0 & 4 & 0 & 72.77 $\pm$ 31.73 & 69.98 $\pm$ 34.67 & $\circ$ & \textbf{66.38 $\pm$ 30.50} & $\circ$ & 69.77 $\pm$ 31.60 & $\circ$ & 73.88 $\pm$ 31.73 & $\bullet$ & 69.79 $\pm$ 31.75 & $\circ$ \\                  
Body fat & 252 & 0 & 14 & 0 & \textbf{4.73 $\pm$ 3.26} & 5.41 $\pm$ 3.47 & $\bullet$ & 4.95 $\pm$ 3.51 &  & 5.36 $\pm$ 3.34 & $\bullet$ & 10.06 $\pm$ 4.31 & $\bullet$ & 11.55 $\pm$ 4.75 & $\bullet$ \\                             
Bolts & 40 & 0 & 7 & 0 & \textbf{8.62 $\pm$ 8.05} & 10.57 $\pm$ 7.58 & $\bullet$ & 17.45 $\pm$ 13.24 & $\bullet$ & 10.61 $\pm$ 9.66 & $\bullet$ & 22.32 $\pm$ 17.85 & $\bullet$ & 24.61 $\pm$ 19.00 & $\bullet$ \\                   
Breast tumor & 286 & 6 & 3 & 0 & 108.05 $\pm$ 25.03 & 99.96 $\pm$ 23.73 & $\circ$ & \textbf{98.19 $\pm$ 23.15} & $\circ$ & 104.84 $\pm$ 24.13 & $\circ$ & 114.95 $\pm$ 26.53 & $\bullet$ & 105.33 $\pm$ 24.86 & $\circ$ \\           
CA housing & 20640 & 0 & 8 & 0 & 19.46 $\pm$ 1.13 & \textbf{17.55 $\pm$ 1.08} & $\circ$ & 20.29 $\pm$ 1.17 & $\bullet$ & 19.65 $\pm$ 1.13 & $\bullet$ & 21.33 $\pm$ 1.19 & $\bullet$ & 21.90 $\pm$ 1.20 & $\bullet$ \\               
Cholesterol & 303 & 3 & 10 & 0.10 & 99.71 $\pm$ 43.48 & 99.56 $\pm$ 43.08 &  & \textbf{97.62 $\pm$ 41.95} & $\circ$ & 99.25 $\pm$ 43.31 &  & 101.14 $\pm$ 44.32 & $\bullet$ & 98.53 $\pm$ 43.85 & $\circ$ \\                         
Cleveland & 303 & 3 & 10 & 0.10 & 48.15 $\pm$ 14.20 & 49.64 $\pm$ 14.68 & $\bullet$ & 48.31 $\pm$ 15.15 &  & \textbf{47.80 $\pm$ 13.85} & $\circ$ & 48.95 $\pm$ 14.15 & $\bullet$ & 48.71 $\pm$ 13.89 & $\bullet$ \\                 
Cloud & 108 & 2 & 4 & 0 & 22.09 $\pm$ 22.59 & \textbf{20.65 $\pm$ 23.51} & $\circ$ & 26.19 $\pm$ 28.85 & $\bullet$ & 21.65 $\pm$ 23.65 & $\circ$ & 24.08 $\pm$ 27.58 & $\bullet$ & 25.98 $\pm$ 29.25 & $\bullet$ \\                  
CPU & 209 & 1 & 6 & 0 & \textbf{8.07 $\pm$ 20.08} & 13.93 $\pm$ 28.85 & $\bullet$ & 12.95 $\pm$ 42.08 &  & 10.28 $\pm$ 22.11 & $\bullet$ & 12.06 $\pm$ 26.98 & $\bullet$ & 17.20 $\pm$ 32.16 & $\bullet$ \\                          
CPU act & 8192 & 0 & 21 & 0 & \textbf{1.66 $\pm$ 0.14} & 1.75 $\pm$ 0.19 & $\bullet$ & 1.77 $\pm$ 0.23 & $\bullet$ & 1.71 $\pm$ 0.15 & $\bullet$ & 2.63 $\pm$ 0.43 & $\bullet$ & 2.97 $\pm$ 0.52 & $\bullet$ \\                      
CPU small & 8192 & 0 & 12 & 0 & \textbf{2.25 $\pm$ 0.19} & 2.30 $\pm$ 0.24 & $\bullet$ & 2.45 $\pm$ 0.26 & $\bullet$ & 2.29 $\pm$ 0.19 & $\bullet$ & 2.45 $\pm$ 0.22 & $\bullet$ & 2.59 $\pm$ 0.24 & $\bullet$ \\                    
Delta ailerons & 7129 & 0 & 5 & 0 & 28.70 $\pm$ 2.44 & \textbf{28.30 $\pm$ 2.64} & $\circ$ & 30.31 $\pm$ 3.00 & $\bullet$ & 28.53 $\pm$ 2.45 & $\circ$ & 28.99 $\pm$ 2.64 & $\bullet$ & 28.76 $\pm$ 2.67 &  \\                       
Delta elevators & 9517 & 0 & 6 & 0 & 36.40 $\pm$ 2.20 & 36.74 $\pm$ 2.28 & $\bullet$ & \textbf{35.41 $\pm$ 2.24} & $\circ$ & 36.04 $\pm$ 2.19 & $\circ$ & 36.75 $\pm$ 2.25 & $\bullet$ & 36.01 $\pm$ 2.23 & $\circ$ \\               
Detroit & 13 & 0 & 13 & 0 & 39.20 $\pm$ 78.01 & 43.25 $\pm$ 75.49 & $\bullet$ & 102.49 $\pm$ 125.62 & $\bullet$ & 44.25 $\pm$ 90.56 & $\bullet$ & \textbf{35.72 $\pm$ 68.25} & $\circ$ & 41.68 $\pm$ 82.03 & $\bullet$ \\            
Diabetes & 43 & 0 & 2 & 0 & 70.84 $\pm$ 45.65 & \textbf{64.93 $\pm$ 39.75} & $\circ$ & 87.56 $\pm$ 54.48 & $\bullet$ & 71.01 $\pm$ 45.74 &  & 67.54 $\pm$ 43.10 & $\circ$ & 67.77 $\pm$ 43.06 & $\circ$ \\                           
Echocardiogram & 130 & 0 & 9 & 8.29 & 56.94 $\pm$ 19.42 & 54.11 $\pm$ 18.47 & $\circ$ & \textbf{52.09 $\pm$ 17.91} & $\circ$ & 56.53 $\pm$ 18.79 &  & 58.00 $\pm$ 19.32 & $\bullet$ & 56.82 $\pm$ 18.33 &  \\                        
Elevators & 16599 & 0 & 18 & 0 & \textbf{10.34 $\pm$ 1.01} & 18.39 $\pm$ 1.76 & $\bullet$ & 14.81 $\pm$ 1.45 & $\bullet$ & 10.46 $\pm$ 1.03 & $\bullet$ & 20.03 $\pm$ 2.06 & $\bullet$ & 21.57 $\pm$ 2.22 & $\bullet$ \\             
Electricity usage & 55 & 0 & 2 & 0 & 21.29 $\pm$ 13.25 & 22.25 $\pm$ 14.09 &  & 28.82 $\pm$ 22.79 & $\bullet$ & \textbf{21.23 $\pm$ 13.34} &  & 23.54 $\pm$ 16.33 &  & 23.55 $\pm$ 16.40 &  \\                                       
Fish catch & 158 & 1 & 6 & 7.87 & \textbf{2.17 $\pm$ 2.35} & 6.15 $\pm$ 7.20 & $\bullet$ & 5.40 $\pm$ 10.14 & $\bullet$ & 2.55 $\pm$ 2.66 & $\bullet$ & 3.24 $\pm$ 3.33 & $\bullet$ & 4.59 $\pm$ 4.83 & $\bullet$ \\                 
Friedman & 40768 & 0 & 10 & 0 & 9.15 $\pm$ 0.19 & \textbf{6.60 $\pm$ 0.17} & $\circ$ & 7.36 $\pm$ 0.21 & $\circ$ & 9.28 $\pm$ 0.20 & $\bullet$ & 8.66 $\pm$ 0.22 & $\circ$ & 9.40 $\pm$ 0.24 & $\bullet$ \\                          
Fruit fly & 125 & 1 & 3 & 0 & 128.60 $\pm$ 66.66 & 124.00 $\pm$ 67.16 & $\circ$ & \textbf{107.97 $\pm$ 65.43} & $\circ$ & 123.39 $\pm$ 65.91 & $\circ$ & 136.63 $\pm$ 68.94 & $\bullet$ & 126.15 $\pm$ 66.98 & $\circ$ \\  
Gas consumption & 27 & 0 & 4 & 0 & \textbf{4.02 $\pm$ 4.68} & 6.71 $\pm$ 7.05 & $\bullet$ & 10.08 $\pm$ 6.80 & $\bullet$ & 4.69 $\pm$ 6.90 &  & 5.83 $\pm$ 5.37 & $\bullet$ & 5.81 $\pm$ 7.25 & $\bullet$ \\     
		\hline 
	\end{tabular}      
	\hspace{20pt}
	\begin{tabular}{cccccHHHHHHHHHHH}   
		\hline     
				Dataset & $N$ & $D_c$ & $D_r$ & \% Miss & CCF & RF & & Rot For & & CCF-Bag & & RRF & & RRF-Bag &\\
				\hline                                                                         
House 16H & 22784 & 0 & 16 & 0 & 36.57 $\pm$ 4.47 & \textbf{35.74 $\pm$ 4.49} & $\circ$ & 40.51 $\pm$ 4.88 & $\bullet$ & 37.04 $\pm$ 4.42 & $\bullet$ & 40.30 $\pm$ 4.68 & $\bullet$ & 41.98 $\pm$ 4.83 & $\bullet$ \\               
House 8L & 22784 & 0 & 8 & 0 & \textbf{29.96 $\pm$ 3.11} & 30.53 $\pm$ 3.13 & $\bullet$ & 32.94 $\pm$ 3.41 & $\bullet$ & 30.08 $\pm$ 3.10 & $\bullet$ & 31.89 $\pm$ 3.26 & $\bullet$ & 32.28 $\pm$ 3.29 & $\bullet$ \\               
Housing & 506 & 0 & 13 & 0 & \textbf{12.89 $\pm$ 8.47} & 13.51 $\pm$ 9.04 & $\bullet$ & 14.86 $\pm$ 9.56 & $\bullet$ & 13.49 $\pm$ 8.68 & $\bullet$ & 13.85 $\pm$ 8.93 & $\bullet$ & 15.42 $\pm$ 9.63 & $\bullet$ \\                 
Hungarian & 294 & 3 & 10 & 13.47 & 50.94 $\pm$ 14.55 & 55.42 $\pm$ 15.34 & $\bullet$ & 53.94 $\pm$ 16.01 & $\bullet$ & \textbf{50.22 $\pm$ 14.54} & $\circ$ & 51.04 $\pm$ 14.30 &  & 51.06 $\pm$ 13.76 &  \\                         
Kinematics & 8192 & 0 & 8 & 0 & 20.33 $\pm$ 1.05 & 27.97 $\pm$ 1.46 & $\bullet$ & 21.33 $\pm$ 1.16 & $\bullet$ & 20.92 $\pm$ 1.09 & $\bullet$ & \textbf{18.81 $\pm$ 1.00} & $\circ$ & 20.98 $\pm$ 1.10 & $\bullet$ \\                
Longley & 16 & 0 & 6 & 0 & \textbf{6.61 $\pm$ 7.56} & 14.75 $\pm$ 12.00 & $\bullet$ & 16.67 $\pm$ 17.19 & $\bullet$ & 8.77 $\pm$ 9.75 & $\bullet$ & 9.57 $\pm$ 8.02 & $\bullet$ & 14.32 $\pm$ 10.72 & $\bullet$ \\                   
Low birth weight & 189 & 1 & 8 & 0 & 40.09 $\pm$ 13.68 & 39.63 $\pm$ 12.92 &  & \textbf{37.97 $\pm$ 13.03} & $\circ$ & 38.89 $\pm$ 13.27 & $\circ$ & 45.71 $\pm$ 16.38 & $\bullet$ & 43.58 $\pm$ 15.95 & $\bullet$ \\                
Machine CPU & 209 & 0 & 6 & 0 & \textbf{12.80 $\pm$ 23.68} & 16.69 $\pm$ 30.55 & $\bullet$ & 23.39 $\pm$ 37.67 & $\bullet$ & 14.79 $\pm$ 24.89 & $\bullet$ & 13.10 $\pm$ 26.48 &  & 17.53 $\pm$ 30.16 & $\bullet$ \\                 
MBA grade & 61 & 0 & 2 & 0 & 106.10 $\pm$ 69.36 & 106.04 $\pm$ 68.30 &  & \textbf{91.00 $\pm$ 66.94} & $\circ$ & 106.95 $\pm$ 69.62 & $\bullet$ & 106.49 $\pm$ 68.01 &  & 107.25 $\pm$ 68.51 &  \\                                   
Meta & 528 & 2 & 19 & 4.55 & \textbf{93.29 $\pm$ 159.35} & 95.34 $\pm$ 165.36 & $\bullet$ & 100.27 $\pm$ 185.32 &  & 95.28 $\pm$ 165.56 &  & 97.12 $\pm$ 175.56 &  & 99.64 $\pm$ 184.72 &  \\                                        
MV & 40768 & 3 & 7 & 0 & \textbf{0.00 $\pm$ 0.00} & 0.02 $\pm$ 0.01 & $\bullet$ & 0.02 $\pm$ 0.00 & $\bullet$ & 0.00 $\pm$ 0.00 & $\bullet$ & 0.29 $\pm$ 0.01 & $\bullet$ & 0.36 $\pm$ 0.02 & $\bullet$ \\                           
PBC & 418 & 1 & 17 & 16.47 & 62.81 $\pm$ 12.95 & 65.11 $\pm$ 13.09 & $\bullet$ & \textbf{60.50 $\pm$ 13.39} & $\circ$ & 62.45 $\pm$ 12.79 & $\circ$ & 64.24 $\pm$ 13.04 & $\bullet$ & 65.00 $\pm$ 13.14 & $\bullet$ \\               
Pharynx & 195 & 2 & 10 & 0.09 & 36.35 $\pm$ 16.41 & 37.24 $\pm$ 17.90 & $\bullet$ & \textbf{32.47 $\pm$ 17.89} & $\circ$ & 37.83 $\pm$ 16.34 & $\bullet$ & 46.21 $\pm$ 19.35 & $\bullet$ & 49.02 $\pm$ 19.39 & $\bullet$ \\          
Pole & 15000 & 0 & 48 & 0 & 2.43 $\pm$ 0.21 & 2.78 $\pm$ 0.22 & $\bullet$ & \textbf{1.27 $\pm$ 0.12} & $\circ$ & 2.65 $\pm$ 0.23 & $\bullet$ & 2.25 $\pm$ 0.21 & $\circ$ & 2.75 $\pm$ 0.23 & $\bullet$ \\                            
Pollution & 60 & 0 & 15 & 0 & \textbf{43.35 $\pm$ 30.47} & 48.87 $\pm$ 34.14 & $\bullet$ & 47.37 $\pm$ 31.65 & $\bullet$ & 43.75 $\pm$ 30.44 &  & 47.81 $\pm$ 32.00 & $\bullet$ & 50.93 $\pm$ 34.02 & $\bullet$ \\                   
Puma 32H & 8192 & 0 & 32 & 0 & 65.10 $\pm$ 3.48 & 26.07 $\pm$ 1.76 & $\circ$ & \textbf{14.59 $\pm$ 0.93} & $\circ$ & 67.43 $\pm$ 3.53 & $\bullet$ & 66.53 $\pm$ 3.19 & $\bullet$ & 69.08 $\pm$ 3.30 & $\bullet$ \\                   
Puma 8NH & 8192 & 0 & 8 & 0 & 32.68 $\pm$ 1.77 & 32.64 $\pm$ 1.81 &  & 33.18 $\pm$ 1.78 & $\bullet$ & \textbf{32.60 $\pm$ 1.76} & $\circ$ & 34.93 $\pm$ 1.86 & $\bullet$ & 35.59 $\pm$ 1.83 & $\bullet$ \\                           
PW Linear & 200 & 0 & 10 & 0 & 15.33 $\pm$ 5.61 & 16.66 $\pm$ 6.09 & $\bullet$ & \textbf{15.19 $\pm$ 5.55} &  & 15.86 $\pm$ 5.83 & $\bullet$ & 21.26 $\pm$ 8.34 & $\bullet$ & 24.02 $\pm$ 9.18 & $\bullet$ \\                        
Pyrimidines & 74 & 0 & 27 & 0 & 58.21 $\pm$ 113.22 & 63.96 $\pm$ 125.51 & $\bullet$ & 69.33 $\pm$ 115.62 & $\bullet$ & 59.33 $\pm$ 115.76 & $\bullet$ & \textbf{57.73 $\pm$ 113.21} &  & 60.47 $\pm$ 112.85 & $\bullet$ \\           
Quake & 2178 & 0 & 3 & 0 & 102.42 $\pm$ 11.28 & 105.75 $\pm$ 11.56 & $\bullet$ & \textbf{99.75 $\pm$ 11.67} & $\circ$ & 102.45 $\pm$ 11.31 &  & 106.23 $\pm$ 11.60 & $\bullet$ & 106.25 $\pm$ 11.56 & $\bullet$ \\                   
SCHL vote & 37 & 0 & 5 & 0 & 78.31 $\pm$ 141.49 & \textbf{58.17 $\pm$ 125.34} & $\circ$ & 93.58 $\pm$ 179.95 &  & 76.87 $\pm$ 140.83 & $\circ$ & 75.34 $\pm$ 137.82 & $\circ$ & 79.09 $\pm$ 147.01 &  \\                             
Sensory & 576 & 0 & 11 & 0 & 77.27 $\pm$ 12.74 & \textbf{70.44 $\pm$ 11.72} & $\circ$ & 71.99 $\pm$ 11.75 & $\circ$ & 77.80 $\pm$ 12.83 & $\bullet$ & 77.44 $\pm$ 12.77 &  & 78.74 $\pm$ 13.20 & $\bullet$ \\                        
Servo & 167 & 4 & 0 & 0 & \textbf{10.93 $\pm$ 14.87} & 13.85 $\pm$ 18.54 & $\bullet$ & 14.75 $\pm$ 20.91 & $\bullet$ & 12.34 $\pm$ 14.96 & $\bullet$ & 18.54 $\pm$ 19.51 & $\bullet$ & 23.10 $\pm$ 20.77 & $\bullet$ \\              
Sleep & 58 & 0 & 7 & 1.97 & 48.24 $\pm$ 23.37 & 56.58 $\pm$ 24.68 & $\bullet$ & 53.74 $\pm$ 25.09 & $\bullet$ & \textbf{48.00 $\pm$ 22.71} &  & 48.21 $\pm$ 24.02 &  & 48.82 $\pm$ 22.74 &  \\                                       
Stock & 950 & 0 & 9 & 0 & 1.11 $\pm$ 0.22 & 1.35 $\pm$ 0.34 & $\bullet$ & 1.27 $\pm$ 0.26 & $\bullet$ & 1.25 $\pm$ 0.25 & $\bullet$ & \textbf{1.06 $\pm$ 0.23} & $\circ$ & 1.27 $\pm$ 0.26 & $\bullet$ \\                            
Strike & 625 & 1 & 5 & 0 & 87.35 $\pm$ 100.00 & \textbf{84.22 $\pm$ 99.55} & $\circ$ & 90.00 $\pm$ 102.21 & $\bullet$ & 87.84 $\pm$ 99.23 &  & 87.19 $\pm$ 100.97 &  & 85.03 $\pm$ 100.80 & $\circ$ \\                               
Triazines & 186 & 0 & 60 & 0 & 71.99 $\pm$ 36.69 & 71.01 $\pm$ 35.37 &  & \textbf{70.25 $\pm$ 34.49} &  & 71.56 $\pm$ 36.57 &  & 72.44 $\pm$ 36.78 &  & 72.87 $\pm$ 37.27 & $\bullet$ \\                                             
Veteran & 137 & 1 & 6 & 0 & 80.39 $\pm$ 77.45 & 88.10 $\pm$ 88.31 & $\bullet$ & 84.10 $\pm$ 89.71 &  & 78.38 $\pm$ 73.80 & $\circ$ & 79.03 $\pm$ 78.11 & $\circ$ & \textbf{78.35 $\pm$ 77.85} & $\circ$ \\                           
Vineyard & 52 & 0 & 3 & 0 & 38.41 $\pm$ 29.36 & 39.40 $\pm$ 36.17 &  & 46.64 $\pm$ 32.90 & $\bullet$ & \textbf{38.31 $\pm$ 28.94} &  & 38.73 $\pm$ 33.31 &  & 38.68 $\pm$ 33.11 &  \\                                                
Wisconsin & 194 & 0 & 32 & 0 & 87.82 $\pm$ 24.18 & 92.91 $\pm$ 24.91 & $\bullet$ & 90.38 $\pm$ 24.95 & $\bullet$ & \textbf{87.33 $\pm$ 23.96} & $\circ$ & 89.87 $\pm$ 25.33 & $\bullet$ & 88.52 $\pm$ 24.80 &  \\  &&&&&&&&&&&&&&&     \\      
		\hline                                         
	\end{tabular}                                                          
\end{table*}

\begin{table*}[p]                        
	\centering    
	\scriptsize             
	\setlength\tabcolsep{3pt}	  
	\renewcommand{\arraystretch}{1}    
			\caption{Mean squared error as a percentage of the data variance.  Conventions as
				 per Table~\ref{table:ResultsTable}. \label{table:RegressionTable}}   \begin{tabular}{|c|HHHHc|cc|cc|cc|cc|cc|}                  
		\hline                                                       
		Dataset & $N$ & $D_c$ & $D_r$ & \% Miss & CCF & RF & & Rot For & & CCF-Bag & & RRF & & RRF-Bag &\\
		\hline           
2D planes & 40768 & 0 & 10 & 0 & 6.37 $\pm$ 0.12 & \textbf{5.57 $\pm$ 0.10} & $\circ$ & 5.87 $\pm$ 0.11 & $\circ$ & 6.09 $\pm$ 0.11 & $\circ$ & 6.57 $\pm$ 0.12 & $\bullet$ & 6.08 $\pm$ 0.11 & $\circ$ \\                           
Abalone & 4177 & 1 & 7 & 0 & 40.83 $\pm$ 4.91 & 43.88 $\pm$ 5.39 & $\bullet$ & 42.08 $\pm$ 5.11 & $\bullet$ & \textbf{40.72 $\pm$ 4.86} & $\circ$ & 43.62 $\pm$ 5.18 & $\bullet$ & 43.24 $\pm$ 5.16 & $\bullet$ \\                   
Ailerons & 13750 & 0 & 40 & 0 & 17.46 $\pm$ 1.15 & 18.60 $\pm$ 1.24 & $\bullet$ & \textbf{17.23 $\pm$ 1.10} & $\circ$ & 17.48 $\pm$ 1.14 &  & 19.03 $\pm$ 1.30 & $\bullet$ & 19.74 $\pm$ 1.35 & $\bullet$ \\                         
Auto 93 & 93 & 3 & 19 & 0.68 & \textbf{27.31 $\pm$ 32.94} & 34.42 $\pm$ 38.42 & $\bullet$ & 34.92 $\pm$ 34.50 & $\bullet$ & 28.15 $\pm$ 34.09 & $\bullet$ & 30.46 $\pm$ 35.70 & $\bullet$ & 34.41 $\pm$ 39.39 & $\bullet$ \\         
Auto horsepower & 203 & 8 & 17 & 1.04 & \textbf{10.89 $\pm$ 14.89} & 15.37 $\pm$ 18.44 & $\bullet$ & 12.11 $\pm$ 15.11 & $\bullet$ & 11.23 $\pm$ 15.10 & $\bullet$ & 13.21 $\pm$ 16.42 & $\bullet$ & 15.00 $\pm$ 17.55 & $\bullet$ \\
Auto MPG & 398 & 1 & 6 & 0.22 & 11.53 $\pm$ 4.56 & 12.61 $\pm$ 5.44 & $\bullet$ & 11.85 $\pm$ 4.91 & $\bullet$ & 11.68 $\pm$ 4.53 & $\bullet$ & \textbf{11.44 $\pm$ 4.73} &  & 11.89 $\pm$ 4.82 & $\bullet$ \\                       
Auto price & 159 & 0 & 15 & 0 & \textbf{10.91 $\pm$ 7.57} & 12.20 $\pm$ 9.60 & $\bullet$ & 11.83 $\pm$ 9.62 & $\bullet$ & 11.95 $\pm$ 8.57 & $\bullet$ & 12.75 $\pm$ 11.81 & $\bullet$ & 14.59 $\pm$ 13.41 & $\bullet$ \\            
Bank 32NH & 8192 & 0 & 32 & 0 & 48.67 $\pm$ 3.83 & 50.35 $\pm$ 3.94 & $\bullet$ & \textbf{48.61 $\pm$ 3.88} &  & 48.89 $\pm$ 3.82 & $\bullet$ & 52.50 $\pm$ 4.02 & $\bullet$ & 53.59 $\pm$ 4.05 & $\bullet$ \\                       
Bank 8FM & 8192 & 0 & 8 & 0 & \textbf{3.70 $\pm$ 0.25} & 4.19 $\pm$ 0.29 & $\bullet$ & 4.38 $\pm$ 0.27 & $\bullet$ & 3.74 $\pm$ 0.25 & $\bullet$ & 4.58 $\pm$ 0.28 & $\bullet$ & 4.86 $\pm$ 0.30 & $\bullet$ \\                      
Basketball & 96 & 0 & 4 & 0 & 72.77 $\pm$ 31.73 & 69.98 $\pm$ 34.67 & $\circ$ & \textbf{66.38 $\pm$ 30.50} & $\circ$ & 69.77 $\pm$ 31.60 & $\circ$ & 73.88 $\pm$ 31.73 & $\bullet$ & 69.79 $\pm$ 31.75 & $\circ$ \\                  
Body fat & 252 & 0 & 14 & 0 & \textbf{4.73 $\pm$ 3.26} & 5.41 $\pm$ 3.47 & $\bullet$ & 4.95 $\pm$ 3.51 &  & 5.36 $\pm$ 3.34 & $\bullet$ & 10.06 $\pm$ 4.31 & $\bullet$ & 11.55 $\pm$ 4.75 & $\bullet$ \\                             
Bolts & 40 & 0 & 7 & 0 & \textbf{8.62 $\pm$ 8.05} & 10.57 $\pm$ 7.58 & $\bullet$ & 17.45 $\pm$ 13.24 & $\bullet$ & 10.61 $\pm$ 9.66 & $\bullet$ & 22.32 $\pm$ 17.85 & $\bullet$ & 24.61 $\pm$ 19.00 & $\bullet$ \\                   
Breast tumor & 286 & 6 & 3 & 0 & 108.1 $\pm$ 25.03 & 99.96 $\pm$ 23.73 & $\circ$ & \textbf{98.19 $\pm$ 23.15} & $\circ$ & 104.8 $\pm$ 24.13 & $\circ$ & 115.0 $\pm$ 26.53 & $\bullet$ & 105.3 $\pm$ 24.86 & $\circ$ \\               
CA housing & 20640 & 0 & 8 & 0 & 19.46 $\pm$ 1.13 & \textbf{17.55 $\pm$ 1.08} & $\circ$ & 20.29 $\pm$ 1.17 & $\bullet$ & 19.65 $\pm$ 1.13 & $\bullet$ & 21.33 $\pm$ 1.19 & $\bullet$ & 21.90 $\pm$ 1.20 & $\bullet$ \\               
Cholesterol & 303 & 3 & 10 & 0.10 & 99.71 $\pm$ 43.48 & 99.56 $\pm$ 43.08 &  & \textbf{97.62 $\pm$ 41.95} & $\circ$ & 99.25 $\pm$ 43.31 &  & 101.1 $\pm$ 44.32 & $\bullet$ & 98.53 $\pm$ 43.85 & $\circ$ \\                          
Cleveland & 303 & 3 & 10 & 0.10 & 48.15 $\pm$ 14.20 & 49.64 $\pm$ 14.68 & $\bullet$ & 48.31 $\pm$ 15.15 &  & \textbf{47.80 $\pm$ 13.85} & $\circ$ & 48.95 $\pm$ 14.15 & $\bullet$ & 48.71 $\pm$ 13.89 & $\bullet$ \\                 
Cloud & 108 & 2 & 4 & 0 & 22.09 $\pm$ 22.59 & \textbf{20.65 $\pm$ 23.51} & $\circ$ & 26.19 $\pm$ 28.85 & $\bullet$ & 21.65 $\pm$ 23.65 & $\circ$ & 24.08 $\pm$ 27.58 & $\bullet$ & 25.98 $\pm$ 29.25 & $\bullet$ \\                  
CPU & 209 & 1 & 6 & 0 & \textbf{8.07 $\pm$ 20.08} & 13.93 $\pm$ 28.85 & $\bullet$ & 12.95 $\pm$ 42.08 &  & 10.28 $\pm$ 22.11 & $\bullet$ & 12.06 $\pm$ 26.98 & $\bullet$ & 17.20 $\pm$ 32.16 & $\bullet$ \\                          
CPU act & 8192 & 0 & 21 & 0 & \textbf{1.66 $\pm$ 0.14} & 1.75 $\pm$ 0.19 & $\bullet$ & 1.77 $\pm$ 0.23 & $\bullet$ & 1.71 $\pm$ 0.15 & $\bullet$ & 2.63 $\pm$ 0.43 & $\bullet$ & 2.97 $\pm$ 0.52 & $\bullet$ \\                      
CPU small & 8192 & 0 & 12 & 0 & \textbf{2.25 $\pm$ 0.19} & 2.30 $\pm$ 0.24 & $\bullet$ & 2.45 $\pm$ 0.26 & $\bullet$ & 2.29 $\pm$ 0.19 & $\bullet$ & 2.45 $\pm$ 0.22 & $\bullet$ & 2.59 $\pm$ 0.24 & $\bullet$ \\                    
Delta ailerons & 7129 & 0 & 5 & 0 & 28.70 $\pm$ 2.44 & \textbf{28.30 $\pm$ 2.64} & $\circ$ & 30.31 $\pm$ 3.00 & $\bullet$ & 28.53 $\pm$ 2.45 & $\circ$ & 28.99 $\pm$ 2.64 & $\bullet$ & 28.76 $\pm$ 2.67 &  \\                       
Delta elevators & 9517 & 0 & 6 & 0 & 36.40 $\pm$ 2.20 & 36.74 $\pm$ 2.28 & $\bullet$ & \textbf{35.41 $\pm$ 2.24} & $\circ$ & 36.04 $\pm$ 2.19 & $\circ$ & 36.75 $\pm$ 2.25 & $\bullet$ & 36.01 $\pm$ 2.23 & $\circ$ \\               
Detroit & 13 & 0 & 13 & 0 & 39.20 $\pm$ 78.01 & 43.25 $\pm$ 75.49 & $\bullet$ & 102.5 $\pm$ 125.6 & $\bullet$ & 44.25 $\pm$ 90.56 & $\bullet$ & \textbf{35.72 $\pm$ 68.25} & $\circ$ & 41.68 $\pm$ 82.03 & $\bullet$ \\              
Diabetes & 43 & 0 & 2 & 0 & 70.84 $\pm$ 45.65 & \textbf{64.93 $\pm$ 39.75} & $\circ$ & 87.56 $\pm$ 54.48 & $\bullet$ & 71.01 $\pm$ 45.74 &  & 67.54 $\pm$ 43.10 & $\circ$ & 67.77 $\pm$ 43.06 & $\circ$ \\                           
Echocardiogram & 130 & 0 & 9 & 8.29 & 56.94 $\pm$ 19.42 & 54.11 $\pm$ 18.47 & $\circ$ & \textbf{52.09 $\pm$ 17.91} & $\circ$ & 56.53 $\pm$ 18.79 &  & 58.00 $\pm$ 19.32 & $\bullet$ & 56.82 $\pm$ 18.33 &  \\                        
Elevators & 16599 & 0 & 18 & 0 & \textbf{10.34 $\pm$ 1.01} & 18.39 $\pm$ 1.76 & $\bullet$ & 14.81 $\pm$ 1.45 & $\bullet$ & 10.46 $\pm$ 1.03 & $\bullet$ & 20.03 $\pm$ 2.06 & $\bullet$ & 21.57 $\pm$ 2.22 & $\bullet$ \\             
Electricity usage & 55 & 0 & 2 & 0 & 21.29 $\pm$ 13.25 & 22.25 $\pm$ 14.09 &  & 28.82 $\pm$ 22.79 & $\bullet$ & \textbf{21.23 $\pm$ 13.34} &  & 23.54 $\pm$ 16.33 &  & 23.55 $\pm$ 16.40 &  \\                                       
Fish catch & 158 & 1 & 6 & 7.87 & \textbf{2.17 $\pm$ 2.35} & 6.15 $\pm$ 7.20 & $\bullet$ & 5.40 $\pm$ 10.14 & $\bullet$ & 2.55 $\pm$ 2.66 & $\bullet$ & 3.24 $\pm$ 3.33 & $\bullet$ & 4.59 $\pm$ 4.83 & $\bullet$ \\                 
Friedman & 40768 & 0 & 10 & 0 & 9.15 $\pm$ 0.19 & \textbf{6.60 $\pm$ 0.17} & $\circ$ & 7.36 $\pm$ 0.21 & $\circ$ & 9.28 $\pm$ 0.20 & $\bullet$ & 8.66 $\pm$ 0.22 & $\circ$ & 9.40 $\pm$ 0.24 & $\bullet$ \\                          
Fruit fly & 125 & 1 & 3 & 0 & 128.6 $\pm$ 66.66 & 124.0 $\pm$ 67.16 & $\circ$ & \textbf{108.0 $\pm$ 65.43} & $\circ$ & 123.4 $\pm$ 65.91 & $\circ$ & 136.6 $\pm$ 68.94 & $\bullet$ & 126.2 $\pm$ 66.98 & $\circ$ \\                  
Gas consumption & 27 & 0 & 4 & 0 & \textbf{4.02 $\pm$ 4.68} & 6.71 $\pm$ 7.05 & $\bullet$ & 10.08 $\pm$ 6.80 & $\bullet$ & 4.69 $\pm$ 6.90 &  & 5.83 $\pm$ 5.37 & $\bullet$ & 5.81 $\pm$ 7.25 & $\bullet$ \\                         
House 16H & 22784 & 0 & 16 & 0 & 36.57 $\pm$ 4.47 & \textbf{35.74 $\pm$ 4.49} & $\circ$ & 40.51 $\pm$ 4.88 & $\bullet$ & 37.04 $\pm$ 4.42 & $\bullet$ & 40.30 $\pm$ 4.68 & $\bullet$ & 41.98 $\pm$ 4.83 & $\bullet$ \\               
House 8L & 22784 & 0 & 8 & 0 & \textbf{29.96 $\pm$ 3.11} & 30.53 $\pm$ 3.13 & $\bullet$ & 32.94 $\pm$ 3.41 & $\bullet$ & 30.08 $\pm$ 3.10 & $\bullet$ & 31.89 $\pm$ 3.26 & $\bullet$ & 32.28 $\pm$ 3.29 & $\bullet$ \\               
Housing & 506 & 0 & 13 & 0 & \textbf{12.89 $\pm$ 8.47} & 13.51 $\pm$ 9.04 & $\bullet$ & 14.86 $\pm$ 9.56 & $\bullet$ & 13.49 $\pm$ 8.68 & $\bullet$ & 13.85 $\pm$ 8.93 & $\bullet$ & 15.42 $\pm$ 9.63 & $\bullet$ \\                 
Hungarian & 294 & 3 & 10 & 13.47 & 50.94 $\pm$ 14.55 & 55.42 $\pm$ 15.34 & $\bullet$ & 53.94 $\pm$ 16.01 & $\bullet$ & \textbf{50.22 $\pm$ 14.54} & $\circ$ & 51.04 $\pm$ 14.30 &  & 51.06 $\pm$ 13.76 &  \\                         
Kinematics & 8192 & 0 & 8 & 0 & 20.33 $\pm$ 1.05 & 27.97 $\pm$ 1.46 & $\bullet$ & 21.33 $\pm$ 1.16 & $\bullet$ & 20.92 $\pm$ 1.09 & $\bullet$ & \textbf{18.81 $\pm$ 1.00} & $\circ$ & 20.98 $\pm$ 1.10 & $\bullet$ \\                
Longley & 16 & 0 & 6 & 0 & \textbf{6.61 $\pm$ 7.56} & 14.75 $\pm$ 12.00 & $\bullet$ & 16.67 $\pm$ 17.19 & $\bullet$ & 8.77 $\pm$ 9.75 & $\bullet$ & 9.57 $\pm$ 8.02 & $\bullet$ & 14.32 $\pm$ 10.72 & $\bullet$ \\                   
Low birth weight & 189 & 1 & 8 & 0 & 40.09 $\pm$ 13.68 & 39.63 $\pm$ 12.92 &  & \textbf{37.97 $\pm$ 13.03} & $\circ$ & 38.89 $\pm$ 13.27 & $\circ$ & 45.71 $\pm$ 16.38 & $\bullet$ & 43.58 $\pm$ 15.95 & $\bullet$ \\                
Machine CPU & 209 & 0 & 6 & 0 & \textbf{12.80 $\pm$ 23.68} & 16.69 $\pm$ 30.55 & $\bullet$ & 23.39 $\pm$ 37.67 & $\bullet$ & 14.79 $\pm$ 24.89 & $\bullet$ & 13.10 $\pm$ 26.48 &  & 17.53 $\pm$ 30.16 & $\bullet$ \\                 
MBA grade & 61 & 0 & 2 & 0 & 106.1 $\pm$ 69.36 & 106.0 $\pm$ 68.30 &  & \textbf{91.00 $\pm$ 66.94} & $\circ$ & 107.0 $\pm$ 69.62 & $\bullet$ & 106.5 $\pm$ 68.01 &  & 107.3 $\pm$ 68.51 &  \\                                        
Meta & 528 & 2 & 19 & 4.55 & \textbf{93.29 $\pm$ 159.3} & 95.34 $\pm$ 165.4 & $\bullet$ & 100.3 $\pm$ 185.3 &  & 95.28 $\pm$ 165.6 &  & 97.12 $\pm$ 175.6 &  & 99.64 $\pm$ 184.7 &  \\                                               
MV & 40768 & 3 & 7 & 0 & \textbf{0.00 $\pm$ 0.00} & 0.02 $\pm$ 0.01 & $\bullet$ & 0.02 $\pm$ 0.00 & $\bullet$ & 0.00 $\pm$ 0.00 & $\bullet$ & 0.29 $\pm$ 0.01 & $\bullet$ & 0.36 $\pm$ 0.02 & $\bullet$ \\                           
PBC & 418 & 1 & 17 & 16.47 & 62.81 $\pm$ 12.95 & 65.11 $\pm$ 13.09 & $\bullet$ & \textbf{60.50 $\pm$ 13.39} & $\circ$ & 62.45 $\pm$ 12.79 & $\circ$ & 64.24 $\pm$ 13.04 & $\bullet$ & 65.00 $\pm$ 13.14 & $\bullet$ \\               
Pharynx & 195 & 2 & 10 & 0.09 & 36.35 $\pm$ 16.41 & 37.24 $\pm$ 17.90 & $\bullet$ & \textbf{32.47 $\pm$ 17.89} & $\circ$ & 37.83 $\pm$ 16.34 & $\bullet$ & 46.21 $\pm$ 19.35 & $\bullet$ & 49.02 $\pm$ 19.39 & $\bullet$ \\          
Pole & 15000 & 0 & 48 & 0 & 2.43 $\pm$ 0.21 & 2.78 $\pm$ 0.22 & $\bullet$ & \textbf{1.27 $\pm$ 0.12} & $\circ$ & 2.65 $\pm$ 0.23 & $\bullet$ & 2.25 $\pm$ 0.21 & $\circ$ & 2.75 $\pm$ 0.23 & $\bullet$ \\                            
Pollution & 60 & 0 & 15 & 0 & \textbf{43.35 $\pm$ 30.47} & 48.87 $\pm$ 34.14 & $\bullet$ & 47.37 $\pm$ 31.65 & $\bullet$ & 43.75 $\pm$ 30.44 &  & 47.81 $\pm$ 32.00 & $\bullet$ & 50.93 $\pm$ 34.02 & $\bullet$ \\                   
Puma 32H & 8192 & 0 & 32 & 0 & 65.10 $\pm$ 3.48 & 26.07 $\pm$ 1.76 & $\circ$ & \textbf{14.59 $\pm$ 0.93} & $\circ$ & 67.43 $\pm$ 3.53 & $\bullet$ & 66.53 $\pm$ 3.19 & $\bullet$ & 69.08 $\pm$ 3.30 & $\bullet$ \\                   
Puma 8NH & 8192 & 0 & 8 & 0 & 32.68 $\pm$ 1.77 & 32.64 $\pm$ 1.81 &  & 33.18 $\pm$ 1.78 & $\bullet$ & \textbf{32.60 $\pm$ 1.76} & $\circ$ & 34.93 $\pm$ 1.86 & $\bullet$ & 35.59 $\pm$ 1.83 & $\bullet$ \\                           
PW Linear & 200 & 0 & 10 & 0 & 15.33 $\pm$ 5.61 & 16.66 $\pm$ 6.09 & $\bullet$ & \textbf{15.19 $\pm$ 5.55} &  & 15.86 $\pm$ 5.83 & $\bullet$ & 21.26 $\pm$ 8.34 & $\bullet$ & 24.02 $\pm$ 9.18 & $\bullet$ \\                        
Pyrimidines & 74 & 0 & 27 & 0 & 58.21 $\pm$ 113.2 & 63.96 $\pm$ 125.5 & $\bullet$ & 69.33 $\pm$ 115.6 & $\bullet$ & 59.33 $\pm$ 115.8 & $\bullet$ & \textbf{57.73 $\pm$ 113.2} &  & 60.47 $\pm$ 112.8 & $\bullet$ \\                 
Quake & 2178 & 0 & 3 & 0 & 106.1 $\pm$ 11.97 & 105.2 $\pm$ 11.57 & $\circ$ & \textbf{99.75 $\pm$ 11.67} & $\circ$ & 103.0 $\pm$ 11.51 & $\circ$ & 109.6 $\pm$ 12.23 & $\bullet$ & 104.5 $\pm$ 11.60 & $\circ$ \\                     
SCHL vote & 37 & 0 & 5 & 0 & 78.31 $\pm$ 141.5 & \textbf{58.17 $\pm$ 125.3} & $\circ$ & 93.58 $\pm$ 179.9 &  & 76.87 $\pm$ 140.8 & $\circ$ & 75.34 $\pm$ 137.8 & $\circ$ & 79.09 $\pm$ 147.0 &  \\                                   
Sensory & 576 & 0 & 11 & 0 & 77.27 $\pm$ 12.74 & \textbf{70.44 $\pm$ 11.72} & $\circ$ & 71.99 $\pm$ 11.75 & $\circ$ & 77.80 $\pm$ 12.83 & $\bullet$ & 77.44 $\pm$ 12.77 &  & 78.74 $\pm$ 13.20 & $\bullet$ \\                        
Servo & 167 & 4 & 0 & 0 & \textbf{10.93 $\pm$ 14.87} & 13.85 $\pm$ 18.54 & $\bullet$ & 14.75 $\pm$ 20.91 & $\bullet$ & 12.34 $\pm$ 14.96 & $\bullet$ & 18.54 $\pm$ 19.51 & $\bullet$ & 23.10 $\pm$ 20.77 & $\bullet$ \\              
Sleep & 58 & 0 & 7 & 1.97 & 48.24 $\pm$ 23.37 & 56.58 $\pm$ 24.68 & $\bullet$ & 53.74 $\pm$ 25.09 & $\bullet$ & \textbf{48.00 $\pm$ 22.71} &  & 48.21 $\pm$ 24.02 &  & 48.82 $\pm$ 22.74 &  \\                                       
Stock & 950 & 0 & 9 & 0 & 1.11 $\pm$ 0.22 & 1.35 $\pm$ 0.34 & $\bullet$ & 1.27 $\pm$ 0.26 & $\bullet$ & 1.25 $\pm$ 0.25 & $\bullet$ & \textbf{1.06 $\pm$ 0.23} & $\circ$ & 1.27 $\pm$ 0.26 & $\bullet$ \\                            
Strike & 625 & 1 & 5 & 0 & 87.35 $\pm$ 100.0 & \textbf{84.22 $\pm$ 99.55} & $\circ$ & 90.00 $\pm$ 102.2 & $\bullet$ & 87.84 $\pm$ 99.23 &  & 87.19 $\pm$ 101.0 &  & 85.03 $\pm$ 100.8 & $\circ$ \\                                   
Triazines & 186 & 0 & 60 & 0 & 71.99 $\pm$ 36.69 & 71.01 $\pm$ 35.37 &  & \textbf{70.25 $\pm$ 34.49} &  & 71.56 $\pm$ 36.57 &  & 72.44 $\pm$ 36.78 &  & 72.87 $\pm$ 37.27 & $\bullet$ \\                                             
Veteran & 137 & 1 & 6 & 0 & 80.39 $\pm$ 77.45 & 88.10 $\pm$ 88.31 & $\bullet$ & 84.10 $\pm$ 89.71 &  & 78.38 $\pm$ 73.80 & $\circ$ & 79.03 $\pm$ 78.11 & $\circ$ & \textbf{78.35 $\pm$ 77.85} & $\circ$ \\                           
Vineyard & 52 & 0 & 3 & 0 & 34.86 $\pm$ 27.98 & 36.99 $\pm$ 33.40 &  & 46.64 $\pm$ 32.90 & $\bullet$ & 35.52 $\pm$ 28.49 & $\bullet$ & \textbf{34.77 $\pm$ 28.92} &  & 35.50 $\pm$ 31.08 &  \\                                       
Wisconsin & 194 & 0 & 32 & 0 & 87.82 $\pm$ 24.18 & 92.91 $\pm$ 24.91 & $\bullet$ & 90.38 $\pm$ 24.95 & $\bullet$ & \textbf{87.33 $\pm$ 23.96} & $\circ$ & 89.87 $\pm$ 25.33 & $\bullet$ & 88.52 $\pm$ 24.80 &  \\  
		\hline                                                             
\end{tabular}  
\end{table*}

\begin{table}[t]	
	\renewcommand{\arraystretch}{1.2}
	\footnotesize
	\centering
	\caption{Number of victories column vs row at 1\% significance level of Wilcoxon signed rank test for regression experiments (61 total).
		CCF victories and losses are given in blue and red respectively. \label{table:nVicsReg}}
	\begin{tabular}{lcccccc} \toprule
		& \text{\textbf{\color{blue} CCF}} & \text{RF} & \text{Rotation Forest} & \text{CCF-Bag} & \text{RRF} & \text{RRF-Bag}  \\ \midrule
		\textbf{\color{red} CCF} & - & \textbf{\color{red} 16} & \textbf{\color{red} 17} & \textbf{\color{red} 17} & \textbf{\color{red} 8} & \textbf{\color{red} 10}\\  
		\text{RF} & \textbf{\color{blue} 38} & - & 28 & 31 & 17 & 13 \\  
		\text{Rotation Forest} & \textbf{\color{blue} 35}& 24 & - & 31 & 22 & 16\\  
		\text{CCF-Bag} & \textbf{\color{blue} 33} & 13 & 18 & - & 11 & 3 \\  
		\text{RRF} & \textbf{\color{blue} 41} & 31 & 32 & 40 & - & 13 \\
		\text{RRF-Bag} & \textbf{\color{blue} 41} & 33 & 36 & 41 & 38 & - \\ \bottomrule
	\end{tabular}
\end{table}

\begin{table}[t]
	\renewcommand{\arraystretch}{1.2}
	\footnotesize
	\centering
	\caption{Mean rank across datasets of test mean squared error.
	 Lower rank corresponds to better performance. \label{table:rank-reg}}
	\begin{tabular}{lc} \toprule
		\text{Algorithm} & \text{Mean Rank}  \\ \midrule
		\text{CCF} & 2.43 \\
		\text{CCF-Bag} & 2.67 \\  
		\text{Rotation Forest} & 3.56 \\  
		\text{RF} & 3.64 \\
		\text{RRF} &  4.02 \\ 
		\text{RRF-Bag} &  4.69 \\   \bottomrule
	\end{tabular}
\end{table}

These results show that, as with the classification experiments, CCFs outperformed all the other methods.
Interestingly, the performance of RFs was noticeably improved relative to the other
approaches.  Though still comfortably outperformed by CCFs and CCF-Bag, RFs performed similarly
to rotation forests and substantially better than the random rotation approaches for these problems.
These results were confirmed by repeating the RF tests using the same code base as for CCFs, 
producing very similar results (this check was also performed for the classification experiments).
One possible reason for this difference is that it stems from differences in the nature of the
error criteria used for the two methods - misclassification rate and mean squared error respectively -
rather than inherent differences between classification and regression.  For example, the latter error
criteria is more influenced by rare large errors.
\pagebreak

\section{Multiple Outputs}
\label{sec:multi-out}


We now demonstrate how CCFs can be used for multiple output problems such as
multivariate regression, where we wish to regress multiple outputs simultaneously, 
and multi-output classification, where we wish to learn a single estimator to predict
multiple classification tasks.  We note that multilabel classification, where we wish
to predict a number of classes which are not mutually exclusive, can be thought
of as a special case of multi-output classification were all of the classification
tasks are binary.  In principle, CCFs can also handle problems with arbitrary
combinations of classification and regression outputs in the same way as
RFs (see e.g.~\cite{glocker2012joint,linusson2013multi}), but we will not
actively consider this case further here.

Various multiple-output axis-aligned 
decision trees and forests have previously been developed in the
literature~\citep{zhang1998classification,de2002multivariate,geurts2006kernelizing,segal2011multivariate,glocker2012joint,faddoul2012learning,linusson2013multi}.
There are three key ways these approaches tend to vary from
single axis aligned approaches: the calculation of split gains
which must now assess the utility of the split to multiple different tasks
simultaneously, the leaf models which must now predict multiple outputs,
and the manner in which the individual tree predictions are combined.

In the simplest case, the leafs have independent models for each of the outputs
and the ensemble prediction for each output is the average over trees of the corresponding leaf
predictions in the same manner as the single output case.
Improvements on this approach have been suggested.  For example,~\cite{geurts2006kernelizing} 
introduce the idea of kernelizing the decision tree outputs to ensure the ensemble
prediction respects the structure of the output data.
As these ideas apply equally well to RFs and CCFs, we leave such extensions to future work
and focus on the aforementioned independent leaf model case in the interest of simplicity. 
Using this approach, we note that if we define each $y_n$ in the multi-output case to be the concatenation of
the different outputs (i.e. such that we concatenate columns of $Y$), then the definition for
the leaf model given in~\eqref{eq:theta} and the combination of tree predictions given in~\eqref{eq:voting}
are unchanged.  Furthermore, as the partitioning procedure given in~\eqref{eq:partitioning} does not change, the
CCF prediction process is unchanged from the single output case and the training process varies
only in the split selection procedure.

Choosing a split for a multi-output decision tree requires a balancing of the
utility of the split to each of the different output tasks.  Two common approaches are
to use the same split criteria for individual outputs and then either
average or maximum gain across each of the 
outputs~\citep{faddoul2012learning}.  We consider the former and note that this is
equivalent to taking the joint information gain and reduction in multivariate mean squared error
for our default choices of impurity criterion in classification and regression cases respectively.
Specifically we have that the impurity criterion for multi-output classification is the same
as~\eqref{eq:entropy} (except that the $p_k$ now sum to the number of classification
tasks be considered) and the impurity criterion for mutlivariate regression is
\begin{align}
\label{eq:impurity-multireg}
g\left(Y_{(\omega_j,:)}\right) = \frac{1}{M} \sum_{m=1}^{M} \frac{1}{\sigma_m} \left( \frac{1}{N_j} \sum_{n \in \omega_j} Y_{(n,m)}^2 - \left(\frac{1}{N_j} \sum_{n \in \omega_j} Y_{(n,m)}\right)^2 \right)
\end{align}
where $M$ is the number of outputs and $\sigma_m$ is the standard deviation of $Y_{(:,m)}$.
The rational for the $\frac{1}{\sigma_m}$ term is to ensure invariance to the scaling of the outputs.
In practice this is done by scaling each column of $Y$ to have unit variance as a preprocessing step
such that each $\sigma_m=1$.  If one wishes to give preference to a particular output in
the estimation (e.g. because prediction of the parameter is more important or more difficult), then
one can also use a weighted average across the split gains, i.e. we can replace the $\frac{1}{\sigma_m}$
term in~\eqref{eq:impurity-multireg} with $\frac{w_m}{\sigma_m}$
where $\sum_{m=1:M} w_m =1$.  This can similarly done for classification by weighting
the groups of $p_k$ corresponding to each output.

\begin{table}[t]                        
	\centering    
	\scriptsize             
	\setlength\tabcolsep{4pt}	  
	\renewcommand{\arraystretch}{1.2}          
	\caption{Errors for multiple output tasks averaged over outputs (misclassification rate for classification, mean squared error for regression).  Method with best accuracy is shown in bold. $\bullet$ and $\circ$ indicate that CCFs were significantly better and worse respectively at the 1\% level of a Wilcoxon signed rank test.  Type $\in\{$MVR,MOC$\}$ gives the problem type corresponding to multivariate regression and multi-output classification respectively; $N_{out}=$ number of targets; $N=$ number of data points; $D_c=$ number of non-ordinal features; and $D_r=$ number of ordinal features.  Note there was no missing data for any of these datasets \label{table:multioutput}}                    
	\begin{tabular}{|c|c|c|c|c|c|c|cc|cc|}                    
		\hline           
		Dataset & Type & $N_{\mathrm{out}}$ & $N$ & $D_c$ & $D_r$ & CCF & RF & & RRF & \\
		\hline                                            
		Andro & MVR & 6 & 49 & 0 & 30 & 35.55 $\pm$ 16.07 & \textbf{23.08 $\pm$ 11.51} & $\circ$ & 29.93 $\pm$ 14.22 & $\circ$ \\    
		EDM & MVR & 2 & 154 & 0 & 16 & \textbf{42.32 $\pm$ 13.67} & 42.52 $\pm$ 13.56 &  & 42.95 $\pm$ 13.91 & $\bullet$ \\          
		ENB & MVR & 2 & 768 & 0 & 8 & \textbf{1.70 $\pm$ 0.42} & 2.19 $\pm$ 0.46 & $\bullet$ & 1.86 $\pm$ 0.42 & $\bullet$ \\        
		Jura & MVR & 3 & 359 & 0 & 15 & \textbf{33.97 $\pm$ 12.18} & 35.69 $\pm$ 12.39 & $\bullet$ & 35.36 $\pm$ 13.23 & $\bullet$ \\
		Slump & MVR & 3 & 103 & 0 & 7 & \textbf{40.00 $\pm$ 16.50} & 49.53 $\pm$ 20.94 & $\bullet$ & 46.43 $\pm$ 19.01 & $\bullet$ \\
		Emotions & MOC & 6 & 593 & 0 & 72 & \textbf{17.85 $\pm$ 1.93} & 18.15 $\pm$ 1.99 & $\bullet$ & 17.99 $\pm$ 1.98 &  \\     
		Flags & MOC & 4 & 194 & 3 & 20 & 60.41 $\pm$ 6.99 & \textbf{58.12 $\pm$ 7.30} & $\circ$ & 60.61 $\pm$ 6.93 &  \\             
		Solar Flare & MOC & 3 & 1066 & 3 & 7 & 8.25 $\pm$ 1.59 & 8.27 $\pm$ 1.55 &  & \textbf{8.21 $\pm$ 1.56} &  \\          
		\hline                                         
	\end{tabular}                                                            
\end{table}  

The final piece of the puzzle for using CCFs for multi-output problems is to note that when $Y$ is
the concatenation of all the outputs, then the CCA procedure given in~\eqref{eq:canonFeat}, 
and thus the process for calculating $\Phi$, is unchanged.  This seamless transition is
an advantage to our choice of using CCA projections.  In theory, one could also make
use the canonical components of the outputs for the multi-output case.  However, preliminary
investigation suggests that a na\"{i}ve approach, where these are used for the split criterion
rather than the original outputs, is not beneficial.

To assess the performance of CCFs for multi-output problems we considered a selection
of datasets taken from the 
Mulan\footnote{\scriptsize \url{http://mulan.sourceforge.net/datasets-mtr.html}}
~\citep{tsoumakas2010mining} and UCI~\citep{Lichman2013UCI}
databases.  The experimental procedures was as for the single output
experiments (using the same respective default parameters for regression 
and classification) and we make comparisons to RF and RRF.  As the RF package used previously does not support multiple outputs, an adaptation of our CCF package was used instead.  Dataset summaries and results
are shown in Table~\ref{table:multioutput} with win/draw/loss ratios for CCFs of $4/2/2$ and $4/3/1$
compared to RFs and RRFs respectively.  More extensive investigation is necessary to
fully assess the relative merit of CCFs compared to alternatives for multi-output problems, but these
results are promising and suggest gains transfer from the single output case.

\section{Discussion}
\label{sec:Discussion}


\subsection{Effect of Correlation between Input Features}
\label{sec:effectOfCorrelation}

As shown by \citet{menze2011oblique}, RFs often struggle on data with highly correlated features.  CCA naturally incorporates information about feature correlations, therefore CCFs do not suffer the same issues. This is demonstrated, for example, by their superior performance on the highly correlated \textit{hill valley} datasets.

\begin{figure}[t]
	\centering
	\begin{subfigure}[t]{0.49\textwidth}
		\caption{Banknote}
		\includegraphics[width=0.99\textwidth,trim={5.5cm 0 0.5cm 0},clip]{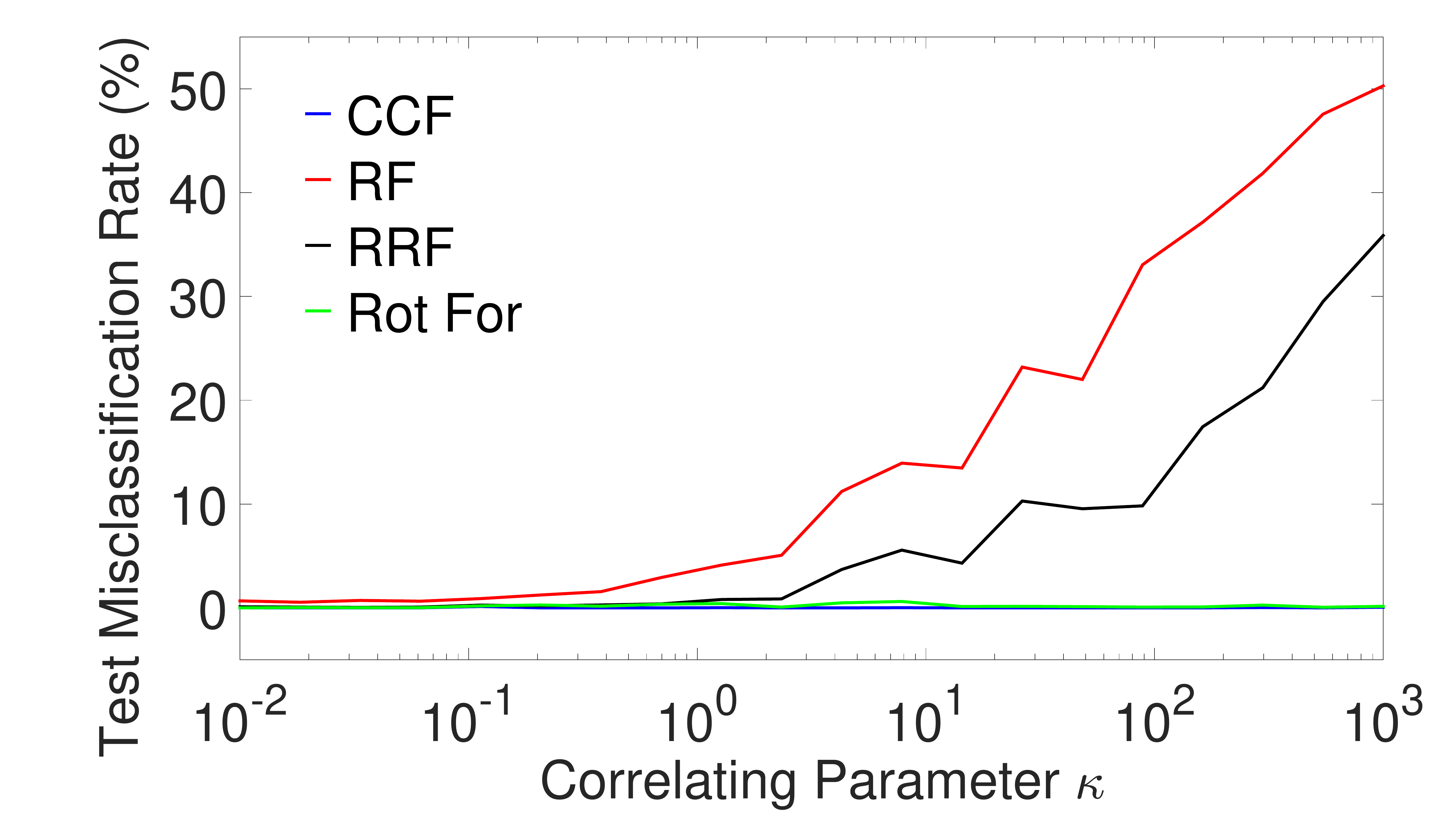}
		\centering
	\end{subfigure}
	~
	\begin{subfigure}[t]{0.49\textwidth}
		\caption{Iris}
		\includegraphics[width=0.99\textwidth,trim={5.5cm 0 0.5cm 0},clip]{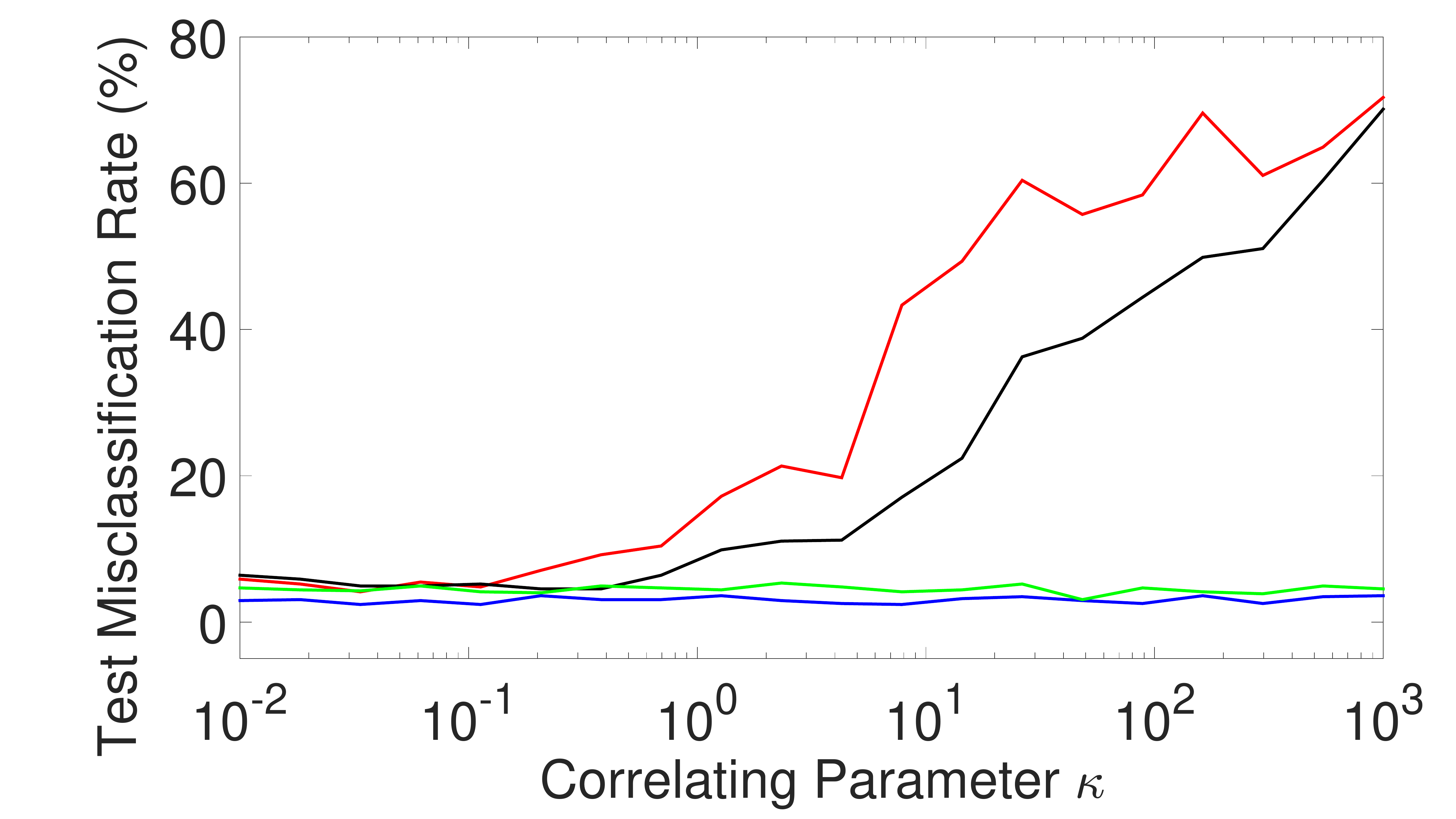}
		\centering
	\end{subfigure}	
	\caption{Misclassification rate on artificially correlated data.  For each method, 5 separate 10-fold cross validations were run and the mean test accuracy across the 50 tests reported.
		\label{fig:corrAccsup}}
\end{figure}

To more rigorously investigate the effect of correlation, we used the following method of artificially correlating the data (after carrying out the preprocessing described in Section \ref{sec:DataPrepro}):
\begin{enumerate}
	\item Add a new random feature 	
	\[
	X_{\left(D+1,n\right)} \sim \mathcal{N} \left(0, \kappa\right), \quad \forall n=1, \dots, N
	\]	where $\kappa$ is a parameter which will control the degree of correlation to be added.
	\item To each of the existing features, randomly either add or subtract the new feature, i.e.
	\[
	X_{\left(:,d\right)} \leftarrow X_{\left(:,d\right)} + \zeta_d X_{\left(:,D+1\right)}, \; \forall d = 1,\dots,D
	\]
	where each $\zeta_d \overset{\text{i.i.d}}{\sim} \textsc{Uniform-Discrete} \left\{-1,1\right\}$.
\end{enumerate}
As shown in Figure \ref{fig:corrAccsup}, this transformation has no effect on the accuracy of CCFs, whereas the accuracy of RF decreases with increasing $\kappa$, eventually giving the same accuracy as random prediction.  
We also see that RRF fare only marginally better than RF under this artificial correlation. 
The use of PCA means rotation forests, on the other hand, exhibit similar robustness to global correlations as CCFs.

We postulate that CCFs are better than rotation forests at incorporating class dependent and localized correlations.  This is because the rotation step of rotation forests does not incorporate any class information other than in the random elimination of classes.  Further, individual trees in a rotation forest are orthogonal and therefore cannot incorporate spatial variation in correlation.  The self similar nature of the growth algorithm for CCTs, on the other hand, means that the local correlations of a partition can be incorporated as naturally as the global correlations.  

To investigate these suggestions formally, we tested performance on compound datasets in which localized and class dependent correlations had been artificially added.  To do this, a replica of the dataset was created where a datapoint with class $k$ in the original is assigned to class $k+K$ in the replica.  
The features in the original and the replica were independently correlated and a constant added to replica to separate it in the input space from the original.  A new dataset was then created
composing of the datapoints from the original dataset and from the replica. Specifically we construct the new dataset as
\begin{align}
\label{eq:corrClassDep}
X \leftarrow \begin{bmatrix} \Call{Corr}{X} \\ \Call{Corr}{X} + \beta \mathds{1}^{N\times \left(D+1\right)}\end{bmatrix}, \quad Y \leftarrow \begin{bmatrix} Y & \mathbf{0} \\ \mathbf{0} & Y \end{bmatrix}
\end{align}
where \textsc{Corr} denotes the correlation process described earlier in the section and $\beta$ is a scalar dictating the degree of separation.  Note that the \textsc{Corr} process is applied separately for the points in the
original dataset and the replica, such that these correlation are class dependent.

We performed a single crossfold validation on a selection of the datasets from Table \ref{table:ResultsTable}, taking $\beta = 2000$ and $\kappa = 100$.
The results, given in Table \ref{table:artCorr}, show that CCFs demonstrated
either no change or a small loss in accuracy on all datasets, whereas there was a large loss of accuracy in all cases for RFs and RRFs, and on most of the datasets for rotation forests. 
These results supports our hypothesis that CCFs are better than rotation forests at dealing with localized and class dependent correlations.  This is an important advancement as such correlations cannot be easily
accounted for using a data preprocessing step such as PCA.
We note that the only cases that did not
exhibit a large drop in accuracy for rotation forests had $K=2$, suggesting the use of class elimination in
the calculation of the rotations~\citep{rodriguez2006rotation} may offer some robustness to
class dependent correlations when the number of classes is small.

\begin{table}[t]                        
	\centering    
	\scriptsize             
	\setlength\tabcolsep{4pt}	  
	\renewcommand{\arraystretch}{1.2}          
	\caption{Mean and standard deviations of percentage of test cases misclassified for compound datasets.  Method with best accuracy is shown in bold. $\bullet$ and $\circ$ indicate that CCFs were significantly better and worse respectively at the 1\% level of a Wilcoxon signed rank test. \label{table:artCorr}}                    
	\begin{tabular}{|c|c|cc|cc|cc|}                    
		\hline                                                       
		Dataset & CCF & RF & & Rot For & & RRF &\\
		\hline           
Balance Scale & \textbf{7.36 $\pm$ 2.09} & 46.72 $\pm$ 5.92 & $\bullet$ & 20.24 $\pm$ 4.46 & $\bullet$ & 40.48 $\pm$ 5.42 & $\bullet$ \\    
Banknote & \textbf{0.00 $\pm$ 0.00} & 34.64 $\pm$ 2.94 & $\bullet$ & 2.85 $\pm$ 0.62 & $\bullet$ & 23.54 $\pm$ 1.98 & $\bullet$ \\          
Iris & \textbf{3.00 $\pm$ 3.31} & 69.33 $\pm$ 7.83 & $\bullet$ & 19.67 $\pm$ 9.87 & $\bullet$ & 65.33 $\pm$ 8.78 & $\bullet$ \\             
Landsat satellite & \textbf{8.69 $\pm$ 0.97} & 32.75 $\pm$ 1.44 & $\bullet$ & 11.56 $\pm$ 0.82 & $\bullet$ & 19.04 $\pm$ 0.89 & $\bullet$ \\
Parkinsons & \textbf{9.74 $\pm$ 3.78} & 33.33 $\pm$ 8.11 & $\bullet$ & 12.56 $\pm$ 6.22 &  & 27.18 $\pm$ 9.06 & $\bullet$ \\                
Vowel-n & \textbf{3.13 $\pm$ 0.46} & 85.91 $\pm$ 2.82 & $\bullet$ & 20.20 $\pm$ 2.79 & $\bullet$ & 78.69 $\pm$ 2.87 & $\bullet$ \\           
Wisconsin cancer & \textbf{2.79 $\pm$ 1.76} & 30.00 $\pm$ 2.65 & $\bullet$ & 2.93 $\pm$ 1.66 &  & 8.79 $\pm$ 2.88 & $\bullet$ \\            
Yeast & \textbf{38.72 $\pm$ 1.89} & 70.13 $\pm$ 3.15 & $\bullet$ & 56.09 $\pm$ 1.85 & $\bullet$ & 63.13 $\pm$ 3.36 & $\bullet$ \\              
		\hline                                         
	\end{tabular}                                                            
\end{table}

\subsection{Computational Complexity}
\label{sec:CompComp}

We now consider the computational complexity of the CCF training algorithm.  
We focus on the classification case, but
note that the results similarly apply to regression by setting $K=1$.  For multiple outputs then the computational
complexity is a little more complex, but the quoted results roughly hold for 
multivariate regression by setting $K$ to the number of outputs and to multiple outputs by replacing
$K$ with the total number of classes across all outputs.

The two limiting factors for the complexity of CCFs are the search over possible splits and the CCA calculations.  We first consider the cost of this calculation at a single node compared to
a RF.  The theoretically achievable computational 
complexity\footnote{
	This is based on exploiting the LDA equivalence.  
	The numerically stable method outlined in Appendix~\ref{sec:numStabCCASup} that was used for our experiments 
	is actually $\mathrm{O}\left(N_j (\lambda+K)^2 +\lambda K \min(\lambda,K) \right)$, 
	with the additional terms occuring due to the QR decomposition of the centred $Y$, the multiplication
	of the $Q$ matrices, and
	the SVD decomposition.  In practise we found the constant 
	factor for these terms to be small and therefore that this was not a problem.  However, for datasets 
	with a very large $K$, an alternative method for CCA might be preferable, for example by exploiting 
	the LDA equivalence, if numerical stability can be ensured by other means.  This is also why
	the computational complexity for the multi-output cases is more complicated.} 
for each CCA calculation is $\mathrm{O}(N_j \lambda^2)$~\citep{golub2012matrix}.  
Assuming all features are 
ordinal,\footnote{In the presence of non-ordinal features, $\lambda$ is replaced 
	in the CCF complexity with the number of features used after the 1-of-K encoding is applied. 
	}
the cost of the
split search a single node for CCFs given $\Phi$ is $\mathrm{O}(N_j \nu_{\max} \log N_j)$ where
$\nu_{\max} \le \min \left(\lambda,K-1\right)$ is the number dimensions that need to be searched over.
This compares with $\mathrm{O}(N_j \lambda \log N_j)$
for RFs, with the dominant cost for both 
coming from sorting $U$ (or equivalently $X_{(\omega_j,\delta_j)}$ for RFs).  
As CCA and the split search are the two dominant costs, we can say that the per node
cost of CCFs is $\mathrm{O}(N_j \lambda^2 + N_j \nu_{\max} \log N_j)$ and of RFs
is $\mathrm{O}(N_j \lambda \log N_j)$.  \cite{louppe2014understanding} showed
that this per node cost for RFs leads to an overall complexity of 
$\mathrm{O}\left(NL\lambda \left(\log N\right)^2\right)$.  Presuming that CCFs
have the same structural properties as RFs, using their as~\cite{louppe2014understanding} gives a complexity
of 
\[
\mathrm{O}\left(NL \lambda^2\log N+ NL\nu_{\max} \left(\log N\right)^2 \right)
\]
for CCFs. Clearly the second term is bounded by the complexity of RFs (as $\nu_{\max}\le\lambda$) 
and may be substantially smaller when $K$ is small, e.g. for regression problems.
The first term, and thus the overall CCF complexity, 
will also be bounded by the complexity of RF if $\lambda < \log N$. 
Using the recommended value of $\lambda = \ceil*{\log_2 \left(D\right) + 1}$  this means the complexity of CCF is upper bounded by that of RF whenever $N>D$ and is at worst a factor of $\log(D)/\log(N)$ larger.  

If $\Phi$ is instead calculated using random rotations then this can be done in 
$O(\lambda^2)$ time~\citep{blaser2016random}.  Consequently it immediately follows that the computational
complexity for random rotation approaches is $\mathrm{O}\left(NL\lambda \left(\log N\right)^2 + L\lambda^2\log N \right)$ which is the same as RFs whenever $\lambda < N\log N$.

Rotation forest training requires each tree to search the full set of features and therefore has a training complexity of $\mathrm{O} \left(NLD\left(\log N\right)^2\right)$.  This is exponentially more expensive in the number of features than RFs and CCFs when $\lambda$ is set to a logarithmic factor of $D$.  This makes their implementation impractical for datasets with more than a modest number of features, as demonstrated by the \textit{ORL} and \textit{polya} datasets for which we were unable to carry out most tests with rotation forests, experiencing both memory issues and an impractically long training time.

Should one wish to use CCFs with a $\lambda$ that is significantly larger than the suggested default, speeds gains should be achievable by separately generating $M$ projection matrices $\left\{\Phi_m \right\}_{m=1:M}$, each using $\lambda / M$ features, and searching for the optimal split over all generated projections.  
When $K$ is small, such an approach could also be used to increase the size of the set of candidate splits
generated and might therefore also be beneficial to performance in some scenarios.
We leave assessing the effect of this change on predictive performance to future work.

\subsection{Empirical Run Times}
\label{sec:empirical_speed}

\begin{table*}[t]                        
	\centering    
	\scriptsize             
	\setlength\tabcolsep{3pt}	  
	\renewcommand{\arraystretch}{1.2}    
	\caption{Combined time for training and testing (using a $90\%/10\%$ train/test split, $500$ trees, and the default parameters) 
		and average number of nodes in each tree
		for CCFs and the other algorithms.  CCF results are quoted in absolute values, while others are given
		as ratios of the equivalent value for CCFs.  Results that correspond to faster run times or smaller
		trees than the corresponding CCF are shown in bold. Results are averaged across $4$ separate train / test splits.  Variations in
		run time and average tree size between independent runs were small (typically in the range $0.5\%-5\%$).
		\label{table:Timing}}                            
\begin{tabular}{|c|c|c|c|c|c|c|c|c|c|c|c|c|}                                                                                                                 
	\hline                                                                                                                                                       
	Dataset & CCF & \multicolumn{5}{c|}{Run Time / CCF Run Time} & CCF & \multicolumn{5}{c|}{Avg. \#Nodes  / CCF Avg. \#Nodes} \\                                
	\cline{3-7} \cline{9-13}                                                                                                                                     
	& Time (s) & RF & Rot-For & CCF-Bag & RRF & RRF-Bag & \#Nodes & RF & Rot-For & CCF-Bag & RRF & RRF-Bag \\                                                    
	\hline                                                                                                                                                       
	Balance scale & 2.41 & 1.44 & 1.24 & \textbf{0.81} & 1.29 & 1.09 & 125 & 1.49 & 1.34 & \textbf{0.613} & 1.31 & \textbf{0.857} \\                             
	Banknote & 0.716 & 1.72 & 1.65 & 1.13 & 1.37 & 1.53 & 16.5 & 2.18 & 2.16 & \textbf{0.943} & 1.88 & 1.72 \\                                                   
	Breast tissue & 0.928 & 1.13 & 1.68 & \textbf{0.925} & 1.35 & 1.19 & 43.4 & \textbf{0.758} & 1.03 & \textbf{0.579} & 1.16 & \textbf{0.801} \\                
	Climate crashes & 0.835 & 1.27 & 2.36 & \textbf{0.912} & 1.54 & 1.34 & 35.0 & \textbf{0.76} & \textbf{0.899} & \textbf{0.562} & 1.20 & \textbf{0.83} \\      
	Fertility & 0.804 & 1.16 & 1.26 & \textbf{0.915} & 1.14 & 1.02 & 26.1 & \textbf{0.879} & \textbf{0.984} & \textbf{0.525} & \textbf{0.947} & \textbf{0.616} \\
	Heart-SPECT & 1.54 & 1.48 & 2.63 & \textbf{0.78} & 1.47 & 1.18 & 79.7 & \textbf{0.974} & 1.05 & \textbf{0.526} & \textbf{0.966} & \textbf{0.608} \\          
	Heart-SPECTF & 1.05 & 1.51 & 4.65 & \textbf{0.821} & 1.67 & 1.43 & 58.0 & \textbf{0.765} & \textbf{0.718} & \textbf{0.502} & 1.02 & \textbf{0.697} \\        
	Hill valley & 0.847 & 14.6 & 7.55 & 1.02 & 11.7 & 10.6 & 25.1 & 14.0 & \textbf{0.576} & \textbf{0.785} & 13.0 & 10.1 \\                                      
	Hill valley noisy & 4.39 & 2.64 & 12.6 & \textbf{0.764} & 2.93 & 2.47 & 319 & 1.00 & \textbf{0.709} & \textbf{0.511} & 1.42 & \textbf{0.985} \\              
	ILPD & 3.14 & 1.25 & 1.90 & \textbf{0.812} & 1.58 & 1.26 & 196 & \textbf{0.746} & \textbf{0.878} & \textbf{0.553} & \textbf{0.992} & \textbf{0.672} \\       
	Ionosphere & 0.945 & 1.56 & 4.07 & \textbf{0.922} & 1.77 & 1.61 & 47.7 & \textbf{0.773} & \textbf{0.782} & \textbf{0.571} & 1.11 & \textbf{0.80} \\          
	Iris & 0.706 & 1.14 & 1.05 & 1.02 & 1.05 & 1.12 & 12.0 & 1.03 & 1.27 & \textbf{0.721} & 1.47 & 1.09 \\                                                       
	Landsat satellite & 15.7 & 1.76 & 5.96 & \textbf{0.824} & 1.94 & 1.66 & 946 & \textbf{0.853} & \textbf{0.81} & \textbf{0.54} & 1.10 & \textbf{0.756} \\      
	Letter & 124 & 1.15 & 2.85 & \textbf{0.936} & 1.34 & 1.24 & 4656 & \textbf{0.849} & \textbf{0.932} & \textbf{0.658} & 1.28 & \textbf{0.933} \\               
	Libras & 2.74 & 1.63 & 9.58 & \textbf{0.89} & 1.80 & 1.59 & 130 & 1.01 & \textbf{0.819} & \textbf{0.632} & 1.23 & \textbf{0.938} \\                          
	Magic & 47.0 & 1.43 & 2.53 & \textbf{0.804} & 1.82 & 1.54 & 3901 & \textbf{0.641} & \textbf{0.874} & \textbf{0.553} & 1.03 & \textbf{0.69} \\                
	Nursery & 11.0 & 1.25 & 2.03 & 1.04 & 2.14 & 2.25 & 562 & \textbf{0.733} & 1.12 & \textbf{0.784} & 1.73 & 1.47 \\                                            
	ORL & 15.9 & 1.22 & 613 & \textbf{0.968} & 1.41 & 1.35 & 200 & \textbf{0.756} & \textbf{0.542} & \textbf{0.611} & 1.26 & \textbf{0.92} \\                    
	Optical digits & 20.9 & 1.40 & 6.75 & \textbf{0.899} & 1.70 & 1.51 & 849 & \textbf{0.91} & \textbf{0.724} & \textbf{0.63} & 1.36 & \textbf{0.986} \\         
	Parkinsons & 0.782 & 1.26 & 2.19 & \textbf{0.942} & 1.37 & 1.22 & 30.3 & \textbf{0.808} & \textbf{0.78} & \textbf{0.607} & 1.06 & \textbf{0.785} \\          
	Pen digits & 18.0 & 1.39 & 3.05 & \textbf{0.993} & 1.49 & 1.44 & 590 & 1.05 & \textbf{0.957} & \textbf{0.721} & 1.36 & 1.06 \\                               
	Polya & 31.6 & 2.43 & 21.3 & \textbf{0.703} & 2.41 & 2.00 & 2245 & \textbf{0.94} & \textbf{0.648} & \textbf{0.459} & 1.07 & \textbf{0.726} \\                
	Seeds & 0.724 & 1.29 & 1.35 & 1.01 & 1.27 & 1.20 & 20.8 & 1.08 & 1.23 & \textbf{0.665} & 1.43 & 1.03 \\                                                      
	Skin seg & 29.3 & 1.93 & 1.66 & 1.28 & 1.47 & 1.78 & 367 & 1.26 & \textbf{0.791} & \textbf{0.763} & 1.11 & \textbf{0.866} \\                                 
	Soybean & 4.00 & 1.67 & 5.61 & \textbf{0.924} & 1.79 & 1.72 & 119 & 1.02 & 1.01 & \textbf{0.699} & 1.20 & \textbf{0.949} \\                                  
	Spirals & 9.00 & 1.06 & \textbf{0.923} & \textbf{0.982} & \textbf{0.945} & \textbf{0.929} & 263 & 1.17 & 1.36 & 1.00 & \textbf{0.835} & \textbf{0.839} \\    
	Splice & 9.21 & 4.52 & 21.0 & \textbf{0.751} & 6.25 & 5.14 & 429 & 1.20 & \textbf{0.57} & \textbf{0.403} & 1.85 & 1.28 \\                                    
	Vehicle & 3.39 & 1.66 & 3.47 & \textbf{0.807} & 2.01 & 1.68 & 237 & \textbf{0.881} & \textbf{0.926} & \textbf{0.562} & 1.24 & \textbf{0.865} \\              
	Vowel-c & 4.59 & 1.91 & 4.39 & \textbf{0.959} & 2.12 & 2.02 & 225 & 1.10 & 1.14 & \textbf{0.70} & 1.47 & 1.21 \\                                             
	Vowel-n & 4.66 & 1.36 & 2.20 & \textbf{0.956} & 1.57 & 1.52 & 255 & \textbf{0.933} & 1.08 & \textbf{0.728} & 1.25 & 1.01 \\                                  
	Waveform (1) & 12.5 & 1.75 & 4.10 & \textbf{0.765} & 2.16 & 1.80 & 991 & \textbf{0.791} & \textbf{0.834} & \textbf{0.522} & 1.16 & \textbf{0.789} \\         
	Waveform (2) & 13.4 & 1.92 & 6.84 & \textbf{0.763} & 2.67 & 2.19 & 1038 & \textbf{0.786} & \textbf{0.777} & \textbf{0.506} & 1.34 & \textbf{0.916} \\        
	Wholesale-c & 1.17 & 1.21 & 1.83 & \textbf{0.845} & 1.47 & 1.30 & 70.8 & \textbf{0.655} & \textbf{0.872} & \textbf{0.57} & 1.05 & \textbf{0.705} \\          
	Wholesale-r & 2.94 & 1.10 & 1.74 & \textbf{0.803} & 1.37 & 1.11 & 225 & \textbf{0.657} & \textbf{0.906} & \textbf{0.56} & 1.01 & \textbf{0.65} \\            
	Wisconsin cancer & 0.883 & 1.61 & 1.73 & \textbf{0.931} & 1.44 & 1.31 & 43.9 & 1.00 & \textbf{0.887} & \textbf{0.564} & 1.03 & \textbf{0.707} \\             
	Yeast & 12.0 & 1.12 & 1.84 & \textbf{0.793} & 1.24 & 1.03 & 862 & \textbf{0.748} & \textbf{0.966} & \textbf{0.575} & 1.03 & \textbf{0.662} \\                
	Zoo & 0.749 & 1.17 & 1.46 & 1.11 & 1.17 & 1.30 & 20.6 & \textbf{0.932} & \textbf{0.841} & \textbf{0.786} & 1.07 & \textbf{0.885} \\                          
	\hline                                                                                                                                           
	Average & 11.2 & 1.92 & 20.9 & \textbf{0.906} & 2.03 & 1.83 & 548 & 1.30 & \textbf{0.941} & \textbf{0.627} & 1.54 & 1.15 \\                                  
	\hline                                                                                                                                                       
\end{tabular}                                              
\end{table*}  

To more explicitly evaluate the practical speed of CCFs compared to competitors, we now consider
empirical run times.  To ensure fair comparison, we used our code base for the CCF, CCF-Bag, RRF,
and RRF-Bag implementations for calculating the RFs and rotation forests timings.  All methods thus shared the same code 
except in the calculation of $\Phi$
(which is just set to an identity matrix for RFs and rotation forests), whether bagging is used,
and the tree projection calculation and different
value of $\lambda$ required for rotation forests.  
Runs were carried out on a single machine with 64 AMD Opteron $1.4$GHz processors and using
$2$ threads per core.

Table~\ref{table:Timing} shows the absolute run times for CCFs and
the run times of the alternatives as a ratio of the corresponding CCF run time.  To characterise whether changes
originate from differences in the size of the learnt trees, the average number of nodes for each tree
are also shown.  As expected, 
the run times for rotation forests were significantly larger that
the other algorithms for any datasets with more than a few input features.  
CCF-Bag was (marginally) faster than CCFs and produced
smaller trees, as would also be expected because bagging reduces 
the effective size of the dataset each tree is trained on.
Perhaps
surprisingly though, RFs, RRFs, and RRF-Bag were all substantially slower than CCFs, each having
longer run times on every dataset except \textit{spirals} and respectively taking on average
$1.92$, $2.03$, and $1.83$ times longer to run than CCFs.  These numbers rise further to
$2.57$, $2.62$, and $2.35$ respectively if one omits time spent on parallelization overheads.
Although the relative speeds will of course vary between implementations, these results suggest
not only do CCFs offer better predictive accuracy than
the alternatives, they are faster to train as well.

Though occasionally producing substantially larger trees, RFs and RRF-Bag more often than
not produced smaller trees than CCFs, again presumably because of the reduction in effective
training dataset size caused by bagging.  The
speed differences for these algorithms cannot thus originate predominantly from changes in the tree size.  
CCFs are in fact faster because
they consider, on average, a smaller number of candidate splits than the other approaches,
with the cost of searching of these splits typically being larger than the cost of carrying out CCA.
This is easy to see in the case were $K<\lambda$ as CCFs search
over $\nu_{\max} \le \max(K,\lambda)$ projections, compared to $\lambda$ for the alternatives.
Somewhat more subtly, then this effect still occurs for
the tree as a whole even when $K>\lambda$.  Although it is still necessary to search over $\lambda$ splits at the root of the tree when $K>\lambda$, the partitioning
procedure attempts to create the ``purest'' possible children and so the average number of classes
present at a node decreases the deeper one goes down the tree.  As $\nu_{\max}$ for a particular node is less than
the number of classes present, once there are fewer classes than
$\lambda$, CCFs needs to search over fewer splits than the alternatives at that node and all
its descendants.  Therefore fewer candidate splits are considered for the tree as a whole, leading to
a faster training speed.

\begin{figure}[t]
	\centering
	\includegraphics[width=0.54\textwidth]{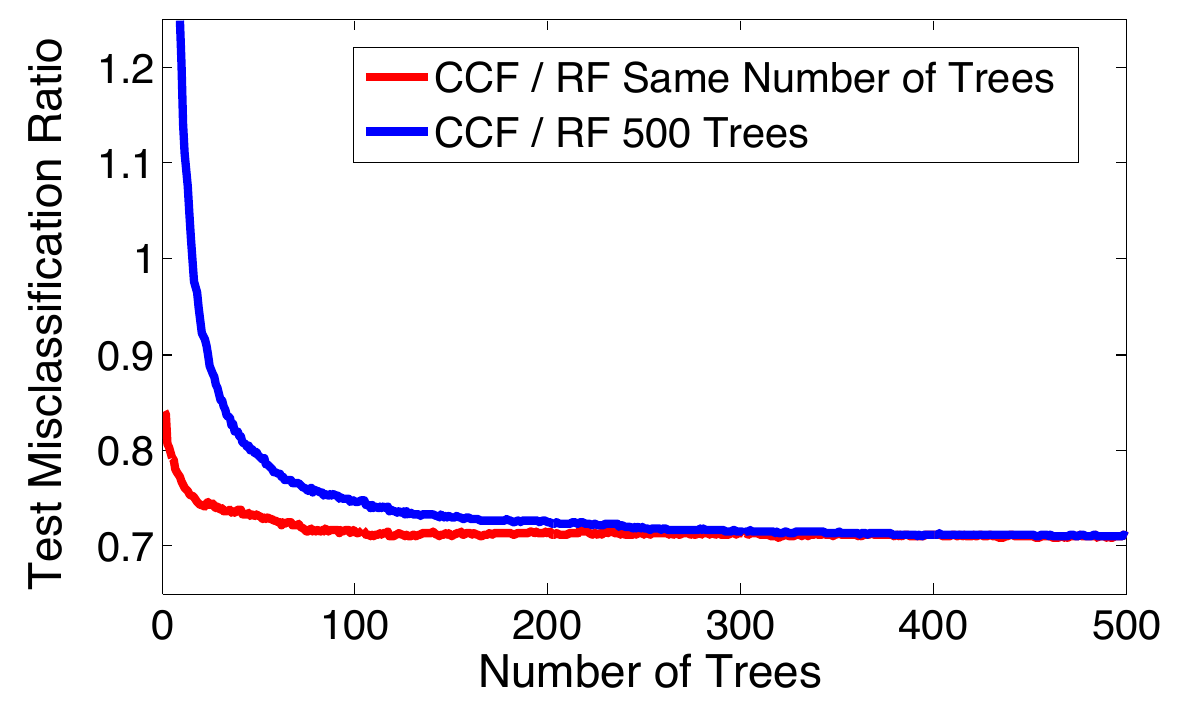}
	\centering
	\caption{Mean misclassification ratios for 37 datasets in Table \ref{table:ResultsTable}. The red line shows the average across the datasets of the ratio between the number of CCF and RF test misclassifications for forests of the same size. The blue line gives the same ratio when comparing CCFs of varying number of trees to a RF with 500 trees.\label{fig:varyNoTrees}}
\end{figure}

\subsection{Effect of Number of Trees}
\label{sec:noTrees}

To investigate the variation in performance with the number of trees we used the individual tree predictions from Section~\ref{sec:treeComparison} to evaluate the accuracy of RF and CCF for different ensemble sizes. Figure \ref{fig:varyNoTrees} summarises the results and shows that, as expected, it is always beneficial to use more trees, but that there are diminishing returns.  Some datasets require more trees than others before the accuracy begins to saturate (not shown), but the advantages of CCFs over RFs is maintained across all ensemble sizes (though for a very small number of trees this is slightly less pronounced).  A further result of interest is that on average it only takes around 15 CCTs to match the accuracy of 500 RF trees.  One could therefore envisage using a smaller CCF as a fast and accurate alternative to a larger RF under a restricted computational budget.

\section{Conclusions}
\label{sec:Conclusions}


We have introduced canonical correlation forests (CCFs), a new decision tree ensemble method 
for classification, regression, and multiple output prediction.  
The improvements of CCFs over previous decision tree ensemble approaches are 
rooted in two core innovations: the use of CCA for generating 
candidate hyperplane splits and a novel new alternative to bagging for oblique decision trees, 
projection bootstrapping, which retains the full dataset for split selection in the projected space.  
We believe that
CCFs represent a new performance benchmark for decision tree ensembles and
have provided substantial empirical evidence to suggest that they outperform a number
of prominent approaches such as random forests (RFs), rotation forests, and randomly
rotated alternatives, while maintaining a similar or better computational complexity.
We have also provided evidence suggesting that classification CCFs outperform all of the 179
classifiers considered in a recent survey~\citep{fernandez2014we}, including a number of
prominent SVM and neural network packages, over a large selection of datasets, even when 
CCFs are run in an out-of-box fashion using default parameters and extensive parameter
tuning is applied for the alternatives.  Although it would be presumptive to conclude from this
that CCFs create a new benchmark for black-box classifiers 
more generally, it does emphatically underline their robustness and
wide ranging potential utility.
%

\newpage

\appendix


\section*{Acknowledgments}

Tom Rainforth is supported by a BP industrial grant.
Frank Wood is supported under DARPA PPAML through the U.S. AFRL under Cooperative 
Agreement number FA8750-14-2-0006, Sub Award number 61160290-111668.

\renewcommand{\thesection}{\Alph{section}}
\setcounter{section}{0}

\section{Numerically Stable CCA Algorithm}
\label{sec:numStabCCASup}

As explained in the main paper, the aim of CCA is to find pairs of linear projections $\{A_{\nu},B_{\nu}\}_{\nu=1:\nu_{\max}}$
that maximize the correlations between two matrices $W\in\mathbb{R}^{n\times d}$ and $V\in\mathbb{R}^{n\times k}$
in the co-projected space $\{WA_{\nu},VB_{\nu}\}$. 
Let $\Sigma_{WV}$ be the cross covariance matrix between $W$ and $V$ such that
the $(i,j)$ entry of $\Sigma_{WV}$ (which we denote $\left(\Sigma_{WV}\right)_{(i,j)}$)
is the empirical covariance between the $i^{\mathrm{th}}$ column of $W$ and the $j^{\mathrm{th}}$
column of $V$, namely
\begin{align}
\left(\Sigma_{WV}\right)_{(i,j)} &= \frac{1}{N-1}\sum_{n=1}^{N}\left(W_{(n,i)}-\bar{W}_i\right)\cdot\left(V_{(n,j)}-\bar{V}_j\right) \\
\mathrm{where} \quad
\bar{W}_i &= \frac{1}{N} \sum_{n=1}^{N} W_{(n,i)}, \quad \mathrm{and} \quad \bar{V}_j = \frac{1}{N} \sum_{n=1}^{N} V_{(n,j)}. \nonumber
\end{align}
Defining $\Sigma_{WW}$ and $\Sigma_{VV}$ in the same way, then a simple approach to
calculate the canonical coefficients $\{A_{\nu},B_{\nu}\}_{\nu=1:\nu_{\max}}$ is to
note that the  satisfy (see e.g. \cite{borga2001canonical})
\begin{align}
\label{eq:solCCA}
\begin{split}
\Sigma_{WW}^{-1} \Sigma_{WV} \Sigma_{VV}^{-1} \Sigma_{VW} A_{\nu} & = \rho_{\nu}^2 A_{\nu} \\
\Sigma_{VV}^{-1} \Sigma_{VW} \Sigma_{WW}^{-1} \Sigma_{WV} B_{\nu} & = \rho_{\nu}^2 B_{\nu}
\end{split}
\end{align}
such that they are the eigenvectors of $\Sigma_{WW}^{-1} \Sigma_{WV} \Sigma_{VV}^{-1} \Sigma_{VW}$ and
$\Sigma_{VV}^{-1} \Sigma_{VW} \Sigma_{WW}^{-1} \Sigma_{WV}$ respectively.
The common eigenvalues, which correspond to the squares of the canonical correlations $\rho_{\nu}$, make the pairing between the coefficients of $W$ and $V$ apparent and the order of the coefficients is found by the corresponding decreasing order of the canonical correlations, $\rho_1 \geq \rho_2 \geq \dots \geq \rho_{\nu_{\max}}$.  

However, this solution can be numerically unstable as it requires an inversion of the often degenerate covariance matrices.
We instead take the numerically stable approach laid out in Algorithm \ref{alg:stableCCA}, 
corresponding to the approach introduced by \cite{bjorck1973numerical}.  Here $\left[q,r,p\right] = \Call{QR}{\alpha}$ refers to a QR decomposition with pivoting such that $q r = \alpha_{\left(:, p\right)}$ where $q$ is an orthogonal matrix, $r$ is upper triangular matrix and $p$ is a column ordering defined implicitly such that $\left|r\left(i,i\right)\right| > \left|r\left(j,j\right)\right| \; \forall i < j$.  $\left[u, \Omega, z\right] = \Call{SVD}{\alpha}$ refers to a singular value decomposition such that $u \Omega z^T = \alpha$ where $u$ and $z$ are unitary matrix and $\Omega$ is diagonal matrix of singular values, with the ordering defined such that $\Omega\left(i,i\right) > \Omega \left(j,j\right) \; \forall i < j$.

After centring both inputs, a QR decomposition with pivoting is carried out on each. For example $qr=wp$ where $q$ is a unitary matrix, $r$ is an upper triangular matrix and $p$ is a pivot matrix such that the diagonal elements of $p$ are of decreasing magnitude.  If
\begin{align}
\label{eq:rankRedCCA}
\zeta = \max \left\{i : \left|r_{\left(i,i\right)}\right| > \varepsilon \left|r_{\left(1,1\right)}\right| \right\}
\end{align}
is the number of non-zero main diagonal terms within some tolerance $\varepsilon$, then the first $\zeta$ columns of $q$ will describe an orthonormal basis for the span of $w$.  Therefore by applying the reductions $q' \leftarrow q_{\left(\colon, 1:\zeta\right)}$ and $r' \leftarrow r_{\left(1:\zeta,1:\zeta\right)}$, then $q'r'$ will be a pivoted reduction of $w$ that is full rank and $r'$ will be invertible.  The algorithm then proceeds with the reduced matrices $q'$ and $r'$ for each input to carry out the CCA in a numerical stable manner.

Although for analytical application then the rank tolerance parameter $\varepsilon$ should be taken as $0^+$, we recommend taking a finite value (we use $\varepsilon = 10^{-4}$) to guard against numerical error and because this can act as a regularization term against individual splits overfitting the inputs.  Note that packaged applications of this algorithm are available, for example \textsc{canoncorr} in \textsc{matlab}, but in general these do not allow $\varepsilon$ to be set manually.

\begin{figure}[t]
	\vspace{-0.2em}
	\rule{\columnwidth}{1.3pt} 	
	\vspace{-1.8em}
\begin{algorithm}[H]
		\small
		\vspace{-0.5em}
		\captionsetup{labelfont=bf, justification=justified,singlelinecheck=false}
		\rule{\columnwidth}{0.5pt}
	\caption{Numerically Stable CCA \label{alg:stableCCA}}
	\begin{spacing}{1.2}
		\begin{algorithmic}[1]
			\renewcommand{\algorithmicrequire}{\textbf{Inputs:}}
			\renewcommand{\algorithmicensure}{\textbf{Outputs:}}	
			\Require First array $w \in \mathbb{R}^{N \times D}$, second array $v \in \mathbb{R}^{N \times K}$, tolerance parameter $\varepsilon \in \left[0^+, 1^-\right]$
			\Ensure Projection matrices for first and second arrays $A$ and $B$, correlations $\rho$
			\State $\mu^w = \frac{1}{N} \sum_{n=1:N}^{} w_{\left(n,:\right)}, \quad \mu^v = \frac{1}{N} \sum_{n=1:N}^{} v_{\left(n,:\right)}$ \Comment First centre the inputs
			\State $w_{\left(:, d\right)} \leftarrow w_{\left(:, d\right)}-\mu^w_{\left(d\right)} \quad \forall d \in \left\{1, \dots, D\right\}$
			\State $v_{\left(:, k\right)} \leftarrow v_{\left(:, k\right)}-\mu^v_{\left(k\right)} \quad \forall k \in \left\{1, \dots, K\right\}$
			\State $\left[q^w, r^w, p^w\right] = \Call{QR}{w}, \quad \left[q^v, r^v, p^v\right] = \Call{QR}{v}$ \Comment Carry out pivoted QR decompositions
			\State $\zeta^w = \max \{i : |r^w_{\left(i,i\right)}| < \varepsilon |r^w_{\left(1,1\right)}|\}, \quad
			\zeta^v = \max \{i : |r^v_{\left(i,i\right)}| < \varepsilon |r^v_{\left(1,1\right)}|\}$ 
			\State $q^w \leftarrow q^w_{\left(:,1:\zeta^w\right)}, \quad r^w \leftarrow r^w_{\left(1:\zeta^w ,1:\zeta^w\right)}, \quad q^v \leftarrow q^v_{\left(:,1:\zeta^v\right)}, \quad  r^v \leftarrow r^v_{\left(1:\zeta^v ,1:\zeta^v\right)}$ \Comment Reduce to full rank
			\State $\nu_{\max} = \min \left(\zeta^w, \zeta^v\right)$ \Comment Number of coefficient pairs that will be returned
			\If{$\zeta^w > \zeta^v$} \Comment Select which of two equivalent SVD decompositions is faster	
			\State $\left[u, \Omega, z\right] = \Call{SVD}{\left({q^w}\right)^T q^v}$
			\Else
			\State $\left[z, \Omega, u\right] = \Call{SVD}{\left({q^v}\right)^T q^w}$
			\EndIf
			\State $u \leftarrow u_{\left(:,1:\nu_{\max}\right)}, \quad z \leftarrow z_{\left(:,1:\nu_{\max}\right)}$ \Comment Remove meaningless components
			\State $A = \left({r^w}\right)^{-1} u, \quad B = \left({r^v}\right)^{-1} z$   \label{stableCCA:line:backsub} \Comment Using back substitution as $r^w$ and $r^v$ are upper triangular
			\State $\rho = \Call{diag}{\Omega}$
			\State $A_{\left(p^w, :\right)} \leftarrow \left[A^T, \mathbf{0}\right]^T$ \Comment Reorder rows to match corresponding columns in original matrices
			\State $B_{\left(p^v, :\right)} \leftarrow \left[B^T, \mathbf{0}\right]^T$ \Comment Some rows are set to zero if original matrix was not full rank
			\State \Return $A, B, \rho$
		\end{algorithmic}
				\vspace{-0.5em}
				\rule{\columnwidth}{0.5pt}
				\vspace{-2em}
	\end{spacing}
\end{algorithm}
\end{figure}
\begin{figure}[t]
	\vspace{-0.2em}
	\rule{\columnwidth}{1.3pt} 	
	\vspace{-1.8em}
	\begin{algorithm}[H]
		\small
		\captionsetup{labelfont=bf, justification=justified,singlelinecheck=false}
		\caption{CCF training algorithm (detailed version)\label{alg:CCFtrainDet}}
		\vspace{-0.5em}
		\rule{\columnwidth}{0.5pt}
		\begin{algorithmic}[1]
			\renewcommand{\algorithmicrequire}{\textbf{Inputs:}}
			\renewcommand{\algorithmicensure}{\textbf{Outputs:}}				 
			\Require ordinal features $X^{r} \in \mathbb{R}^{N \times D_r}$, categorical features $X^{c} \in \mathcal{S}^{N \times D_c}$, outputs $Y \in \mathbb{Y}^N$, 
			number of trees $L$ (default is $500$),
			number of features to sub-sample $\lambda \in \left\{1,\dots,D_r+D_c\right\}$ (default is  
			$\ceil*{\log_2 (D_r+D_c)+1}$),
			impurity measure $g$ (default is entropy as per~\eqref{eq:entropy} for classification
			and mean squared error as per~\eqref{eq:mse-gain} for regression), stopping criteria $c$ (see Section~\ref{sec:algorithm})
			\Ensure CCF $T$
			\State Convert $X^{c}$ to 1-of-K encoding $X^{b} \in \mathbb{I}^{N \times D_b}$
			\State $X = \left\{X^{r}, X^{b}\right\}$
			\For{$d \in 1:D_b+D_c$}
			\State $\mu_{\left(d\right)} = \frac{1}{N}\sum_{n=1}^{N} X_{\left(n,d\right)}$
			\Comment Calculate feature means and stand deviations (ignore missing terms) \label{train:line:pre1}
			\State $\sigma_{\left(d\right)}  = \sqrt{\frac{1}{N}\sum_{n=1}^{N} {{X_{\left(n,d\right)}}^2 - {\mu_{\left(d\right)}}^2}}$ 
			\State $X_{\left(:,d\right)} \leftarrow \left(X_{\left(:,d\right)}-\mu_{\left(d\right)}\right)/{\sigma_{\left(d\right)}}$
			\Comment Convert $X$ to $z$ scores
			\EndFor
			\If{$\lambda < \left(D_r+D_c\right)$} \Comment Don't projection bootstrap if $\lambda = \left(D_r+D_c\right)$
			\State $b=\mathrm{true}$ 
			\Else \State $\; b=\mathrm{false}$ 
			\EndIf
			\For{$i = 1 \colon L$}
			\State $X' \leftarrow X$
			\State Randomly assign each missing value in $X'$ to an independent draw from a unit normal
			\If{$b$} 
			\State $Y'\leftarrow Y$ 
			\Else
			\State $\; \left\{X', Y'\right\} \leftarrow $ bootstrap sample $N$ rows from $\{X', Y\}$ 
			\EndIf
			\State $ t_i \leftarrow \Call{growTree}{0, X', Y', \left\{1,\dots,D_r+D_c\right\}, \lambda, g, b, c}$
			\EndFor
			\State \Return $T = \left\{t_i \right\}_{i = 1 \dots L}$
		\end{algorithmic}
		\vspace{-0.5em}
		\rule{\columnwidth}{0.5pt}
		\vspace{-2em}
	\end{algorithm}
	\end{figure}
	
\section{Detailed Training Algorithms}
\label{sec:detail-alg}

In this section we provide more detailed versions of the training Algorithms~\ref{alg:CCFtrain}
and~\ref{alg:growTree} given in Algorithms~\ref{alg:CCFtrainDet} and~\ref{alg:growTreeDet} respectively.
These differ by providing explicit consideration for categorical features, missing data, and
edge cases (e.g. when a node only has two unique nodes).	
	\begin{figure}[p]
	\vspace{-0.2em}
	\rule{\columnwidth}{1.3pt} 	
	\vspace{-1.8em}	
	\begin{algorithm}[H]
		\small
		\captionsetup{labelfont=bf, justification=justified,singlelinecheck=false}
		\caption{\textsc{GrowTree} (detailed version) \label{alg:growTreeDet}}
		\vspace{-0.5em}
		\rule{\columnwidth}{0.5pt}
		\begin{algorithmic}[1]	
			\renewcommand{\algorithmicrequire}{\textbf{Inputs:}}
			\renewcommand{\algorithmicensure}{\textbf{Outputs:}}
			\Require unique node identifier for root node $j$, $X^j = X_{(\omega_j,:)} \in \mathbb{R}^{N_j \times \left(D_r+D_b\right)}$, 
			$Y^j = Y_{(\omega_j,:)} \in \mathbb{Y}^{N_j}$, available feature ids $D_j \subseteq \left\{1,\dots,D_r+D_c\right\}$, $\lambda$, $g$, whether to projection bootstrap $b$, $c$
			\Ensure subtree $\{\Psi_j,\Theta_j\}$ where $\Psi_j$ are discriminant nodes and $\Theta_j$ leaf nodes
			\State Sample $\delta_j \subseteq D_j$ by taking $\min \left(\lambda, \left|D_j\right|\right)$ samples without replacement from $D_j$
			\State While $\delta_j$ contains features without variation, eliminate these from $D_j$ and $\delta_j$ and resample
			\State $\gamma_j = \delta_j$ mapped to the column indices of $X^j$ in accordance with the 1-of-K encoding of $X^c$
			\If {$b$} 
			\State $\left\{\mathcal{X}', \mathcal{Y}'\right\} \leftarrow $ bootstrap sample $N_j$ data points from $\{X^j_{\left(:,\gamma\right)}, Y^j\}$ 
			\Else 
			\State $\left\{\mathcal{X}',\mathcal{Y}'\right\} \leftarrow \left\{X^j_{\left(:,\gamma\right)},Y^j\right\}$  
			\EndIf
			\If{all rows in $\mathcal{X}'$ or $\mathcal{Y}'$ are identical}
			\IIf{all rows in $\mathcal{X}^j_{\left(:,\gamma\right)}$ or $Y^j$ are identical} 
			\Return $\left\{\cdot,\left\{j,\frac{1}{N_j} \sum_{n=1}^{N_j} Y^j_{(n,:)}\right\}\right\}$
			\EndIIf
			\State $\left\{\mathcal{X}', \mathcal{Y}'\right\} \leftarrow \{X^j_{\left(:,\gamma\right)}, Y^j\}$ 
			\EndIf
			\If{$\mathcal{X}'$ contains only two unique rows}
			\State $\tilde{\mathcal{X}} = \Call{UniqueRows}{X'}$
			\State ${\phi_j} \leftarrow \tilde{\mathcal{X}}_{\left(2, :\right)} - \tilde{\mathcal{X}}_{\left(1, :\right)}$
			\State
			$s_j \leftarrow \frac{1}{2} \left(\tilde{\mathcal{X}}_{\left(1, :\right)} + \tilde{\mathcal{X}}_{\left(2, :\right)}\right) \phi_j$
			\label{growTree:line:twopoints}
			\Else				
			\State $\left\{\Phi, \Omega\right\} \leftarrow \Call{CCA}{\mathcal{X}', \mathcal{Y}'}$ \Comment Calculate CCA coefficients using bootstrap sample
			\State $U \leftarrow \mathcal{X} \Phi$ \Comment Project \textit{original features} into canonical component space
			\State $G_{\mathrm{base}} \leftarrow g\left(Y^j \right)$
			\For{$\nu\in1:\nu_{\max}$} \Comment $\nu_{\max}= \min \left(\mathrm{rank}\left(\mathcal{X}'\right),\mathrm{rank}\left(\mathcal{Y}'\right) \right)$ is number of canonical coefficient pairs
			\State $u \leftarrow \Call{Sort}{U_{(:,\nu)}}$
			\For{$i\in2:N_j$} \Comment Exhaustive search on unique splits
			\State $S_{(i,\nu)} \leftarrow (u_{(i-1)}+u_{(i)})/2$ \Comment Split halfway between consecutive points.
\State $\tau^{\ell} \leftarrow \left\{n \in \left\{1,\dots,N_j\right\} : U_{(n,\nu)} \le S_{(i,\nu)}\right\}$
\State $N_{\chi_{j,\ell}}\leftarrow$ number of elements in $\tau^{\ell}$
\State $\tau^{r} \leftarrow \left\{1,\dots,N_j\right\} \backslash \tau^{\ell}$
\State $G_{(i,\nu)} \leftarrow G_{\mathrm{base}}
-\frac{N_{\chi_{j,\ell}}}{N_j} g\left(Y_{(\tau^{\ell} ,:)}\right)
-\frac{N_j-N_{\chi_{j,\ell}}}{N_j} g\left(Y_{(\tau^{r} ,:)}\right)$  \Comment Split gain
			\EndFor \Comment In practise gain is calculated by iterating from previous value to avoid $N_j^2$ complexity. \EndFor
			\State $\left\{i^*, \nu^*\right\} \leftarrow \argmax_{i,\nu} G_{(i,\nu)}$ \Comment Choose best split
			\If{$G_{(i^*, \nu^*)} \le 0$ or any of $c$ satisfied} \Comment Node is a leaf
			\State $\theta_j \leftarrow \frac{1}{N_j} \sum_{n=1}^{N_j} Y^j_{(n,:)}$ 
			\State \Return $\left\{\cdot,\left\{j,\theta_j\right\}\right\}$
			\EndIf
			\State $\phi_j \leftarrow \Phi_{(:,\nu^*)}, \quad s_j \leftarrow S_{(i^*, \nu^*)}$ \Comment Chosen projection and split point
			\EndIf 
			\State Generate unique identifiers for children $\chi_{j,\ell}$ and $\chi_{j,r}$
			\State $\psi_j \leftarrow \left\{j, \delta_j,  \phi_j, s_j, \chi_{j,\ell}, \chi_{j,r} \right\}$ 
			\State $\tau^{\ell} \leftarrow \left\{n \in \left\{1,\dots,N_j\right\} : U_{(n,\nu^*)} \le S_{(i^*,\nu^*)}\right\}, \;\;
			\tau^{r} \leftarrow \left\{1,\dots,N_j\right\} \backslash \tau^{\ell}$
			\Comment Assign datapoints to children
			\State $\left\{\Psi_{\ell},\Theta_{\ell}\right\} \leftarrow 
			\Call{growTree}{\chi_{j,\ell}, X^j_{(\tau^{\ell},:)}, Y^j_{(\tau^{\ell},:)}, D_j,\lambda, g,b,c}$
			\Comment Recurse for left child and right child
			\State $\left\{\Psi_{r},\Theta_{r}\right\} \leftarrow 
			\Call{growTree}{\chi_{j,r}, X^j_{(\tau^{r},:)}, Y^j_{(\tau^{r},:)}, D_j,\lambda, g,b,c}$
			\State \Return $\left\{\psi_j \cup \Psi_{\ell} \cup \Psi_r,  \Theta_{\ell} \cup \Theta_{r}\right\}$
		\end{algorithmic}
		\vspace{-0.5em}
		\rule{\columnwidth}{0.5pt}
		\vspace{-2em}
	\end{algorithm} 
\end{figure}

\newpage

\section{Inverted Cross Validation}
\label{sec:invCrossVal}

To investigate the performance of CCFs on datasets containing few data points relative to the complexity of the underlying structure, we carried out inverted cross validations, training on one fold and testing on the other nine.  15 such tests were carried out, using the same folds as the original cross validation.  Table \ref{table:InvertedResultsTable} gives the results on each of the individual datasets, Table \ref{table:InvertedVics} gives a summary of the significant victories and losses, and Table~\ref{table:rankInverted} shows the average rank.  The parameters used were the same as for the standard cross validation except that only 200 trees were used.

The comparative performances were overall similar to the standard cross-validation case, though there
were a number of relative changes for individual datasets.  
Two results of particular note were that all methods performed worse than predicting the most popular class for the \textit{fertility} and \textit{wholesale-r} datasets, which gives misclassification rates of 12.0\% and 28.2\% respectively.  This suggests that it may be beneficial to incorporate information about the overall class ratios into the training algorithms.

\begin{table*}[p]                        
	\centering    
	\scriptsize             
	\setlength\tabcolsep{3.25pt}	  
	\renewcommand{\arraystretch}{1.2}      
	\caption{Mean and standard deviations of percentage of test cases misclassified for inverted cross validation.  Method with best accuracy is shown in bold. $\bullet$ and $\circ$ indicate that CCFs were significantly better and worse respectively at the 1\% level of a Wilcoxon signed rank test. \label{table:InvertedResultsTable}}                              
	\begin{tabular}{|c|c|cc|cc|cc|cc|cc|}                    
		\hline                                                       
		Dataset & CCF & RF & & Rot-For & & CCF-Bag & & RRF & & RRF-Bag &\\
		\hline           
Balance scale & 14.86 $\pm$ 2.54 & 21.02 $\pm$ 2.74 & $\bullet$ & 15.27 $\pm$ 2.52 & $\bullet$ & \textbf{14.08 $\pm$ 2.59} & $\circ$ & 16.10 $\pm$ 2.32 &  & 14.75 $\pm$ 2.21 &  \\                        
Banknote & \textbf{0.64 $\pm$ 0.57} & 3.96 $\pm$ 1.66 & $\bullet$ & 0.89 $\pm$ 0.70 & $\bullet$ & 0.78 $\pm$ 0.58 & $\bullet$ & 1.16 $\pm$ 0.93 & $\bullet$ & 1.46 $\pm$ 0.96 & $\bullet$ \\               
Breast tissue & \textbf{50.51 $\pm$ 8.19} & 52.06 $\pm$ 7.72 & $\bullet$ & 51.03 $\pm$ 8.10 &  & 54.76 $\pm$ 8.87 & $\bullet$ & 50.70 $\pm$ 7.11 &  & 52.05 $\pm$ 7.64 &  \\                               
Climate crashes & 7.23 $\pm$ 0.65 & 7.22 $\pm$ 0.51 &  & 7.22 $\pm$ 0.79 &  & 7.22 $\pm$ 0.54 &  & \textbf{7.20 $\pm$ 0.53} &  & 7.21 $\pm$ 0.48 &  \\                                                     
Fertility & 18.97 $\pm$ 8.16 & 14.20 $\pm$ 5.36 & $\circ$ & 15.38 $\pm$ 5.92 & $\circ$ & 16.76 $\pm$ 7.20 & $\circ$ & 15.73 $\pm$ 5.05 & $\circ$ & \textbf{13.28 $\pm$ 2.88} & $\circ$ \\                  
Heart-SPECT & 20.63 $\pm$ 3.08 & 20.76 $\pm$ 3.06 &  & 20.09 $\pm$ 3.22 & $\circ$ & 20.04 $\pm$ 2.83 & $\circ$ & 19.10 $\pm$ 2.50 & $\circ$ & \textbf{18.88 $\pm$ 2.51} & $\circ$ \\                       
Heart-SPECTF & 21.11 $\pm$ 1.87 & 20.79 $\pm$ 1.93 & $\circ$ & 20.93 $\pm$ 2.05 & $\circ$ & 21.38 $\pm$ 2.56 &  & \textbf{20.60 $\pm$ 1.99} & $\circ$ & 20.61 $\pm$ 1.74 & $\circ$ \\                      
Hill valley & \textbf{0.14 $\pm$ 0.39} & 48.34 $\pm$ 1.75 & $\bullet$ & 7.18 $\pm$ 1.77 & $\bullet$ & 0.17 $\pm$ 0.45 &  & 34.89 $\pm$ 3.69 & $\bullet$ & 37.71 $\pm$ 2.94 & $\bullet$ \\                  
Hill valley noisy & \textbf{20.37 $\pm$ 4.07} & 49.27 $\pm$ 1.48 & $\bullet$ & 22.22 $\pm$ 3.50 & $\bullet$ & 21.63 $\pm$ 4.57 & $\bullet$ & 41.16 $\pm$ 2.75 & $\bullet$ & 42.90 $\pm$ 2.34 & $\bullet$ \\
ILPD & 30.65 $\pm$ 1.79 & 30.68 $\pm$ 1.71 &  & \textbf{30.14 $\pm$ 1.52} & $\circ$ & 30.29 $\pm$ 1.62 & $\circ$ & 31.14 $\pm$ 1.70 & $\bullet$ & 30.43 $\pm$ 1.63 &  \\                                   
Ionosphere & \textbf{11.90 $\pm$ 4.05} & 13.90 $\pm$ 4.44 & $\bullet$ & 12.95 $\pm$ 4.28 & $\bullet$ & 16.78 $\pm$ 4.82 & $\bullet$ & 12.81 $\pm$ 4.34 & $\bullet$ & 15.45 $\pm$ 5.26 & $\bullet$ \\       
Iris & 5.93 $\pm$ 3.83 & 7.59 $\pm$ 4.31 & $\bullet$ & 9.74 $\pm$ 5.64 & $\bullet$ & \textbf{5.65 $\pm$ 4.13} &  & 8.35 $\pm$ 4.71 & $\bullet$ & 9.42 $\pm$ 4.98 & $\bullet$ \\                            
Landsat satellite & 11.58 $\pm$ 0.41 & 12.07 $\pm$ 0.49 & $\bullet$ & \textbf{11.41 $\pm$ 0.45} & $\circ$ & 12.19 $\pm$ 0.42 & $\bullet$ & 11.60 $\pm$ 0.46 &  & 12.10 $\pm$ 0.46 & $\bullet$ \\           
Letter & \textbf{10.14 $\pm$ 0.47} & 13.46 $\pm$ 0.57 & $\bullet$ & 11.14 $\pm$ 0.48 & $\bullet$ & 11.22 $\pm$ 0.48 & $\bullet$ & 11.80 $\pm$ 0.49 & $\bullet$ & 13.24 $\pm$ 0.54 & $\bullet$ \\           
Libras & \textbf{49.86 $\pm$ 4.83} & 61.57 $\pm$ 4.31 & $\bullet$ & 55.27 $\pm$ 4.82 & $\bullet$ & 52.30 $\pm$ 4.66 & $\bullet$ & 53.94 $\pm$ 4.67 & $\bullet$ & 56.17 $\pm$ 4.82 & $\bullet$ \\           
MAGIC & \textbf{13.46 $\pm$ 0.22} & 14.03 $\pm$ 0.28 & $\bullet$ & 14.19 $\pm$ 0.29 & $\bullet$ & 13.55 $\pm$ 0.23 & $\bullet$ & 14.41 $\pm$ 0.26 &  & 14.51 $\pm$ 0.27 &  \\                              
Nursery & 3.15 $\pm$ 0.34 & 3.97 $\pm$ 0.43 & $\bullet$ & \textbf{2.89 $\pm$ 0.39} & $\circ$ & 3.67 $\pm$ 0.36 & $\bullet$ & 3.86 $\pm$ 0.39 & $\bullet$ & 4.48 $\pm$ 0.42 & $\bullet$ \\                  
ORL & \textbf{54.95 $\pm$ 3.80} & 60.35 $\pm$ 4.04 & $\bullet$ & - &  & 61.22 $\pm$ 3.74 & $\bullet$ & 55.74 $\pm$ 3.81 & $\bullet$ & 61.20 $\pm$ 4.13 & $\bullet$ \\                          
Optical digits & \textbf{3.54 $\pm$ 0.38} & 4.79 $\pm$ 0.47 & $\bullet$ & 3.91 $\pm$ 0.42 & $\bullet$ & 3.93 $\pm$ 0.38 & $\bullet$ & 3.91 $\pm$ 0.42 & $\bullet$ & 4.42 $\pm$ 0.45 & $\bullet$ \\         
Parkinsons & \textbf{17.06 $\pm$ 4.05} & 18.75 $\pm$ 4.02 & $\bullet$ & 18.55 $\pm$ 4.43 & $\bullet$ & 18.24 $\pm$ 4.62 & $\bullet$ & 17.64 $\pm$ 4.11 & $\bullet$ & 18.35 $\pm$ 4.23 & $\bullet$ \\       
Pen digits & \textbf{1.43 $\pm$ 0.22} & 2.97 $\pm$ 0.35 & $\bullet$ & 1.84 $\pm$ 0.27 & $\bullet$ & 1.67 $\pm$ 0.24 & $\bullet$ & 1.81 $\pm$ 0.26 & $\bullet$ & 2.15 $\pm$ 0.29 & $\bullet$ \\             
Polya & 23.80 $\pm$ 0.51 & 24.16 $\pm$ 0.61 & $\bullet$ & \textbf{23.08 $\pm$ 0.50} & $\circ$ & 23.84 $\pm$ 0.50 &  & 24.37 $\pm$ 0.67 & $\bullet$ & 25.04 $\pm$ 0.75 & $\bullet$ \\                       
Seeds & 8.84 $\pm$ 3.14 & 13.34 $\pm$ 3.21 & $\bullet$ & 11.71 $\pm$ 3.70 & $\bullet$ & \textbf{8.53 $\pm$ 3.21} & $\circ$ & 10.50 $\pm$ 2.53 & $\bullet$ & 10.34 $\pm$ 2.83 & $\bullet$ \\                
Skin seg & 0.06 $\pm$ 0.01 & 0.13 $\pm$ 0.02 & $\bullet$ & 0.10 $\pm$ 0.01 & $\bullet$ & 0.06 $\pm$ 0.01 &  & \textbf{0.06 $\pm$ 0.01} &  & 0.07 $\pm$ 0.01 & $\bullet$ \\                                 
Soybean & \textbf{18.96 $\pm$ 3.95} & 21.99 $\pm$ 4.00 & $\bullet$ & 21.22 $\pm$ 4.73 & $\bullet$ & 20.23 $\pm$ 4.06 & $\bullet$ & 19.20 $\pm$ 4.01 &  & 20.84 $\pm$ 3.98 & $\bullet$ \\                   
Spirals & \textbf{0.68 $\pm$ 0.16} & 3.29 $\pm$ 0.49 & $\bullet$ & 3.80 $\pm$ 0.57 & $\bullet$ & \textbf{0.68 $\pm$ 0.16} &  & 0.75 $\pm$ 0.16 & $\bullet$ & 0.75 $\pm$ 0.16 & $\bullet$ \\                
Splice & 7.53 $\pm$ 1.75 & \textbf{4.81 $\pm$ 0.68} & $\circ$ & 6.72 $\pm$ 1.11 & $\circ$ & 9.06 $\pm$ 2.12 & $\bullet$ & 22.34 $\pm$ 4.99 & $\bullet$ & 27.43 $\pm$ 5.62 & $\bullet$ \\                   
Vehicle & \textbf{26.02 $\pm$ 2.14} & 32.76 $\pm$ 2.66 & $\bullet$ & 27.79 $\pm$ 2.31 & $\bullet$ & 26.78 $\pm$ 2.26 & $\bullet$ & 30.76 $\pm$ 2.47 & $\bullet$ & 32.29 $\pm$ 2.49 & $\bullet$ \\          
Vowel-c & \textbf{40.71 $\pm$ 3.17} & 46.42 $\pm$ 3.00 & $\bullet$ & 41.63 $\pm$ 2.87 & $\bullet$ & 41.46 $\pm$ 3.44 & $\bullet$ & 50.26 $\pm$ 2.74 & $\bullet$ & 52.28 $\pm$ 2.71 & $\bullet$ \\          
Vowel-n & \textbf{34.15 $\pm$ 2.99} & 41.52 $\pm$ 2.94 & $\bullet$ & 35.82 $\pm$ 3.16 & $\bullet$ & 35.37 $\pm$ 2.92 & $\bullet$ & 37.42 $\pm$ 3.09 & $\bullet$ & 39.74 $\pm$ 3.16 & $\bullet$ \\          
Waveform (1) & 14.76 $\pm$ 0.41 & 16.52 $\pm$ 0.53 & $\bullet$ & 14.91 $\pm$ 0.42 & $\bullet$ & \textbf{14.66 $\pm$ 0.41} & $\circ$ & 14.81 $\pm$ 0.40 &  & 14.74 $\pm$ 0.45 &  \\                         
Waveform (2) & 14.83 $\pm$ 0.48 & 16.24 $\pm$ 0.49 & $\bullet$ & 14.83 $\pm$ 0.49 &  & \textbf{14.70 $\pm$ 0.47} & $\circ$ & 15.66 $\pm$ 0.62 & $\bullet$ & 15.84 $\pm$ 0.73 & $\bullet$ \\                
Wholesale-c & 11.41 $\pm$ 2.14 & 10.82 $\pm$ 2.00 & $\circ$ & \textbf{10.58 $\pm$ 1.84} & $\circ$ & 11.16 $\pm$ 2.13 & $\circ$ & 11.42 $\pm$ 2.16 &  & 11.19 $\pm$ 1.93 &  \\                              
Wholesale-r & 35.64 $\pm$ 4.09 & 33.38 $\pm$ 3.66 & $\circ$ & \textbf{31.23 $\pm$ 3.59} & $\circ$ & 33.08 $\pm$ 3.81 & $\circ$ & 36.74 $\pm$ 4.10 & $\bullet$ & 33.14 $\pm$ 3.76 & $\circ$ \\              
Wisconsin & 4.12 $\pm$ 1.02 & 4.31 $\pm$ 0.98 & $\bullet$ & 3.53 $\pm$ 0.69 & $\circ$ & 4.10 $\pm$ 1.03 &  & 3.55 $\pm$ 0.70 & $\circ$ & \textbf{3.43 $\pm$ 0.63} & $\circ$ \\                      
Yeast & 46.12 $\pm$ 1.86 & 45.98 $\pm$ 1.91 &  & 45.32 $\pm$ 1.82 & $\circ$ & 45.37 $\pm$ 1.83 & $\circ$ & 46.01 $\pm$ 1.90 &  & \textbf{44.96 $\pm$ 1.69} & $\circ$ \\                                    
Zoo & \textbf{23.43 $\pm$ 10.80} & 25.36 $\pm$ 11.26 & $\bullet$ & 24.10 $\pm$ 10.66 &  & 24.12 $\pm$ 11.15 & $\bullet$ & 23.75 $\pm$ 10.65 &  & 24.75 $\pm$ 11.07 & $\bullet$ \\ 
		\hline                                         
	\end{tabular}                                                      
\end{table*}  

\begin{table}[h!]	
	\renewcommand{\arraystretch}{1.2}
	\footnotesize
	\centering
	\caption{Number of victories column vs row at 1\% significance level of Wilcoxon signed rank test for inverted cross-validation.
		Victories and losses for CCFs are shown in blue and red respectively.
		 \label{table:InvertedVics}}
	\begin{tabular}{lcccccc} \toprule
		& \textbf{\color{blue} CCF} & \text{RF} & \text{Rotation Forest} & \text{CCF-Bag} & \text{RRF} & \text{RRF-Bag}
		\\ \midrule
		\textbf{\color{red} CCF} & - & \textbf{\color{red} 5} & \textbf{\color{red} 12} & \textbf{\color{red} 10} 
								& \textbf{\color{red} 4} & \textbf{\color{red} 6} \\  
		\text{RF} & \textbf{\color{blue} 28} & - & 26 & 26 & 23 & 19\\  
		\text{Rotation Forest} & \textbf{\color{blue} 20} & 5 & - & 15 & 10 & 7\\
		\text{CCF-Bag} & \textbf{\color{blue} 19} & 8 & 12 & - & 8 & 7 \\
		\text{RRF} & \textbf{\color{blue} 22} & 7 & 16 & 20 & - & 8 \\
		\text{RRF-Bag} & \textbf{\color{blue} 24} & 8 & 20 & 20 & 22 & - \\
		  		  \bottomrule
	\end{tabular}
\end{table}

\begin{table}[t]
	\renewcommand{\arraystretch}{1.2}
	\footnotesize
	\centering
	\caption{Mean rank across datasets of mean test misclassification rate for inverted cross-validation.  Lower
		rank corresponds to better performance (average rank $=3.5$). \textit{ORL} dataset is
		omitted from comparison to avoid biasing against rotation forests. \label{table:rankInverted}}
	\begin{tabular}{lc} \toprule
		\text{Algorithm} & \text{Mean Rank}  \\ \midrule
		\text{CCF} & 2.51 \\
		\text{CCF-Bag} & 2.89 \\  
		\text{Rotation Forest} & 3.14 \\  
		\text{RRF} &  3.55 \\ 
		\text{RRF-Bag} &  4.03 \\ 
		\text{RF} & 4.88 \\  \bottomrule
	\end{tabular}
\end{table}

\begin{figure}[t]
	\centering
	\begin{subfigure}[t]{0.46\textwidth}
		\caption{Letter \label{fig:q-letter}}
		\includegraphics[width=0.99\textwidth,trim={0 0 0 3cm},clip]{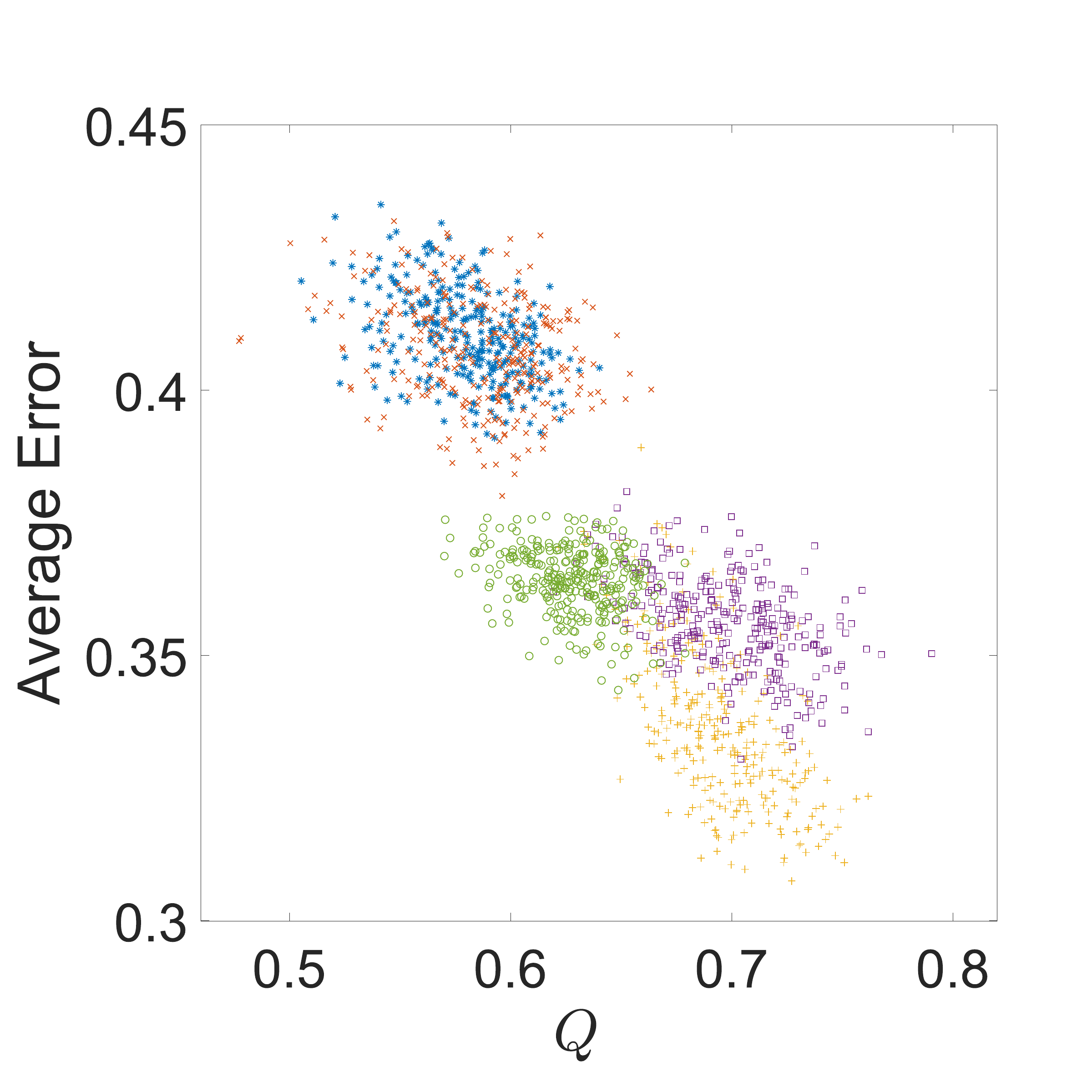}
		\centering
	\end{subfigure}
	~~~~
	\begin{subfigure}[t]{0.46\textwidth}
		\caption{Pen Digits \label{fig:q-pen}}
		\includegraphics[width=0.99\textwidth,trim={0 0 0 3cm},clip]{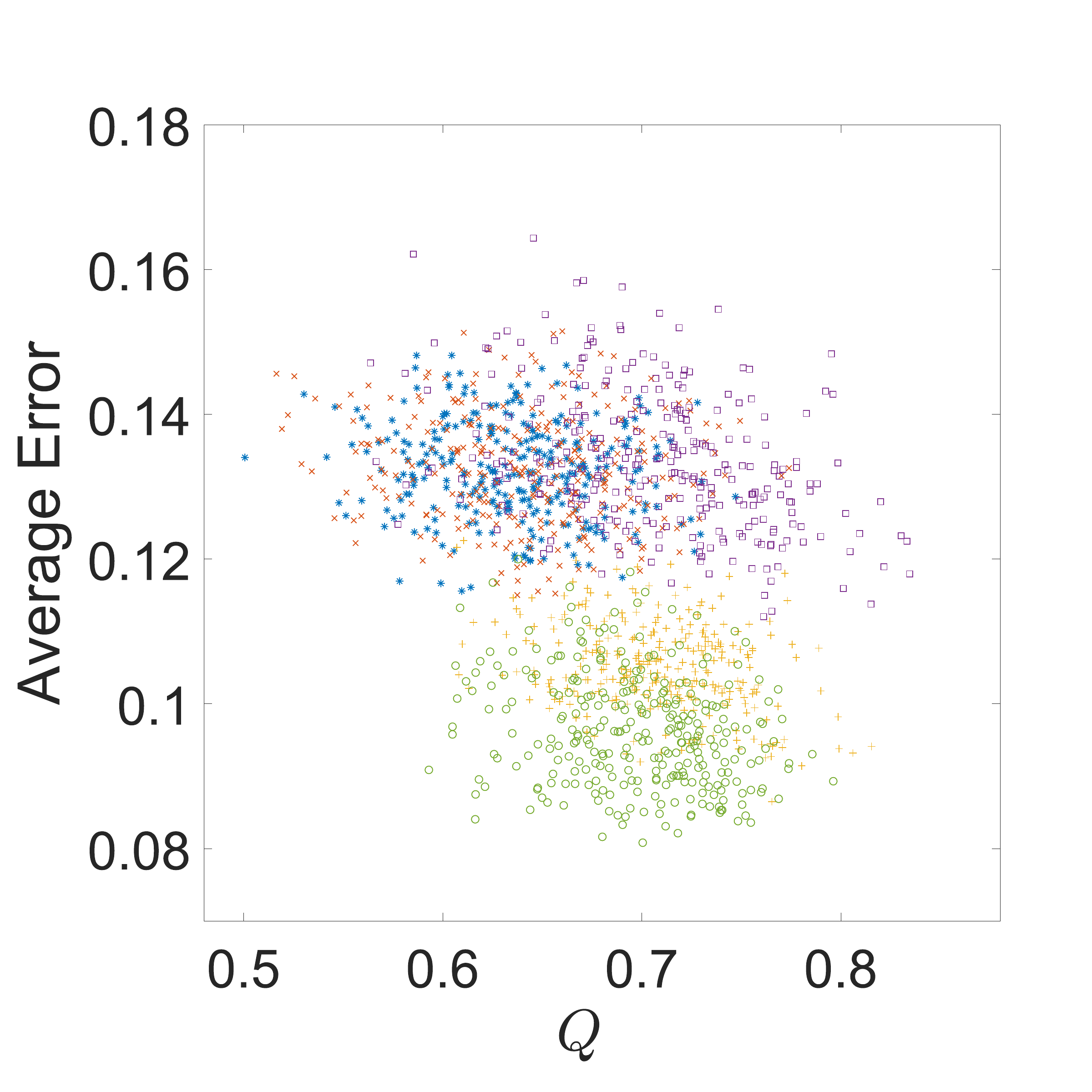}
		\centering
	\end{subfigure}
	\vspace{10pt} \\
	\begin{subfigure}[t]{0.45\textwidth}
		\caption{Skin Segmentation \label{fig:q-skin}}
		\includegraphics[width=0.99\textwidth,trim={0 0 0 1.5cm},clip]{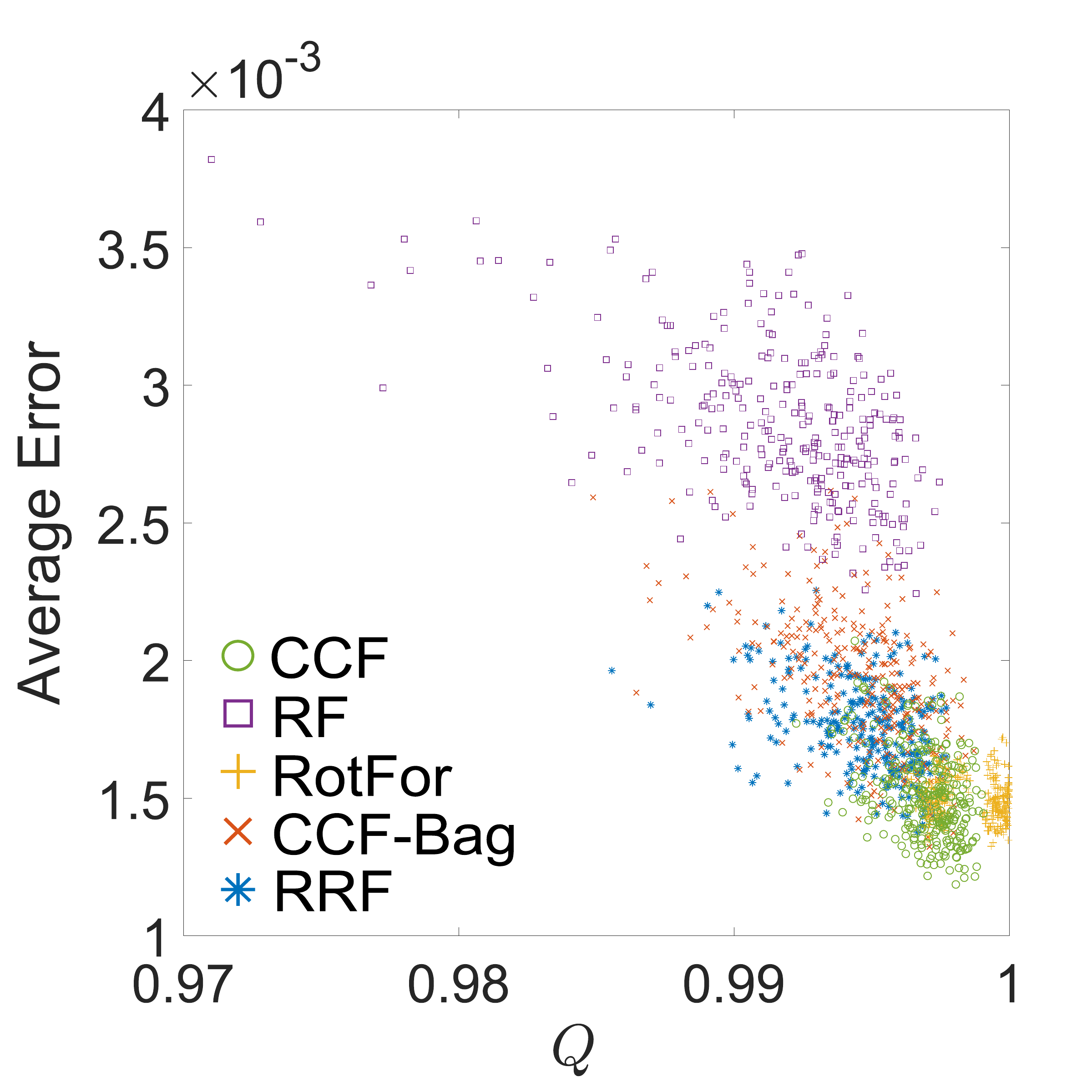}
		\centering
	\end{subfigure}	
	~~~~
	\begin{subfigure}[t]{0.46\textwidth}
		\caption{Splice \label{fig:q-splice}}
		\includegraphics[width=0.99\textwidth,trim={0 0.5 0 2cm},clip]{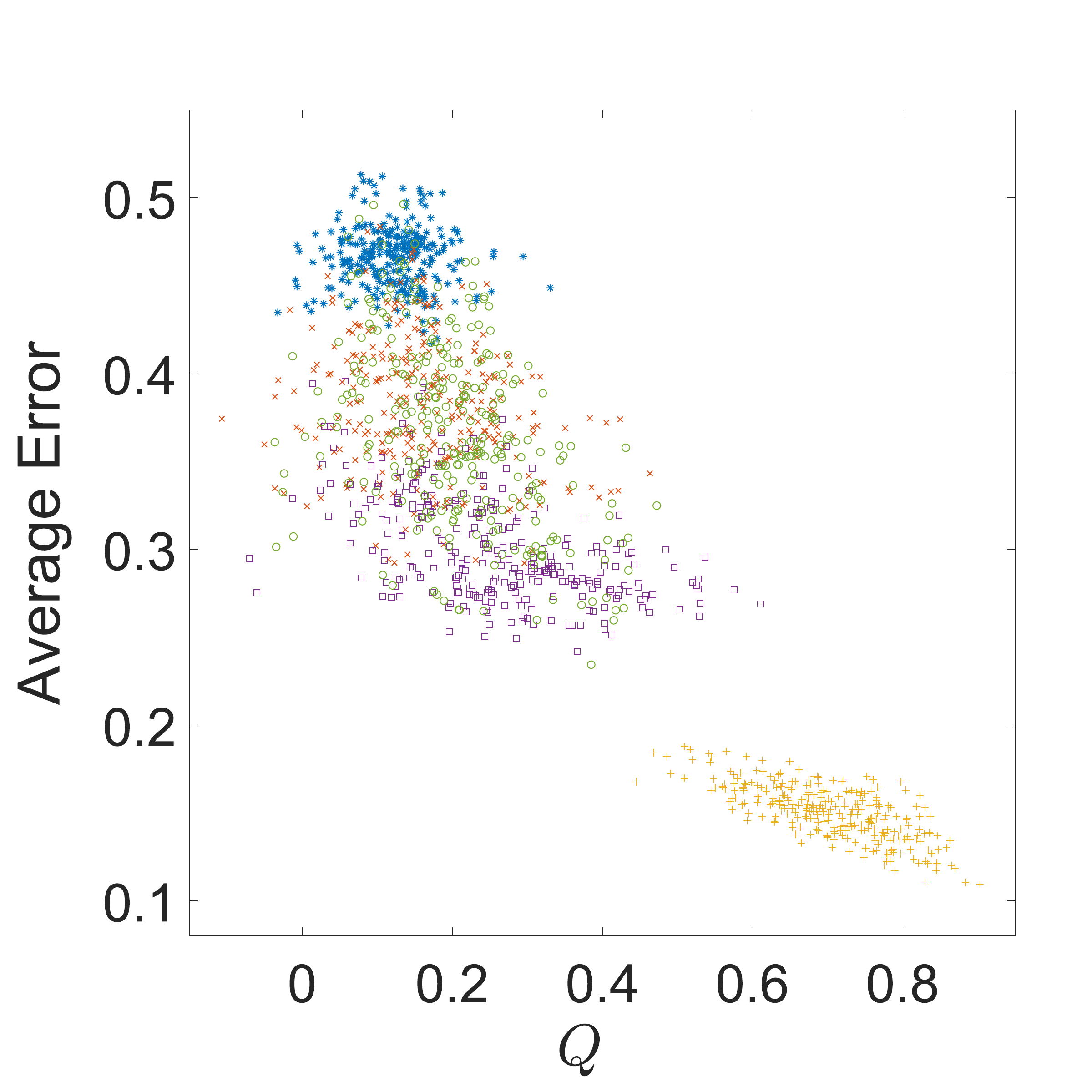}
		\centering
	\end{subfigure}
	\caption{Diversity-error plots for four different datasets.  Each point represents
		a pair of trees in the ensemble of $25$ trees.  $Q$ is a diversity measure as 
		defined in~\ref{eq:Q}.  The error corresponds
		to the average test misclassification rate of the two trees.
		\label{fig:kappa-diags}}
\end{figure}

\section{Trade-off between Tree Diversity and Accuracy}
\label{sec:diversity}

As we asserted in the introduction, the two key factors in the predictive accuracy of any decision tree
ensemble method are the predictive accuracy of the individual trees and the diversity
of the tree predictions.  In this section, we investigate whether the improvements
of CCFs over alternatives originate from one or both of these factors.  To do so we will construct
so-called diversity-error diagrams~\citep{margineantu1997pruning} for the different methods.  
These consider the error and similarity between the predictions between of pairs of trees.  Given
an ensemble of $L$ trees, $L(L-1)/2$ such pairwise comparisons can be made, and therefore a
scatter plot of accuracy vs diversity for pairs of trees
can be constructed, providing a representation for the ensemble as a whole.
As recommended by~\cite{kuncheva2003measures}, we use as our
diversity metric the $Q$ statistic \citep{yule1900association}.
Given a pair of classifiers, then if $E^{11}$, $E^{10}$, $E^{01}$, and $E^{00}$ are respectively 
the number test instances correctly classified by both classifiers, only the first classifier, only
the second classifier, and neither classifier respectively, then the $Q$ statistic is defined as
\begin{align}
\label{eq:Q}
Q = \frac{E^{11}E^{00}-E^{01}E^{10}}{E^{11}E^{00}+E^{01}E^{10}}.
\end{align}
The possible values of $Q$ vary between $-1$ and $+1$ with the former corresponding
to classifiers that \textit{never} make the same error (or are never correct at the same time), the
latter to classifiers that \textit{always} make the same error, and $Q=0$ indicating independence.
We can therefore use $Q$ as a measure of diversity (with lower $Q$ indicating more diversity), 
as it characterizes how often pairs of classifiers have the same errors relative to how often they have 
different errors.

Diversity-error diagrams for CCFs, RFs, rotation forests, CCF-Bag, and RRFs on four different 
datasets are shown in Figure~\ref{fig:kappa-diags}.
These are based on training trees using $10\%$ of the data an then calculating the average
error and $Q$ statistic for each pair of trees using the remaining $90\%$ as per the inverted
cross validation carried out in Appendix~\ref{sec:invCrossVal}.  This inverted
train/test split was used to ensure there were enough test points to calculate stable $Q$ values.
Note that, by necessity, the diagrams are produced by only a single train/test split, but for the presented
datasets then qualitative variations in the generated plots were relatively small.

For the \textit{letter} and \textit{pen digits} datasets, Table~\ref{table:InvertedResultsTable} shows that CCFs
outer-performed all the other approaches in predictive accuracy.  However, Figures~\ref{fig:q-letter}
and~\ref{fig:q-pen} suggest these advantages may stem from different causes.  Figure~\ref{fig:q-pen}
shows a similarly diversity for all the approaches on the \textit{pen digits} dataset, but a better tree
accuracy for CCFs than the  others.  Figure~\ref{fig:q-letter}  suggests that CCFs may have outperformed
CCF-Bag and RRFs on the \textit{letter} dataset due to having better individual tree accuracy, but the
advantages compared with RFs now seem to stem from better diversity.  Rotation forests had a better
tree accuracy but worse diversity than CCFs, with the better ensemble performance of CCFs suggesting 
they had a better trade-off for this dataset.

For the \textit{skin segmentation} dataset then Table~\ref{table:InvertedResultsTable} shows that
CCFs outperformed RFs and rotation forests, but had similar performance to CCF-Bag and RRFs.
Figure~\ref{fig:q-skin} suggests that the advantage over RFs comes from better tree accuracy and the
advantage over rotation forests from better diversity.  It shows that CCFs gave more accurate but
less diverse trees than CCF-Bag and RRFs.

For the \textit{splice} dataset then CCFs outperformed CCF-Bag and RRFs (with the latter doing particularly
poorly), but were outperformed by RFs and rotation forests.  From Figure~\ref{fig:q-splice}, the poor
tree accuracy of RRFs is very apparent, while RFs seem to give better accuracy than CCFs, at the expense
of slightly less diversity.  Rotation forests, which were outperformed by RFs, gave significantly more 
accurate trees than all the other approaches, but far less diversity.

Other than suggesting a tendency for rotation forests to produce relatively accurate but low diversity
trees, the diagrams do not appear to show a clear pattern of differences.  
This suggests that the advantages
of CCFs can originate from either improved tree accuracy or diversity depending on the dataset.

\section{Detailed Results for Comparisons to Other Classifiers}
\label{sec:detailedResults}

Table \ref{table:detailedCCFandTreeBagger} gives a detailed comparison of CCFs to \textsc{matlab}'s \textsc{TreeBagger} algorithm, along with a comparison to a summary of the performance of the 179 classifiers tested in \cite{fernandez2014we} for 82 UCI datasets.  Both CCFs and \textsc{TreeBagger} used 500 trees and the
default tree parameter settings laid out in Section~\ref{sec:param}.  For significance comparisons with the \textsc{TreeBagger} algorithm a 10\% significance level was necessary as the Wilcoxon signed rank test cannot give a p-value lower than 6.25\% for 4-fold cross validation.  This gave a win/draw/loss ratio for CCFs of 22/56/4.  It should be noted that the number of losses is no more than would be expected by chance and that the reason that there are many more draws than in the tests of Section 3.1 is that a many fewer tests were carried out for each dataset (4 instead of 150).  There were a number of datasets were CCFs performed noticeably better then any of the other classifiers, with the performance on the \textit{hill-valley-noisy} dataset particularly impressive, giving a error rate of $6.93\%$ with the next best classifier only achieving $25.7\%$.

\clearpage

{ 			\centering    
	\scriptsize             
	\setlength\tabcolsep{4pt}	  
	\renewcommand{\arraystretch}{1.2}   
	\begin{longtable}{|c|c|c|c|c|cc|c|c|c|c|c|}     
		\caption{Dataset summaries and mean and standard deviations of percentage of test cases misclassified for CCFs and \textsc{TreeBagger} on 82 datasets used for comparison to classifiers used in \cite{fernandez2014we}.  	Datasets were a predefined train / test split was used instead of a 4-fold cross validation have no standard deviation given.   $K$ = number of classes, $N$ = number of data points and $D$ = number of features (note all datasets only contain ordinal features).  Method with best accuracy is shown in bold. $\bullet$ and $\circ$ indicate that CCFs were significantly better and worse respectively at the 10\% level of a Wilcoxon signed rank test.  Also shown is the average and standard deviation, best and worst for the error rate across the 179 classifiers (not including CCF and \textsc{TreeBagger}), and $N_B$ and $N_W$ the number of classifiers that were better and worse than CCFs respectively.  \label{table:detailedCCFandTreeBagger}}           \\             
		\hline                                                       
		Dataset & $K$ & $N$ & $D$ & CCF & \textsc{TreeBagger} & & Average & Best & Worst & $N_B$ & $N_W$ \\
		\hline    
		\endfirsthead
		\hline                                                       
		Dataset & $K$ & $N$ & $D$ & CCF & \textsc{TreeBagger} & & Average & Best & Worst & $N_B$ & $N_W$ \\
		\hline   
		\endhead      
		\hline		
		\endfoot    
		abalone & 3 & 4177 & 8 & \textbf{34.15 $\pm$ 1.53} & 35.42 $\pm$ 0.99 & $\bullet$ & 39.91 $\pm$ 7.52 & 32.6 & 65.4 & 14 & 157 \\                          
		acute-inflammation & 2 & 120 & 6 & \textbf{0.00 $\pm$ 0.00} & \textbf{0.00 $\pm$ 0.00} &  & 5.14 $\pm$ 12.97 & 0.0 & 50.9 & 0 & 50 \\                     
		acute-nephritis & 2 & 120 & 6 & \textbf{0.00 $\pm$ 0.00} & \textbf{0.00 $\pm$ 0.00} &  & 4.04 $\pm$ 10.20 & 0.0 & 41.7 & 0 & 56 \\                        
		balance-scale & 3 & 625 & 4 & \textbf{8.81 $\pm$ 0.95} & 15.87 $\pm$ 1.23 & $\bullet$ & 20.68 $\pm$ 17.84 & 1.0 & 92.8 & 23 & 151 \\                      
		balloons & 2 & 16 & 4 & \textbf{12.50 $\pm$ 12.50} & 25.00 $\pm$ 17.68 &  & 36.79 $\pm$ 15.19 & 0.0 & 81.3 & 7 & 159 \\                                   
		blood & 2 & 748 & 4 & \textbf{24.20 $\pm$ 3.47} & 25.00 $\pm$ 1.75 &  & 24.17 $\pm$ 4.35 & 19.7 & 62.6 & 137 & 41 \\                                      
		breast-cancer-wisc & 2 & 699 & 9 & \textbf{2.71 $\pm$ 1.17} & 2.86 $\pm$ 0.99 &  & 6.54 $\pm$ 7.74 & 2.6 & 34.5 & 1 & 176 \\                              
		breast-cancer-wisc-diag & 2 & 569 & 30 & \textbf{2.46 $\pm$ 1.06} & 3.87 $\pm$ 1.27 &  & 7.24 $\pm$ 8.14 & 1.8 & 37.3 & 10 & 162 \\                       
		breast-cancer-wisc-prog & 2 & 198 & 33 & \textbf{15.31 $\pm$ 1.77} & 17.35 $\pm$ 2.28 &  & 24.53 $\pm$ 4.36 & 17.2 & 44.9 & 0 & 178 \\                    
		breast-tissue & 6 & 106 & 9 & \textbf{28.85 $\pm$ 11.05} & \textbf{28.85 $\pm$ 5.77} &  & 39.08 $\pm$ 15.35 & 20.2 & 83.1 & 33 & 144 \\                   
		car & 4 & 1728 & 6 & \textbf{0.87 $\pm$ 0.60} & 1.39 $\pm$ 0.33 &  & 14.01 $\pm$ 13.27 & 0.8 & 98.3 & 1 & 177 \\                                          
		cardiotocography-10& 10 & 2126 & 21 & 13.14 $\pm$ 1.54 & \textbf{12.05 $\pm$ 1.15} & $\circ$ & 31.68 $\pm$ 20.22 & 11.5 & 84.5 & 4 & 170 \\        
		cardiotocography-3 & 3 & 2126 & 21 & 6.12 $\pm$ 0.21 & \textbf{5.18 $\pm$ 0.54} & $\circ$ & 12.82 $\pm$ 9.35 & 4.4 & 93.7 & 10 & 165 \\             
		congressional-voting & 2 & 435 & 16 & \textbf{39.91 $\pm$ 2.29} & \textbf{39.91 $\pm$ 2.38} &  & 39.58 $\pm$ 3.97 & 36.8 & 84.4 & 131 & 44 \\             
		conn-bench-sonar & 2 & 208 & 60 & \textbf{13.94 $\pm$ 2.50} & 14.90 $\pm$ 3.43 &  & 25.42 $\pm$ 8.74 & 9.6 & 53.4 & 12 & 166 \\               
		conn-bench-vowel & 11 & 462 & 11 & \textbf{0.00} & 1.30 &  & 30.02 $\pm$ 29.52 & 0.0 & 98.1 & 0 & 164 \\                                        
		contrac & 3 & 1473 & 9 & 49.18 $\pm$ 2.60 & \textbf{48.37 $\pm$ 2.31} &  & 50.41 $\pm$ 5.55 & 42.8 & 68.5 & 96 & 78 \\                                    
		dermatology & 6 & 366 & 34 & 2.75 $\pm$ 0.55 & \textbf{2.47 $\pm$ 0.91} &  & 13.81 $\pm$ 20.13 & 1.4 & 69.6 & 30 & 140 \\                                 
		echocardiogram & 2 & 131 & 10 & 17.42 $\pm$ 2.51 & \textbf{15.15 $\pm$ 2.14} &  & 18.49 $\pm$ 5.74 & 12.3 & 46.2 & 106 & 68 \\                            
		ecoli & 8 & 336 & 7 & 14.58 $\pm$ 2.96 & \textbf{12.50 $\pm$ 2.15} &  & 22.39 $\pm$ 12.92 & 9.1 & 79.5 & 50 & 123 \\                                      
		energy-y1 & 3 & 768 & 8 & \textbf{2.86 $\pm$ 1.07} & 3.26 $\pm$ 1.30 &  & 14.05 $\pm$ 14.03 & 2.2 & 94.5 & 6 & 170 \\                                     
		energy-y2 & 3 & 768 & 8 & \textbf{9.24 $\pm$ 1.30} & 9.64 $\pm$ 0.94 &  & 15.74 $\pm$ 12.43 & 6.6 & 95.4 & 39 & 136 \\                                    
		fertility & 2 & 100 & 9 & \textbf{11.00 $\pm$ 1.73} & \textbf{11.00 $\pm$ 1.73} &  & 14.22 $\pm$ 5.65 & 10.0 & 54.0 & 1 & 174 \\                          
		glass & 6 & 214 & 9 & 25.94 $\pm$ 1.56 & \textbf{24.06 $\pm$ 5.72} &  & 39.01 $\pm$ 11.25 & 21.5 & 68.1 & 14 & 163 \\                                     
		haberman-survival & 2 & 306 & 3 & \textbf{29.61 $\pm$ 2.18} & 33.55 $\pm$ 1.47 &  & 27.43 $\pm$ 3.63 & 22.9 & 51.4 & 151 & 26 \\                          
		heart-cleveland & 5 & 303 & 13 & \textbf{41.45 $\pm$ 3.89} & 42.76 $\pm$ 2.87 &  & 44.04 $\pm$ 4.49 & 35.2 & 80.3 & 38 & 134 \\                           
		heart-hungarian & 2 & 294 & 12 & \textbf{16.10 $\pm$ 3.54} & 18.84 $\pm$ 3.27 & $\bullet$ & 20.04 $\pm$ 5.15 & 13.4 & 36.1 & 25 & 153 \\                  
		heart-switzerland & 5 & 123 & 12 & \textbf{56.45 $\pm$ 3.61} & 59.68 $\pm$ 2.79 &  & 62.00 $\pm$ 5.77 & 46.8 & 87.1 & 24 & 149 \\                         
		heart-va & 5 & 200 & 12 & \textbf{67.00 $\pm$ 3.00} & 67.50 $\pm$ 4.56 &  & 68.19 $\pm$ 3.83 & 60.0 & 77.5 & 68 & 103 \\                                  
		hepatitis & 2 & 155 & 19 & \textbf{17.95 $\pm$ 3.14} & 18.59 $\pm$ 2.13 &  & 19.97 $\pm$ 5.96 & 10.3 & 69.2 & 54 & 123 \\                                 
		hill-valley-noisy & 2 & 606 & 100 & \textbf{6.93} & 50.83 &  & 45.91 $\pm$ 7.53 & 25.7 & 93.9 & 0 & 176 \\                                                      
		ilpd-indian-liver & 2 & 583 & 9 & 29.79 $\pm$ 1.03 & \textbf{29.28 $\pm$ 2.44} &  & 30.08 $\pm$ 5.18 & 22.4 & 70.5 & 115 & 60 \\                          
		ionosphere & 2 & 351 & 33 & \textbf{5.11 $\pm$ 1.70} & 7.10 $\pm$ 1.68 &  & 14.05 $\pm$ 9.58 & 4.5 & 64.2 & 3 & 173 \\                                    
		iris & 3 & 150 & 4 & \textbf{2.70 $\pm$ 1.91} & 4.73 $\pm$ 2.24 &  & 10.61 $\pm$ 17.57 & 0.7 & 99.3 & 29 & 147 \\                                         
		led-display & 10 & 1000 & 7 & 27.30 $\pm$ 1.77 & \textbf{27.20 $\pm$ 1.47} &  & 39.76 $\pm$ 22.15 & 25.2 & 91.2 & 37 & 135 \\                             
		lenses & 3 & 24 & 4 & \textbf{16.67 $\pm$ 16.67} & \textbf{16.67 $\pm$ 16.67} &  & 26.02 $\pm$ 13.60 & 4.2 & 83.3 & 30 & 119 \\                           
		letter & 26 & 20000 & 16 & \textbf{2.27 $\pm$ 0.07} & 3.55 $\pm$ 0.14 & $\bullet$ & 37.28 $\pm$ 32.22 & 2.6 & 96.1 & 0 & 163 \\                           
		libras & 15 & 360 & 90 & \textbf{11.39 $\pm$ 3.46} & 20.00 $\pm$ 2.22 & $\bullet$ & 42.91 $\pm$ 25.55 & 10.8 & 93.4 & 1 & 176 \\                          
		low-res-spect & 9 & 531 & 100 & \textbf{7.52 $\pm$ 0.53} & 8.08 $\pm$ 0.98 &  & 19.99 $\pm$ 12.21 & 6.6 & 62.0 & 1 & 176 \\                               
		magic & 2 & 19020 & 10 & \textbf{12.31 $\pm$ 0.19} & 12.76 $\pm$ 0.26 & $\bullet$ & 19.98 $\pm$ 6.75 & 11.7 & 41.4 & 7 & 159 \\                           
		miniboone & 2 & 130064 & 50 & \textbf{6.13 $\pm$ 0.07} & 6.32 $\pm$ 0.08 & $\bullet$ & 17.41 $\pm$ 15.01 & 6.2 & 71.6 & 0 & 132 \\                        
		musk-1 & 2 & 476 & 166 & \textbf{11.55 $\pm$ 1.82} & 11.97 $\pm$ 4.88 &  & 20.59 $\pm$ 9.65 & 6.3 & 56.6 & 29 & 148 \\                                    
		musk-2 & 2 & 6598 & 166 & 2.96 $\pm$ 0.20 & \textbf{2.29 $\pm$ 0.16} & $\circ$ & 7.50 $\pm$ 7.94 & 0.2 & 73.5 & 38 & 134 \\                               
		nursery & 5 & 12960 & 8 & \textbf{0.15 $\pm$ 0.02} & 0.34 $\pm$ 0.12 & $\bullet$ & 16.15 $\pm$ 21.35 & 0.0 & 76.7 & 3 & 157 \\                            
		oocytes\_m\_nucleus\_4d & 0 & 1022 & 41 & \textbf{15.59 $\pm$ 2.36} & 22.45 $\pm$ 3.67 & $\bullet$ & 27.33 $\pm$ 10.82 & 14.0 & 91.7 & 2 & 175 \\
		oocytes\_m\_states\_2f & 0 & 1022 & 25 & \textbf{6.76 $\pm$ 0.98} & 7.75 $\pm$ 0.85 & $\bullet$ & 14.18 $\pm$ 12.19 & 6.0 & 99.7 & 1 & 176 \\    
		oocytes\_t\_nucleus\_2f & 0 & 912 & 25 & \textbf{14.91 $\pm$ 1.24} & 19.41 $\pm$ 1.62 & $\bullet$ & 28.79 $\pm$ 12.16 & 13.2 & 90.2 & 5 & 171 \\
		oocytes\_t\_states\_5b & 0 & 912 & 32 & \textbf{6.36 $\pm$ 1.30} & 7.46 $\pm$ 0.44 &  & 15.58 $\pm$ 12.61 & 4.9 & 100.0 & 12 & 163 \\           
		optical & 10 & 1797 & 62 & 2.62 & \textbf{2.56} &  & 26.67 $\pm$ 29.52 & 1.3 & 90.1 & 5 & 163 \\                                                          
		ozone & 2 & 2536 & 72 & 2.96 $\pm$ 0.13 & \textbf{2.92 $\pm$ 0.08} &  & 5.02 $\pm$ 6.92 & 2.6 & 48.9 & 84 & 72 \\                                         
		page-blocks & 5 & 5473 & 10 & 2.87 $\pm$ 0.35 & \textbf{2.76 $\pm$ 0.39} &  & 5.93 $\pm$ 5.55 & 2.5 & 59.7 & 20 & 148 \\                                  
		parkinsons & 2 & 195 & 22 & \textbf{6.63 $\pm$ 3.02} & 10.71 $\pm$ 2.65 & $\bullet$ & 14.94 $\pm$ 7.27 & 5.6 & 62.2 & 8 & 169 \\                          
		pendigits & 10 & 3498 & 16 & \textbf{3.20} & 4.35 &  & 25.31 $\pm$ 28.39 & 2.2 & 89.7 & 16 & 157 \\                                                       
		pima & 2 & 768 & 8 & \textbf{24.48 $\pm$ 1.61} & 25.39 $\pm$ 1.00 &  & 25.73 $\pm$ 4.05 & 21.0 & 41.8 & 85 & 87 \\                                        
		planning & 2 & 182 & 12 & 32.22 $\pm$ 3.69 & \textbf{31.67 $\pm$ 1.84} &  & 32.32 $\pm$ 5.57 & 27.2 & 58.8 & 123 & 55 \\                                  
		plant-margin & 100 & 1600 & 64 & \textbf{11.06 $\pm$ 0.48} & 14.44 $\pm$ 1.30 & $\bullet$ & 50.61 $\pm$ 31.22 & 12.8 & 99.0 & 0 & 163 \\                  
		plant-shape & 100 & 1600 & 64 & \textbf{24.31 $\pm$ 0.89} & 34.94 $\pm$ 1.79 & $\bullet$ & 62.16 $\pm$ 23.43 & 27.7 & 99.4 & 0 & 161 \\                   
		plant-texture & 100 & 1599 & 64 & \textbf{14.19 $\pm$ 0.84} & 16.63 $\pm$ 1.35 & $\bullet$ & 49.80 $\pm$ 30.84 & 13.4 & 99.1 & 2 & 156 \\                 
		ringnorm & 2 & 7400 & 20 & \textbf{2.24 $\pm$ 0.36} & 4.38 $\pm$ 0.40 & $\bullet$ & 16.09 $\pm$ 14.33 & 1.3 & 50.5 & 25 & 149 \\                          
		seeds & 3 & 210 & 7 & 5.29 $\pm$ 2.50 & \textbf{4.81 $\pm$ 2.15} &  & 14.02 $\pm$ 16.34 & 2.8 & 87.0 & 21 & 146 \\                                        
		semeion & 10 & 1593 & 256 & 4.96 $\pm$ 1.44 & \textbf{4.65 $\pm$ 1.30} &  & 30.74 $\pm$ 27.22 & 3.6 & 89.9 & 9 & 164 \\                                   
		spambase & 2 & 4601 & 57 & \textbf{4.02 $\pm$ 0.34} & 4.76 $\pm$ 0.22 & $\bullet$ & 12.89 $\pm$ 9.73 & 3.9 & 43.6 & 1 & 170 \\                            
		spect & 2 & 186 & 22 & 31.72 & \textbf{25.81} &  & 42.88 $\pm$ 6.07 & 27.8 & 92.5 & 6 & 170 \\                                                            
		spectf & 2 & 187 & 44 & \textbf{8.02} & \textbf{8.02} &  & 17.44 $\pm$ 21.91 & 7.5 & 94.1 & 4 & 128 \\                                                    
		statlog-image & 7 & 2310 & 18 & \textbf{1.39 $\pm$ 0.49} & 2.04 $\pm$ 0.72 & $\bullet$ & 18.51 $\pm$ 25.22 & 1.4 & 85.9 & 0 & 176 \\                      
		statlog-landsat & 6 & 2000 & 36 & \textbf{9.10} & \textbf{9.10} &  & 24.28 $\pm$ 19.65 & 8.1 & 78.9 & 6 & 166 \\                                          
		statlog-shuttle & 7 & 14500 & 9 & 0.03 & \textbf{0.01} &  & 6.61 $\pm$ 8.77 & 0.0 & 57.6 & 0 & 159 \\                                                     
		statlog-vehicle & 4 & 846 & 18 & \textbf{17.65 $\pm$ 2.71} & 24.64 $\pm$ 1.74 & $\bullet$ & 34.61 $\pm$ 16.17 & 14.9 & 74.4 & 10 & 168 \\                 
		steel-plates & 7 & 1941 & 27 & \textbf{21.24 $\pm$ 0.87} & 21.34 $\pm$ 0.83 &  & 36.40 $\pm$ 16.11 & 19.6 & 92.0 & 4 & 172 \\                             
		synthetic-control & 6 & 600 & 60 & \textbf{0.83 $\pm$ 0.73} & 1.50 $\pm$ 1.09 &  & 18.19 $\pm$ 26.13 & 0.3 & 87.3 & 5 & 171 \\                            
		trains & 2 & 10 & 29 & \textbf{12.50 $\pm$ 21.65} & \textbf{12.50 $\pm$ 21.65} &  & 31.89 $\pm$ 17.36 & 0.0 & 87.5 & 7 & 131 \\                           
		twonorm & 2 & 7400 & 20 & \textbf{2.11 $\pm$ 0.40} & 2.72 $\pm$ 0.22 & $\bullet$ & 10.44 $\pm$ 13.71 & 2.0 & 50.1 & 2 & 145 \\                            
		vertebral-column-2 & 2 & 310 & 6 & \textbf{16.23 $\pm$ 1.95} & \textbf{16.23 $\pm$ 3.84} &  & 19.85 $\pm$ 6.46 & 12.6 & 67.8 & 57 & 120 \\          
		vertebral-column-3 & 3 & 310 & 6 & \textbf{13.96 $\pm$ 3.36} & 15.91 $\pm$ 2.96 &  & 23.00 $\pm$ 11.63 & 12.6 & 67.8 & 7 & 170 \\                   
		wall-following & 4 & 5456 & 24 & 2.93 $\pm$ 0.45 & \textbf{0.46 $\pm$ 0.20} & $\circ$ & 21.40 $\pm$ 20.01 & 0.1 & 76.7 & 47 & 126 \\                      
		waveform & 3 & 5000 & 21 & \textbf{13.70 $\pm$ 0.93} & 15.28 $\pm$ 0.99 & $\bullet$ & 23.99 $\pm$ 14.81 & 12.9 & 74.9 & 29 & 145 \\                       
		waveform-noise & 3 & 5000 & 40 & \textbf{12.78 $\pm$ 0.51} & 13.88 $\pm$ 0.72 & $\bullet$ & 24.38 $\pm$ 14.75 & 12.6 & 76.3 & 1 & 174 \\                  
		wine & 3 & 178 & 13 & \textbf{0.00 $\pm$ 0.00} & 1.14 $\pm$ 1.14 &  & 9.95 $\pm$ 16.15 & 0.0 & 98.3 & 0 & 176 \\                                          
		wine-quality-red & 6 & 1599 & 11 & \textbf{29.94 $\pm$ 1.08} & 30.88 $\pm$ 1.66 &  & 44.48 $\pm$ 12.26 & 31.0 & 98.1 & 0 & 177 \\                         
		wine-quality-white & 7 & 4898 & 11 & \textbf{30.29 $\pm$ 0.12} & 31.58 $\pm$ 0.78 &  & 48.44 $\pm$ 12.39 & 30.9 & 98.2 & 0 & 173 \\                       
		yeast & 10 & 1484 & 8 & \textbf{37.60 $\pm$ 3.05} & 37.80 $\pm$ 2.48 &  & 47.56 $\pm$ 10.20 & 36.3 & 70.3 & 3 & 173 \\                                    
		zoo & 7 & 101 & 16 & \textbf{1.00 $\pm$ 1.73} & \textbf{1.00 $\pm$ 1.73} &  & 13.49 $\pm$ 16.79 & 1.0 & 99.0 & 0 & 175 \\    
	\end{longtable}                                                      
}

{ 			\centering    
	\scriptsize             
	\setlength\tabcolsep{4pt}	  
	\renewcommand{\arraystretch}{1.2}   
	\begin{longtable}{|c|c|c|c|c|c|c|c|c|}      
		\caption{Comparison of all classifiers on 82 UCI datasets. R is the mean rank according to error rate; E is the mean error rate (\%); $\kappa$ is the mean Cohen's $\kappa$ \cite{carletta1996assessing}; $\text{E}_{\text{CCF}}$ and $\kappa_{\text{CCF}}$ are the respective values for CCFs on the datasets where the competing classifier successfully ran (note CCFs successfully ran on all datasets),  $N_v$ and $N_l$ are the number of datasets where the CCFs $\kappa$ was higher and lower than the classifier respectively, and p is the p-value for whether the CCFs $\kappa$ mean is higher using a Wilcoxon signed ranks test.   \label{table:fullPerformanceRanking}}           \\             
		\hline                                                       
		Classifier & R & E & $\text{E}_{\text{CCF}}$ & $\kappa$ & $\kappa_{\text{CCF}}$ & $N_v$ & $N_l$ & p \\
		\hline        
		\endfirsthead
		\hline                                                       
		Classifier & R & E & $\text{E}_{\text{CCF}}$ & $\kappa$ & $\kappa_{\text{CCF}}$ & $N_v$ & $N_l$ & p \\
		\hline       
		\endhead      
		\hline		
		\endfoot    
		CCF & \textbf{28.87} & \textbf{14.08} & - & \textbf{70.67} & - & - & - & - \\                                             
		svmPoly\_t & 31.53 & 15.73 & 14.27 & 65.10 & 69.61 & 54 & 25 & 0.00023 \\                              
		svmRadialCost\_t & 31.84 & 15.33 & 14.27 & 66.55 & 69.61 & 43 & 36 & 0.11 \\                           
		svm\_C & 32 & 15.67 & 14.18 & 67.65 & 70.49 & 46 & 32 & 0.18 \\                                        
		elm\_kernel\_m & 32.19 & 15.20 & 14.54 & 69.01 & 69.75 & 42 & 36 & 0.16 \\                             
		parRF\_m & 33.03 & 15.54 & 14.08 & 67.73 & 70.67 & 52 & 27 & 0.01 \\                                   
		svmRadial\_t & 33.77 & 15.68 & 14.27 & 65.88 & 69.61 & 50 & 28 & 0.0016 \\                             
		rf\_caret & 34.48 & 15.56 & 14.18 & 67.67 & 70.49 & 54 & 23 & 0.00063 \\                               
		rforest\_R & 40.70 & 15.82 & 14.18 & 66.67 & 70.49 & 57 & 20 & 2e-05 \\                                
		TreeBagger & 40.91 & 15.75 & 14.08 & 67.51 & 70.67 & 55 & 23 & 3.5e-05 \\                              
		Bag\_LibSVM\_w & 42.28 & 16.65 & 14.25 & 58.13 & 70.27 & 70 & 12 & 3.2e-12 \\                          
		C50\_t & 42.61 & 16.85 & 14.08 & 66.11 & 70.67 & 57 & 22 & 0.0004 \\                                   
		nnet\_t & 42.87 & 18.74 & 14.08 & 64.72 & 70.67 & 54 & 26 & 0.0015 \\                                  
		avNNet\_t & 43.26 & 18.77 & 14.08 & 64.88 & 70.67 & 50 & 29 & 0.001 \\                                 
		RotationForest\_w & 44.62 & 16.64 & 14.08 & 65.34 & 70.67 & 64 & 15 & 7.1e-09 \\                       
		pcaNNet\_t & 45.86 & 19.28 & 14.08 & 63.83 & 70.67 & 54 & 25 & 0.00015 \\                              
		mlp\_t & 46.06 & 17.38 & 14.08 & 66.75 & 70.67 & 54 & 26 & 0.0016 \\                                   
		LibSVM\_w & 46.50 & 16.65 & 14.08 & 63.80 & 70.67 & 57 & 21 & 2.9e-06 \\                               
		MultiBoostAB\_LibSVM\_w & 46.90 & 16.82 & 14.25 & 64.47 & 70.27 & 61 & 17 & 1.8e-06 \\                 
		RRF\_t & 49.56 & 16.71 & 14.18 & 66.30 & 70.49 & 59 & 20 & 1.1e-05 \\                                  
		adaboost\_R & 50.48 & 18.33 & 14.01 & 64.24 & 71.07 & 58 & 21 & 1.8e-06 \\                             
		RRFglobal\_caret & 51.88 & 16.83 & 14.18 & 65.97 & 70.49 & 58 & 20 & 2.7e-07 \\                        
		RandomForest\_weka & 52.66 & 16.75 & 14.24 & 63.42 & 69.64 & 62 & 15 & 1.6e-08 \\                      
		svmLinear\_caret & 53.14 & 17.96 & 14.27 & 61.51 & 69.61 & 60 & 19 & 1.3e-08 \\                        
		MultiBoostAB\_RandomForest\_weka & 53.38 & 16.51 & 13.93 & 62.91 & 69.78 & 66 & 13 & 2.8e-09 \\        
		gaussprRadial\_R & 53.75 & 18.41 & 14.69 & 61.92 & 70.70 & 62 & 17 & 1.9e-09 \\                        
		MultiBoostAB\_MultilayerPerceptron\_weka & 56.49 & 17.20 & 14.08 & 66.29 & 70.67 & 61 & 18 & 2.3e-06 \\
		pnn\_matlab & 56.56 & 18.03 & 14.36 & 62.98 & 70.13 & 66 & 13 & 2.2e-08 \\                             
		mda\_caret & 57.04 & 20.96 & 14.08 & 62.25 & 70.67 & 59 & 20 & 4.3e-07 \\                              
		cforest\_caret & 58.27 & 19.52 & 14.61 & 59.80 & 69.50 & 70 & 9 & 1.9e-11 \\                           
		svmlight\_C & 58.35 & 18.46 & 15.80 & 61.29 & 65.70 & 57 & 21 & 5.2e-07 \\                             
		mlp\_C & 58.50 & 18.47 & 14.08 & 64.35 & 70.67 & 67 & 13 & 9.4e-09 \\                                  
		Decorate\_weka & 58.69 & 17.87 & 14.12 & 64.58 & 70.56 & 65 & 14 & 7.1e-07 \\                          
		Bagging\_RandomForest\_weka & 59.75 & 17.39 & 14.09 & 58.82 & 67.12 & 66 & 12 & 5.8e-11 \\             
		rbfDDA\_caret & 59.91 & 19.23 & 15.80 & 60.96 & 66.34 & 67 & 11 & 8.7e-11 \\                           
		MultiBoostAB\_PART\_weka & 60.27 & 18.08 & 14.08 & 65.01 & 70.67 & 61 & 18 & 2e-06 \\                  
		dkp\_C & 60.57 & 17.93 & 14.08 & 45.14 & 70.67 & 66 & 12 & 1e-10 \\                                    
		knn\_caret & 60.94 & 18.93 & 14.08 & 61.20 & 70.67 & 65 & 14 & 7.7e-09 \\                              
		MultilayerPerceptron\_weka & 61.97 & 18.00 & 14.08 & 64.79 & 70.67 & 64 & 15 & 1.2e-07 \\              
		glmnet\_R & 62 & 19.35 & 14.08 & 61.25 & 70.67 & 60 & 17 & 7.1e-09 \\                                  
		multinom\_caret & 62.04 & 19.77 & 14.08 & 61.20 & 70.67 & 65 & 15 & 2.1e-09 \\                         
		Bagging\_PART\_weka & 62.45 & 18.43 & 14.08 & 63.97 & 70.67 & 63 & 16 & 7.1e-08 \\                     
		rda\_R & 62.63 & 18.40 & 14.08 & 62.68 & 70.67 & 57 & 20 & 1.9e-07 \\                                  
		knn\_R & 62.64 & 18.21 & 14.08 & 62.91 & 70.67 & 63 & 16 & 1.5e-08 \\                                  
		fda\_caret & 63.32 & 19.05 & 14.08 & 62.65 & 70.67 & 67 & 13 & 2.7e-09 \\                              
		elm\_matlab & 63.79 & 18.86 & 14.08 & 61.64 & 70.67 & 67 & 14 & 2e-09 \\                               
		SMO\_weka & 63.94 & 18.41 & 14.08 & 62.63 & 70.67 & 64 & 16 & 2.5e-08 \\                               
		RandomCommittee\_weka & 64.11 & 18.19 & 14.08 & 63.62 & 70.67 & 63 & 16 & 4.5e-09 \\                   
		MultiBoostAB\_J48\_weka & 65.09 & 18.70 & 14.18 & 63.78 & 70.49 & 65 & 14 & 2e-08 \\                   
		mlpWeightDecay\_caret & 65.72 & 22.12 & 14.18 & 59.53 & 70.49 & 64 & 16 & 4.1e-09 \\                   
		Bagging\_RandomTree\_weka & 65.85 & 18.50 & 14.08 & 63.84 & 70.67 & 64 & 15 & 3.1e-09 \\               
		pda\_caret & 66.80 & 19.59 & 14.08 & 61.04 & 70.67 & 62 & 17 & 1e-09 \\                                
		mlm\_R & 66.82 & 19.94 & 14.08 & 60.78 & 70.67 & 65 & 14 & 1.1e-08 \\                                  
		ClassificationViaRegression\_weka & 67.05 & 19.50 & 14.08 & 61.09 & 70.67 & 66 & 14 & 8.8e-10 \\       
		Bagging\_J48\_weka & 67.42 & 19.02 & 14.08 & 62.90 & 70.67 & 66 & 13 & 2.1e-09 \\                      
		AdaBoostM1\_J48\_weka & 69.14 & 18.42 & 14.18 & 64.49 & 70.49 & 63 & 16 & 8.7e-08 \\                   
		treebag\_caret & 69.31 & 18.82 & 14.18 & 62.82 & 70.49 & 65 & 15 & 3.2e-09 \\                          
		rbf\_caret & 69.49 & 22.06 & 14.08 & 58.67 & 70.67 & 65 & 15 & 6e-09 \\                                
		SimpleLogistic\_weka & 69.76 & 20.26 & 14.08 & 58.47 & 70.67 & 66 & 14 & 7.2e-11 \\                    
		fda\_R & 70.26 & 20.09 & 14.08 & 60.35 & 70.67 & 63 & 16 & 4.4e-09 \\                                  
		ldaBag\_R & 70.31 & 20.21 & 14.08 & 60.00 & 70.67 & 62 & 16 & 2e-09 \\                                 
		gcvEarth\_caret & 70.46 & 20.42 & 14.08 & 60.38 & 70.67 & 66 & 14 & 1.1e-10 \\                         
		lda\_R & 70.64 & 20.21 & 14.08 & 60.22 & 70.67 & 63 & 16 & 3.2e-09 \\                                  
		lssvmRadial\_caret & 72.38 & 19.13 & 14.51 & 62.62 & 69.66 & 69 & 10 & 4e-09 \\                        
		LibLINEAR\_weka & 72.68 & 20.36 & 14.08 & 60.29 & 70.67 & 69 & 11 & 1.7e-11 \\                         
		lda2\_caret & 72.76 & 20.19 & 14.08 & 59.75 & 70.67 & 67 & 13 & 2.1e-10 \\                             
		Bagging\_Logistic\_weka & 73.91 & 19.70 & 13.96 & 60.82 & 70.61 & 67 & 13 & 1.2e-10 \\                 
		MultiBoostAB\_RandomTree\_weka & 74.09 & 19.04 & 14.08 & 62.81 & 70.67 & 69 & 11 & 2.8e-10 \\          
		Logistic\_weka & 75.35 & 20.29 & 13.96 & 59.52 & 70.61 & 69 & 11 & 1.1e-11 \\                          
		MultiBoostAB\_REPTree\_weka & 75.37 & 19.99 & 14.08 & 61.17 & 70.67 & 72 & 9 & 1.7e-11 \\              
		END\_weka & 75.48 & 19.34 & 14.08 & 61.30 & 70.67 & 66 & 14 & 3.8e-10 \\                               
		Bagging\_IBk\_weka & 75.65 & 19.74 & 14.08 & 60.75 & 70.67 & 74 & 7 & 2.2e-12 \\                       
		Bagging\_LWL\_weka & 75.65 & 19.74 & 14.08 & 60.75 & 70.67 & 74 & 7 & 2.2e-12 \\                       
		Bagging\_weka & 75.65 & 19.74 & 14.08 & 60.23 & 70.67 & 74 & 7 & 1.7e-12 \\                            
		mda\_R & 76.31 & 20.04 & 14.08 & 59.93 & 70.67 & 68 & 13 & 1.5e-10 \\                                  
		sda\_caret & 76.40 & 20.34 & 14.08 & 59.40 & 70.67 & 65 & 13 & 5e-11 \\                                
		MultiBoostAB\_Logistic\_weka & 76.62 & 20.48 & 14.08 & 60.71 & 70.67 & 65 & 15 & 2e-10 \\              
		svmBag\_R & 76.80 & 25.06 & 13.98 & 55.03 & 70.52 & 65 & 13 & 2.4e-09 \\                               
		RandomSubSpace\_weka & 78.20 & 20.71 & 14.08 & 56.89 & 70.67 & 74 & 5 & 7.3e-14 \\                     
		hdda\_R & 79.29 & 20.48 & 14.08 & 59.99 & 70.67 & 62 & 17 & 6e-09 \\                                   
		lvq\_caret & 79.69 & 19.85 & 14.09 & 58.23 & 70.00 & 70 & 11 & 3.2e-11 \\                              
		pls\_caret & 79.98 & 24.08 & 14.89 & 54.92 & 68.93 & 68 & 12 & 9.4e-12 \\                              
		ctreeBag\_R & 80.06 & 21.01 & 14.27 & 56.26 & 69.82 & 72 & 8 & 7.7e-13 \\                              
		MultiClassClassifier\_weka & 81.43 & 21.75 & 14.08 & 58.42 & 70.67 & 68 & 12 & 2.1e-11 \\              
		LogitBoost\_weka & 83.48 & 20.92 & 14.08 & 58.88 & 70.67 & 69 & 10 & 3.3e-11 \\                        
		C50Rules\_caret & 83.87 & 20.95 & 14.08 & 59.97 & 70.67 & 71 & 8 & 2.5e-11 \\                          
		JRip\_caret & 83.95 & 22.05 & 14.08 & 56.94 & 70.67 & 69 & 12 & 6.3e-12 \\                             
		PART\_caret & 84.03 & 20.95 & 14.08 & 57.65 & 70.67 & 71 & 9 & 1.2e-11 \\                              
		RBFNetwork\_weka & 85.81 & 20.79 & 14.14 & 54.81 & 69.84 & 71 & 7 & 2.5e-12 \\                         
		J48\_caret & 86.10 & 20.88 & 14.08 & 56.85 & 70.48 & 68 & 11 & 1e-11 \\                                
		C50Tree\_caret & 87.52 & 21.49 & 14.08 & 59.16 & 70.67 & 69 & 10 & 5.5e-12 \\                          
		IBk\_weka & 87.93 & 20.30 & 14.08 & 61.11 & 70.67 & 68 & 10 & 1.2e-10 \\                               
		qda\_caret & 89.09 & 21.96 & 14.18 & 55.45 & 70.47 & 71 & 11 & 8.4e-13 \\                              
		PART\_weka & 89.14 & 21.53 & 14.08 & 60.35 & 70.67 & 67 & 13 & 6.4e-10 \\                              
		IB1\_weka & 89.25 & 20.53 & 14.08 & 60.17 & 70.67 & 71 & 7 & 1.4e-11 \\                                
		KStar\_weka & 89.32 & 20.74 & 14.18 & 58.60 & 70.49 & 75 & 4 & 4.6e-13 \\                              
		NBTree\_weka & 89.73 & 21.41 & 14.08 & 58.91 & 70.67 & 69 & 10 & 2.7e-12 \\                            
		J48\_weka & 89.94 & 21.75 & 14.18 & 59.29 & 70.49 & 70 & 10 & 8.1e-11 \\                               
		Bagging\_DecisionTable\_weka & 91.38 & 22.73 & 14.18 & 55.73 & 70.49 & 75 & 4 & 7.3e-14 \\             
		MultiBoostAB\_DecisionTable\_weka & 91.52 & 23.78 & 14.18 & 55.37 & 70.49 & 72 & 8 & 9.5e-13 \\        
		obliqueTree\_R & 92.02 & 23.77 & 14.14 & 55.43 & 69.84 & 71 & 8 & 6.9e-12 \\                           
		AttributeSelectedClassifier\_weka & 92.12 & 22.36 & 14.08 & 56.63 & 70.67 & 71 & 8 & 2e-12 \\          
		NNge\_weka & 93.20 & 21.31 & 14.18 & 58.21 & 70.49 & 73 & 6 & 1.7e-13 \\                               
		bagging\_R & 94.07 & 31.45 & 14.18 & 49.22 & 70.49 & 70 & 11 & 1.7e-12 \\                              
		rbf\_matlab & 94.39 & 26.03 & 16.39 & 51.92 & 65.73 & 66 & 12 & 1.5e-11 \\                             
		DTNB\_weka & 96.04 & 22.10 & 14.08 & 56.56 & 70.67 & 72 & 7 & 3.2e-12 \\                               
		ctree2\_caret & 96.23 & 25.83 & 14.18 & 52.85 & 70.49 & 72 & 10 & 7.9e-12 \\                           
		lvq\_R & 96.51 & 27.97 & 14.08 & 52.58 & 70.67 & 72 & 7 & 2.5e-12 \\                                   
		REPTree\_weka & 96.69 & 23.05 & 14.08 & 55.64 & 70.67 & 75 & 6 & 2.3e-13 \\                            
		rrlda\_R & 97.07 & 24.39 & 14.08 & 56.76 & 70.67 & 69 & 10 & 1e-10 \\                                  
		cascor\_C & 97.54 & 23.07 & 14.08 & 57.12 & 70.67 & 71 & 9 & 7e-13 \\                                  
		ctree\_caret & 98.09 & 24.20 & 14.33 & 53.72 & 69.96 & 73 & 8 & 5.6e-12 \\                             
		JRip\_weka & 98.14 & 23.08 & 14.18 & 56.73 & 70.49 & 73 & 8 & 3.4e-12 \\                               
		mlp\_matlab & 98.26 & 27.00 & 14.08 & 48.38 & 70.67 & 74 & 6 & 1e-13 \\                                
		OrdinalClassClassifier\_weka & 98.77 & 24.40 & 14.08 & 56.17 & 70.67 & 71 & 9 & 6.1e-12 \\             
		nbBag\_R & 98.81 & 22.94 & 14.17 & 56.08 & 70.48 & 70 & 11 & 7.7e-12 \\                                
		Ridor\_weka & 99.05 & 21.65 & 14.14 & 56.21 & 69.84 & 72 & 8 & 5.3e-13 \\                              
		bdk\_R & 99.14 & 21.99 & 14.08 & 58.06 & 70.67 & 74 & 5 & 2.7e-13 \\                                   
		Dagging\_weka & 99.29 & 24.46 & 14.08 & 50.08 & 70.67 & 76 & 4 & 3.5e-14 \\                            
		rpart\_caret & 99.90 & 25.03 & 14.08 & 53.93 & 70.67 & 74 & 8 & 3.4e-12 \\                             
		BayesNet\_weka & 100.64 & 23.16 & 14.08 & 55.20 & 70.67 & 69 & 10 & 8.1e-13 \\                         
		naiveBayes\_R & 101.61 & 23.50 & 14.08 & 56.55 & 70.67 & 70 & 9 & 4.7e-12 \\                           
		FilteredClassifier\_weka & 101.65 & 23.82 & 14.08 & 54.16 & 70.67 & 75 & 4 & 6.7e-14 \\                
		plsBag\_R & 101.70 & 31.34 & 14.18 & 44.50 & 70.49 & 72 & 6 & 3.8e-13 \\                               
		rpart2\_caret & 102.34 & 24.06 & 14.08 & 55.35 & 70.48 & 72 & 10 & 1.8e-11 \\                          
		logitboost\_R & 102.60 & 24.16 & 14.08 & 64.70 & 70.67 & 67 & 12 & 1.5e-08 \\                          
		rpart\_R & 103.37 & 25.84 & 14.08 & 52.04 & 70.67 & 73 & 8 & 7.3e-13 \\                                
		slda\_caret & 105.58 & 26.13 & 14.08 & 48.75 & 70.67 & 76 & 3 & 4e-14 \\                               
		nnetBag\_R & 105.84 & 39.24 & 14.08 & 35.37 & 70.67 & 70 & 10 & 2.5e-12 \\                             
		pam\_caret & 107.69 & 25.52 & 14.08 & 46.49 & 70.67 & 78 & 2 & 1.2e-14 \\                              
		mars\_R & 108.10 & 35.05 & 14.18 & 49.44 & 70.49 & 73 & 7 & 1.9e-13 \\                                 
		MultiBoostAB\_NaiveBayes\_weka & 109.33 & 25.66 & 14.08 & 52.92 & 70.67 & 70 & 11 & 1.4e-12 \\         
		RandomTree\_weka & 111.01 & 23.91 & 14.08 & 55.65 & 70.67 & 76 & 6 & 3.4e-14 \\                        
		sddaLDA\_R & 111.26 & 26.96 & 14.08 & 44.47 & 70.67 & 76 & 4 & 3.8e-14 \\                              
		simpls\_R & 112.09 & 37.47 & 14.08 & 42.78 & 70.67 & 71 & 9 & 3.6e-13 \\                               
		widekernelpls\_R & 112.30 & 36.93 & 14.73 & 41.31 & 69.36 & 71 & 9 & 2.4e-13 \\                        
		Bagging\_NaiveBayes\_weka & 113.50 & 26.54 & 14.08 & 51.39 & 70.67 & 70 & 10 & 1.4e-12 \\              
		NaiveBayes\_weka & 113.84 & 26.49 & 14.08 & 51.54 & 70.67 & 70 & 10 & 8e-13 \\                         
		stepQDA\_caret & 114.34 & 27.16 & 14.26 & 45.82 & 70.12 & 75 & 6 & 5.6e-14 \\                          
		DecisionTable\_weka & 114.41 & 27.07 & 14.08 & 50.31 & 70.67 & 76 & 4 & 3.1e-14 \\                     
		QdaCov\_caret & 114.59 & 26.47 & 14.28 & 50.35 & 70.29 & 77 & 5 & 9.3e-14 \\                           
		kernelpls\_R & 114.73 & 39.72 & 14.08 & 40.72 & 70.67 & 71 & 9 & 3.6e-13 \\                            
		sparseLDA\_R & 114.96 & 30.47 & 13.96 & 43.73 & 70.61 & 72 & 9 & 3.2e-13 \\                            
		NaiveBayesUpdateable\_weka & 115.50 & 27.73 & 14.08 & 51.54 & 70.67 & 70 & 10 & 8e-13 \\               
		bayesglm\_caret & 116.20 & 42.40 & 14.08 & 34.14 & 70.67 & 72 & 6 & 4e-13 \\                           
		PenalizedLDA\_R & 116.24 & 32.62 & 14.08 & 42.79 & 70.67 & 66 & 13 & 2.6e-12 \\                        
		sddaQDA\_R & 116.75 & 29.97 & 14.08 & 41.12 & 70.67 & 77 & 4 & 4.4e-14 \\                              
		stepLDA\_caret & 117 & 27.88 & 14.08 & 43.57 & 70.48 & 78 & 3 & 1.3e-14 \\                             
		NaiveBayesSimple\_weka & 124.85 & 26.95 & 13.07 & 50.31 & 70.52 & 75 & 7 & 1.4e-13 \\                  
		glmStepAIC\_caret & 124.85 & 43.05 & 14.20 & 34.13 & 70.37 & 74 & 5 & 1.3e-13 \\                       
		LWL\_weka & 126.43 & 30.60 & 14.18 & 42.89 & 70.49 & 75 & 4 & 6.1e-14 \\                               
		gpls\_R & 126.52 & 45.94 & 14.84 & 33.86 & 69.16 & 71 & 7 & 2.4e-13 \\                                 
		dpp\_C & 127.03 & 31.59 & 14.08 & 49.22 & 70.67 & 68 & 12 & 3.3e-12 \\                                 
		AdaBoostM1\_weka & 128.16 & 37.45 & 14.08 & 36.59 & 70.67 & 75 & 4 & 3.7e-14 \\                        
		glm\_R & 130.88 & 51.04 & 14.08 & 31.62 & 70.67 & 72 & 8 & 2.8e-13 \\                                  
		Bagging\_HyperPipes\_weka & 133.59 & 36.56 & 14.08 & 32.96 & 70.67 & 80 & 1 & 6.5e-15 \\               
		MultiBoostAB\_weka & 133.60 & 38.57 & 14.08 & 33.58 & 70.67 & 76 & 3 & 2.2e-14 \\                      
		MultiBoostAB\_IBk\_weka & 133.60 & 38.57 & 14.08 & 31.80 & 70.67 & 76 & 3 & 2e-14 \\                   
		MultiBoostAB\_OneR\_weka & 133.72 & 35.98 & 14.08 & 36.85 & 70.67 & 78 & 2 & 8.8e-15 \\                
		Bagging\_OneR\_weka & 133.90 & 36.19 & 14.08 & 35.35 & 70.67 & 78 & 3 & 8.1e-15 \\                     
		VFI\_weka & 135.01 & 32.76 & 14.08 & 47.06 & 70.67 & 75 & 6 & 5e-14 \\                                 
		Bagging\_DecisionStump\_weka & 138.01 & 38.98 & 14.08 & 30.20 & 70.67 & 79 & 3 & 6.6e-15 \\            
		OneR\_caret & 138.07 & 37.68 & 14.08 & 38.16 & 70.67 & 77 & 5 & 1.1e-14 \\                             
		HyperPipes\_weka & 139.46 & 39.62 & 14.08 & 31.01 & 70.67 & 80 & 2 & 5.5e-15 \\                        
		OneR\_weka & 139.71 & 37.78 & 14.08 & 34.57 & 70.67 & 79 & 3 & 7.9e-15 \\                              
		spls\_R & 140.37 & 46.35 & 14.08 & 19.96 & 70.67 & 78 & 4 & 9.2e-15 \\                                 
		RacedIncrementalLogitBoost\_weka & 140.37 & 44.08 & 14.08 & 16.78 & 70.67 & 80 & 1 & 6.5e-15 \\        
		DecisionStump\_weka & 140.85 & 40.81 & 14.08 & 27.77 & 70.67 & 79 & 3 & 6.8e-15 \\                     
		ConjunctiveRule\_weka & 140.92 & 41.12 & 14.08 & 28.71 & 70.67 & 79 & 3 & 8.5e-15 \\                   
		Bagging\_MultilayerPerceptron\_weka & 143.11 & 46.63 & 14.08 & 16.33 & 70.67 & 77 & 3 & 1.8e-14 \\     
		StackingC\_weka & 154.88 & 53.05 & 14.10 & 3.43 & 70.62 & 80 & 1 & 6.2e-15 \\                          
		CVParameterSelection\_weka & 154.90 & 53.05 & 14.10 & 3.45 & 70.62 & 80 & 1 & 6.2e-15 \\               
		Grading\_weka & 154.90 & 53.05 & 14.10 & 3.45 & 70.62 & 80 & 1 & 6.2e-15 \\                            
		Stacking\_weka & 154.90 & 53.05 & 14.10 & 3.45 & 70.62 & 80 & 1 & 6.2e-15 \\                           
		MetaCost\_weka & 154.94 & 53.16 & 14.08 & 7.26 & 70.67 & 79 & 2 & 7e-15 \\                             
		CostSensitiveClassifier\_weka & 155.02 & 53.13 & 14.08 & 5.78 & 70.67 & 79 & 2 & 7.5e-15 \\            
		MultiScheme\_weka & 155.02 & 53.13 & 14.08 & 2.74 & 70.67 & 80 & 1 & 6.2e-15 \\                        
		Vote\_weka & 155.02 & 53.13 & 14.08 & 2.74 & 70.67 & 80 & 1 & 6.2e-15 \\                               
		ZeroR\_weka & 155.02 & 53.13 & 14.08 & 2.74 & 70.67 & 80 & 1 & 6.2e-15 \\                              
		ClassificationViaClustering\_weka & 156.78 & 46.81 & 14.08 & 28.61 & 70.67 & 79 & 2 & 7.8e-15 \\ 
	\end{longtable}                                                      
}

\vskip 0.2in
\bibliography{refs}

\end{document}